\documentclass[pdflatex,sn-mathphys-num]{sn-jnl} 
\usepackage{capt-of}
\usepackage{blindtext}
\usepackage{graphicx}
\usepackage{amsmath}
\usepackage{amsfonts}
\usepackage{scalerel}
\usepackage{xcolor}
\usepackage{geometry}
\usepackage{float}
\usepackage{doi}
\usepackage{enumitem}
\usepackage{multirow}
\usepackage{booktabs}\let\cline\cmidrule
\usepackage{caption}
\usepackage{multirow}
\usepackage{subcaption}
\usepackage{mwe}

\usepackage[export]{adjustbox}
\usepackage{graphicx,wrapfig,lipsum}
\usepackage{tikz}
\usepackage{xcolor}

\def\xb{\mathbf{x}}
\def\ub{\mathbf{u}}

\def\bx{\bar{\xb}}
\def\bu{\bar{\ub}}
\def\yb{\mathbf{y}}
\def\zb{\mathbf{z}}

\def\sz{\scaleto{(0)}{7pt}}

\def\sl{\scaleto{(l)}{7pt}}
\def\sL{\scaleto{(L)}{7pt}}
\def\slo{\scaleto{(l+1)}{7pt}}

\def\sitf{\scaleto{t+\delta t}{6.5pt}}

\definecolor{airforceblue}{rgb}{0.36, 0.54, 0.66}
\definecolor{bluebell}{rgb}{0.64, 0.64, 0.82}
\definecolor{arylideyellow}{rgb}{0.91, 0.84, 0.42}
\definecolor{Thistle}{rgb}{0.85, 0.75, 0.85}

\newcommand\mydots{\hbox to 1em{.\hss.\hss.}}
\newcommand{\Poincare}{Poincar\'e }

\newcommand\blfootnote[1]{%
  \begingroup
  \renewcommand\thefootnote{}\footnote{#1}%
  \addtocounter{footnote}{0}%
  \endgroup
}

\makeatletter
\renewcommand\footnotesize{%
   \@setfontsize\footnotesize\@ixpt{9}%
   \abovedisplayskip 7\p@ \@plus2\p@ \@minus4\p@
   \abovedisplayshortskip \z@ \@plus\p@
   \belowdisplayshortskip 4\p@ \@plus2\p@ \@minus2\p@
   \def\@listi{\leftmargin\leftmargini
               \topsep 4\p@ \@plus2\p@ \@minus2\p@
               \parsep 2\p@ \@plus\p@ \@minus\p@
               \itemsep \parsep}%
   \belowdisplayskip \abovedisplayskip
}
\makeatother

\begin{document}

\title[Article Title]{Connecting the geometry and dynamics of many-body complex systems with message passing neural operators}

\author[1]{ Nicholas A. Gabriel }

%\author[2]{Second Author}

\author[1]{ Neil F. Johnson }

\affil[1]{\orgdiv{Department of Physics}, \orgname{The George Washington University}, \orgaddress{ \city{Washington, DC}, \postcode{20052}}\vspace{2mm}}

\author[2]{George Em Karniadakis}

\affil[2]{\orgdiv{Division of Applied Mathematics}, \orgname{Brown University}, \orgaddress{ \city{Providence, RI}, \postcode{02912}}\vspace{0mm}}

%\affil[2]{\orgdiv{Department}, \orgname{Organization}, \orgaddress{ \city{City, State}, \postcode{10587}}}

%\affiliation{$^1$Department of Physics, George Washington University, Washington DC, USA }

\abstract{The relationship between scale transformations and dynamics established by renormalization group techniques is a cornerstone of modern physical theories, from fluid mechanics to elementary particle physics. Integrating renormalization group methods into neural operators for many-body complex systems could provide a foundational inductive bias for learning their effective dynamics, while also uncovering multiscale organization. We introduce a scalable AI framework, ROMA (Renormalized Operators with Multiscale Attention), for learning multiscale evolution operators of many-body complex systems. In particular, we develop a renormalization procedure based on neural analogs of the geometric and laplacian renormalization groups, which can be co-learned with neural operators. An attention mechanism is used to model multiscale interactions by connecting geometric representations of local subgraphs and dynamical operators. We apply this framework in challenging conditions: large systems of more than 1M nodes, long-range interactions, and noisy input-output data for two contrasting examples: Kuramoto oscillators and Burgers-like social dynamics. We demonstrate that the ROMA framework improves scalability and positive transfer between forecasting and effective dynamics tasks compared to state-of-the-art operator learning techniques, while also giving insight into multiscale interactions. Additionally, we investigate power law scaling in the number of model parameters, and demonstrate a departure from typical power law exponents in the presence of hierarchical and multiscale interactions.}

\maketitle
\section{Introduction}
Discovering effective dynamics of many-body complex systems remains challenging~\cite{Kemeth2022,de2024ai,vlachas2022multiscale,kaltenbach2020incorporating,mao2019nonlocal}. In many cases, learning coupled dynamics of complex systems implies solving an inverse problem based on nonlinear, infinite dimensional, and ill-conditioned relationships from incomplete, noisy, and multi-resolution data~\cite{karniadakis2021physics}. Neural operators~\cite{lu2021learning,wang2022improved,kovachki2023neural,cao2024laplace} are a robust data-driven approach to learning physical dynamics, but it is unclear how to leverage the multiscale structure~\cite{betzel2017multi,garcia2018multiscale,garuccio2023multiscale} of complex systems in such approaches, or if multiscale structure is present but not known in advance, how it can be co-learned with neural operators. Graph learning methods~\cite{chami2022machine}, including graph embedding~\cite{node2vec,xu2021understanding} and graph neural networks~\cite{kipf2016semi,GAT2018,MPNN2017,you2022nonlocal}, present a flexible way to learn low-dimensional representations of graph structured data suitable for a wide range of downstream tasks, such as learning coupled dynamics~\cite{hamilton2017inductive,jin2020sympnets}. The flexible nature of graph neural networks, and message passing models more generally, allows integration with neural operators~\cite{li2020neural,CViT2024,bryutkin2024hamlet} specifically aimed at learning the dynamics of many-body complex systems. In particular, message passing architectures representing hyperbolic geometry~\cite{ganea2018hyperbolic,liu2019hyperbolic,chami2019hyperbolic} and dynamics~\cite{cao2021choose,CViT2024,bryutkin2024hamlet} are well suited to represent the structural and functional aspects of complex systems, respectively. Composing these methods in a manner that is appropriate for many-body complex systems has not been investigated. In the remainder of this section, we outline current methods of renormalization for complex systems, and how neural analogs of such methods can be composed with neural operators to jointly uncover the structure and dynamics of complex systems, potentially enhancing both predictive performance and explainability. \blfootnote{${}^{1}$Code and replication data is available at \href{https://github.com/nngabe/roma}{\texttt{https://github.com/nngabe/roma}}} \\

The relationship between scale transformations and dynamics has played a significant role in the development of physical theories~\cite{gell1954quantum,wilson1971renormalization,pelissetto2002critical,verstraete2023density,kraichnan1982hydrodynamic,yakhot1986renormalization,polyakov1993theory,zhou2010renormalization}, described mathematically by renormalization group theory. The key insight is that certain properties of physical systems should be invariant at all scales of observation, and therefore one should be able to coarse grain microscopic observations to obtain effective equations of state and macroscopic parameters of the system. Generalizing renormalization procedures from physical space to the topological space of networks has proven challenging, largely due to correlations between coexisting scales, a phenomenon that co-occurs with the small world property in complex systems. Thus far, two main procedures have emerged to circumvent this difficulty, Geometric Renormalization (GR)~\cite{krioukov2010hyperbolic,garcia2018multiscale,boguna2021network,allard2023geometric} and Laplacian Renormalization (LR)~\cite{villegas2023laplacian,caldarelli2024laplacian}. In GR procedures, one embeds nodes in a latent geometric space, such as a Poincar\'{e} disk or Hyperboloid, where the hierarchical topology can be faithfully represented with much smaller latent dimension compared to Euclidean embeddings. This latent geometric space provides meaningful distances over which one can coarse grain the network while preserving network properties and reproducing behavior at each scale. In LR procedures, one can define coarse-graining operators using the top-$k$ ``slow" Laplacian eigenvectors, without needing to introduce a hidden metric space. The LR approach is systematic and readily understood in terms of well established spectral methods, while GR offers greater flexibility to represent hierarchical structures and provide additional insight beyond topological structure. A limitation of both methods is that there is no way to directly incorporate dynamical data into the inference procedures for coarse-graining operators, though multiscale simulation can be performed once the coarse-graining is determined. \\

Developing neural operators that can leverage the interplay between multiscale structure and dynamics is a principled extension of renormalization methods for complex systems, and the flexibility and composability of neural operators presents a systematic way of defining such methods. Conversely, renormalizability suggests an inductive bias for designing neural operators for complex systems, enabling the modeling of multiscale interactions and potentially improving the performance and explainability of neural operators. With these principles in mind, we develop a neural operator for many-body complex systems with the following elements: 
\begin{enumerate}[topsep=6pt,itemsep=6pt]
    \item[] \textbf{\underline{Renormalization}}: We develop a neural renormalization procedure analogous to geometric and Laplacian renormalization group methods. This renormalization procedure learns multiscale structure concurrently with effective dynamics and provides a basis for modeling multiscale interactions.
    \item[] \textbf{\underline{Multiscale interactions}}: We define multiscale neural operators using an attention mechanism to learn interactions across scales.
    \item[] \textbf{\underline{Scalability}}: By conditioning neural operators locally based on subgraph sampling, we enable batch processing of large graphs (up to 3M nodes in our experiments) and local approximations of high-dimensional dynamics.
    \item[] \textbf{\underline{Effective Dynamics}}: We approximate high-dimensional coupled ODEs with latent, low-dimensional PDEs based on local representations of subgraphs.
    \item[] \textbf{\underline{Robustness}}: We test our method under high uncertainty, 2-10\% noise, and compare the performance of linear and nonlinear neural operators within a renormalization framework.
\end{enumerate}
\vspace*{2mm}

%\section{Renormalized Operators with Multiscale Attention}

%------------------------------------------

\begin{wrapfigure}[25]{r}{7.2cm}
\vspace*{-10mm}
\includegraphics[width=7.8cm]{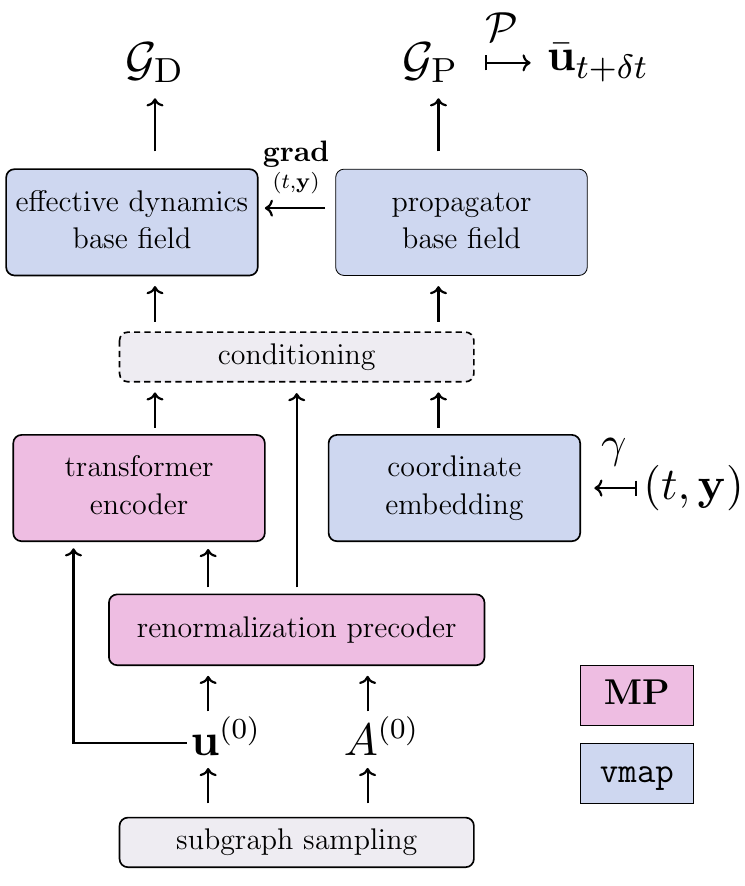}
\caption{Renormalized Operators with Multiscale Attention: The proposed model consists of message passing regions (pink) that conditions operators based on implicit representations of interactions, and vectorized regions (blue) that efficiently compute operators with \texttt{vmap} to forecast system evolution.}
\label{wrap-fig:roma_small}
\end{wrapfigure}

These specifications direct the design of ROMA (Renormalized Operators with Multiscale Attention), see Fig. \ref{wrap-fig:roma_small}, as well as our choice of benchmarks. 
We take the viewpoint of ~\cite{CViT2024} and formulate the ROMA architecture as a conditioned neural field, where the conditioning mechanism uses message passing to learn coupled representations based on a local subgraph. These local representations then modulate propagator and effective dynamics base fields that approximate the full, high-dimensional dynamics. This approach allows training in a batch fashion such that we can process large graphs and enable generalization to unseen subgraphs.\\

We investigate two systems with dynamics dominated by coupled, long-range interactions, representing a range of physical, biological, and social systems. The first is a Kuramoto model with locally normalized interactions 
\begin{equation}
\label{eq:KM}
\begin{aligned}
    \frac{d\phi_i}{dt} &= \omega_i + K\sum_{j}w_{ij}\textrm{sin}(\phi_j - \phi_i) \\
    w_{ij} &= \frac{A_{ij}}{\sum_j A_{ij}}, \\
    {}
\end{aligned}
\end{equation}
where $A_{ij}$ is an adjacency matrix. This model specification mimics many biological and social systems, particularly near the critical coupling $K \sim K_c$ ~\cite{kaiser2010optimal,odor2019critical}. This is a generalization of the classic Kuramoto model, which has an equally weighted, undirected, and fully connected graph. In the classic model configuration, and with sufficiently large coupling $K$ and system size $N$, all oscillators tend towards a globally synchronized phase with an effective frequency $\psi$. Exact solutions are available only in particular limits of system size and coupling strength, but properties such as synchronization conditions can be derived more generally from mean-field treatments~\cite{dorfler2014synchronization}. Generalizations of this model have been applied to understand a wide range of synchronization phenomena including neural dynamics ~\cite{odor2019critical,odor2021effect,anyaeji2021quantitative}, social dynamics~\cite{neda2000sound,strogatz2005crowd}, chemical oscillators~\cite{kiss2002emerging}, and optomechanical arrays~\cite{ludwig2013quantum}. \\

\noindent Solving the Kuramoto model in more general circumstances is still an open problem. We seek an approximate solution to the inverse problem for the Kuramoto model on large networks ($N >$ 1M nodes) near the critical coupling $K \sim K_c$,  which exhibits both long-range interactions and non-periodic dynamics, and is thereby outside of the realm of most numerical and analytical methods. This system configuration mirrors high-resolution connectome dynamics of the human brain~\cite{odor2019critical} and crowd dynamics~\cite{strogatz2000kuramoto,strogatz2005crowd,gu2025emergence}, among other complex systems. \\

\noindent We solve the inverse problem by approximating the high dimensional, coupled ODEs of equation \ref{eq:KM} as local PDE representations based on  subgraphs of size $N_B \ll N$ in emergent coordinates $\mathbf{y} \in \mathbb{R}^d$~\cite{Kemeth2022}
\begin{subequations}
\label{eq:pde}
\begin{align}
        \bar{\sigma}(t,\yb) &= \mathcal{G}_{\textrm{P}}(v)\big(t,\yb;\bar{\mathbf{b}},\bar{\mathbf{x}}\big) \\
        \cfrac{\partial \bar{\sigma}}{\partial t} &\simeq \mathcal{G}_{\textrm{D}}(v)\big(t, \mathbf{y};\bar{\mathbf{b}}, \bar{\xb}, \bar{\sigma}, \nabla \bar{\sigma}, \nabla^2 \bar{\sigma} \big),  
\end{align}
\end{subequations}
where $\bar{\mathbf{x}}$ and $\bar{\mathbf{b}}$ are multiscale representations computed by the renormalization precoder and transformer encoder in Fig.~\ref{wrap-fig:roma_small}, respectively. We note that equation (\ref{eq:pde}b) indicates asymptotic equality based on minimization of a physics-informed residual\footnote{Specifically, the left-hand-side of equation (\ref{eq:pde}b) is computed by automatic differentiation (AD) of (\ref{eq:pde}a), and the right-hand-side of (\ref{eq:pde}b) is an independently computed operator. Equality is enforced by minimizing the residual $|\partial_t\bar{\sigma}-\mathcal{G}_{\textrm{D}}|$.}. Each  quantity computed by a message passing module consists of a fine-grained quantity (e.g. $\mathbf{x}^{(0)}$) concatenated with its coarse-grained counterpart (e.g. $\mathbf{x}^{(1)}, \mathbf{x}^{(2)}\ldots$), i.e., 
\begin{equation}
    \begin{aligned}
        \bar{\sigma} &= [\sigma^{(0)}, \, \sigma^{(1)}, \, \ldots] , \\
        \bar{\mathbf{x}} &= [{\mathbf{x}^{(0)}}, \, {\mathbf{x}^{(1)}}, \, \ldots]  , \ \ \textrm{and} \\
        \bar{\mathbf{b}} &= [{\mathbf{b}^{(0)}}, \, {\mathbf{b}^{(1)}}, \, \ldots]. \\
    \end{aligned}
\end{equation}
The effective dynamics are projected by an operation $\mathcal{P}: \mathbb{R}^{d} \rightarrow \mathbb{R}$ to obtain the multiscale forecast
\begin{equation}
    \bar{\mathbf{u}}_{\sitf} = \mathcal{P}\big(\bar{\sigma}\big). \vspace*{3mm} 
\end{equation}

\noindent This approach aims to connect multiscale structure with dynamics in a generalizable fashion through a low-dimensional continuous field $\bar{\sigma}(t,\yb)$ as follows:

\begin{enumerate}[topsep=4pt,itemsep=7pt,partopsep=4pt]
    \item[1.] \underline{Multiscale Context Representation}: The dynamical and topological context of a sampled subgraph $(\mathbf{u}^{\sz},A^{\sz})$ is encoded as coupled, multiscale graph representations $\bar{\mathbf{x}}$ and $\bar{\mathbf{b}}$.
    \item[2.] \underline{Conditioning}: The low-dimensional coordinates $(t,\mathbf{y})$ are modulated by $\bar{\mathbf{x}}$ and $\bar{\mathbf{b}}$ in a conditioning mechanism such that the gradients $\partial_t \bar{\sigma}$ and $\nabla \bar{\sigma}$ encode contextual information.
    \item[3.] \underline{Physics-Informed Manifold Learning}: The low dimensional gradients $\partial_t \bar{\sigma}$ and $\nabla \bar{\sigma}$ are used to learn effective dynamics in equation (\ref{eq:pde}) that regularize the multiscale forecast $ \bar{\mathbf{u}}_{\sitf} = \mathcal{P}\big(\bar{\sigma}\big)$ via a physics-informed loss function.
\end{enumerate}
Importantly, this formulation allows graph learning methods that can scale to billions of nodes~\cite{hamilton2017inductive,ying2018graph,zeng2019graphsaint} to be tightly integrated with physics-informed learning. Hence, we can learn effective dynamics by training on large-scale complex systems of unprecedented size using fully data-driven methods, and perform inference based on local and partial observations of the system. \\

As a second example, we investigate the formation of social groups on social media platforms~\cite{manrique2018generalized,manrique2023shockwavelike}. In general, the formation of social groups is an emergent phenomenon resulting from pairwise relations such as physical, topological, and socio-cultural distances. Alongside the formation of macro-scale social units, coexisting structure can be found at the micro- and meso-scales due to homophily, clustering, and hierarchality. Multiscale interactions—the mechanism by which phenomena at various scales are related—are a defining feature of complex systems. The aggregation of individuals in the presence of multiscale interactions can exhibit shockwavelike solutions~\cite{manrique2018generalized,manrique2023shockwavelike}, leading to rapid and potentially harmful changes in public opinion with little or no time for intervention by social platforms, governments, or regulatory institutions~\cite{johnson2016new,johnson2020online}. In the context of online interactions, such phenomena can be accelerated and proliferated by content filters, recommender systems, AI powered bots, and inauthentic actors acting on behalf of commercial or geopolitical interests~\cite{avalle2024persistent,huo2024multi,gabriel2023inductive}. \\

Understanding the dynamics of social systems is perhaps the foremost objective of social physics, with broader implications for complex systems. On one hand, the volume of interactions on social platforms is by far the most extensive record of human activity in history, with a high degree of dynamical regularity. However, on the other hand, the dynamics of human interactions take place in non-ergodic, irreversible, and open systems, rendering mean-field treatments challenging, and bottom-up approaches marginal.  At the same time, many near-universal features of social systems such as homophily, clustering, hierarchality, power-law distributions of events, and locally sustained activity, among others, suggest an underlying structure that gives rise to observed dynamical regularities. This situation potentially makes data-driven effective dynamics an appealing alternative, as one can leverage the presence of physical regularities without needing specifying them a priori, and instead introducing them as inductive biases in the architecture or loss function. \\

We propose a benchmark based on social group formation as follows. We begin with $N$ stochastic realizations of shock-wave solutions satisfying a Burgers equation~\cite{manrique2023shockwavelike} 
\begin{equation}
    \frac{\partial u_i}{\partial t} = - F(t) \, \frac{\partial u_i}{\partial x} u_i + \nu(t) \, \frac{\partial^2 u_i}{\partial x^2},  
\end{equation}
where $F(t)$ and $\nu(t)$ are monotonic functions, and each $u_i$ represents the population of a ``hard"  social community such as in-built groups or a ``soft" social community defined by co-interactions with content. Nodes are then coupled through a nonlinear diffusion process
\begin{equation}
\label{eq:BD}
    \frac{\partial u_i}{\partial t} \ \leftarrow \ \beta\frac{\partial u_i}{\partial t} + (1-\beta) \, \sqrt{ \frac{1}{k_i}\sum_{j\in \mathcal{N}(i)} \bigg(\frac{\partial u_j}{\partial t}\bigg)^2 }, 
\end{equation}
where $k_i$ is the (in-)degree and $\mathcal{N}(i)$ the 1-hop neighborhood of the $i$th node. This equation can be iterated $m$ times to produce nonlinear interactions where $m$ and $\beta$ jointly control the range and magnitude of the interactions. In addition to the unsupervised dynamics in equation \ref{eq:pde}, we investigate semi-supervised solutions to equation \ref{eq:BD} of the form
\begin{equation}
\begin{aligned}
\label{eq:pde2}
        \bar{\sigma}(t,\yb) &= \mathcal{G}_{\textrm{P}}(v)\big(t,\yb;\bar{\mathbf{b}},\bar{\mathbf{x}}\big) \\
        \cfrac{\partial \bar{\sigma}}{\partial t} &\simeq - \bar{F} \, (\bar{\sigma} \cdot \nabla) \bar{\sigma} + \bar{\nu} \, \nabla^2 \bar{\sigma} \\
        \bar{\mathbf{u}}_{\sitf}& \, = \  \mathcal{P}\big(\bar{\sigma}\big), 
\end{aligned}
\end{equation}
where $\bar{F}$ and $\bar{\nu}$ are operator-valued coefficients
\begin{equation}
    \begin{aligned}
        \bar{F} &= \mathcal{G}^0_{\textrm{D}}(v)\big(t, \mathbf{y};\bar{\mathbf{b}},\bar{\xb}\big)  \\
        \bar{\nu} &= \mathcal{G}^1_{\textrm{D}}(v)\big(t, \mathbf{y};\bar{\mathbf{b}},\bar{\xb}\big)  \\
    \end{aligned}
\end{equation}
that define a dynamical Burgers equation in emergent coordinates $\mathbf{y}$.
\section{Results}

We compare several neural operators on inverse problems of varying system size, signal-to-noise ratios, and underlying dynamics. We select baseline methods that have simple message passing capabilities (Gaussian Process), no message passing capabilities (DeepONet), as well as baselines with the core message passing capabilites of ROMA, but without renormalization and/or effective dynamics. This allows us to investigate both the advantages of message passing architectures in modeling coupled dynamics, as well as the additional benefit from modeling multiscale interactions in the ROMA architecture. Each neural operator can be summarized as follows: \\

\begin{itemize}[topsep=0pt,partopsep=6pt,itemsep=6pt]

    \item[] \textbf{Gaussian Process (GP)}: As a baseline for message passing, we train and test a GP in-sample with a radial basis function (RBF) kernel to compare how the attention mechanism performs relative to an ideal kernel.

    \item[] \textbf{DeepONet (DON)}: A baseline DeepONet architecture with ResNet blocks and Fourier feature encoding.

    \item[] \textbf{DeepONet-MP (DON-MP)}: A standard DeepONet architecture with a baseline conditioning mechanism consisting of a Euclidean graph network precoder and a transformer encoder.

    \item[] \textbf{NOMAD-MP}: A nonlinear neural operator with the same baseline conditioning mechanism as DeepONet-MP.

    \item[] \textbf{ROMA}: A DeepONet architecture with a multiscale conditioning mechanism consisting of a hyperbolic graph network precoder and transformer encoder, as well as effective dynamics parameterized by a base field and modulated by the conditioning mechanism. \vspace*{3mm}
    
\end{itemize}
Notably, the only architectural difference between DeepONet-MP and ROMA is the inclusion of renormalization inspired elements and an effective dynamics base field.
Additionally, for examples examining performance benefits of effective dynamics, we have physics-informed variants of the two baseline neural operators with message passing: \\

\begin{itemize}[topsep=0pt,partopsep=6pt,itemsep=6pt]

    \item[] \textbf{DeepONet-MP-PI}: A DeepONet architecture with a baseline conditioning mechanism consisting of a Euclidean graph network precoder and transformer encoder, as well as effective dynamics parameterized by a base field and modulated by the conditioning mechanism.

    \item[] \textbf{NOMAD-MP-PI}: A nonlinear neural operator with a baseline conditioning mechanism consisting of a Euclidean graph network precoder and transformer encoder, as well as effective dynamics parameterized by a base field and modulated by the conditioning mechanism.
    
\end{itemize}

\noindent For each experiment, we configure each neural operator such that the number of parameters is equal for a given experiment. For example, the size of the branch and trunk nets of DON, DON-MP, and NOMAD-MP are increased relative to ROMA, which has additional parameters for the coarse-graining networks and effective dynamics base field. \\
\newpage

\vspace*{-4mm}

\begin{wrapfigure}[13]{r}{7.6cm}
\vspace*{-15mm}
\begin{center}
\captionof{table}{Details of the Kuramoto Model (KM) and Burgers Dynamics (BD) datasets. \\ }
      \begin{itemize}[leftmargin=.0cm,topsep=1.8cm,itemsep=0.6mm]
        \item[]{\makebox[8mm]{$N$} -- number of nodes.}
        \item[]{\makebox[8mm]{$T$} -- number of points in the time domain.}
        \item[]{\makebox[8mm]{$N_B$} -- number of nodes per training subgraph.}
        \item[]{\makebox[8mm]{$\alpha$} -- power law exponent of degree distribution.}
        \item[]{\makebox[8mm]{$C$} -- average clustering coefficient.}
        \item[] 
      \end{itemize}
\begin{tabular}{lrrrrr}
\hline \vspace*{-2mm}\\
\multicolumn{1}{c}{ \textbf{Dataset} \ \ \ \ }  & $N$ & $T$ & $N_B$ & $\alpha$ & $C$ \vspace*{1mm} \\
\hline \vspace*{-3.5mm} \\
KM-38k & 38,289 & 1,600 & 512 & 2.29 & 0.69    \\
KM-314k & 314,923 & 1,000 & 512 & 2.24 & 0.57    \\
KM-3M & 3,153,074 & 800 & 512 & 2.24 & 0.52   \\
BD-3M & 3,153,074 & 936 & 512 & 2.24 & 0.52   \\
\hline \vspace*{-0mm}

\label{tab:dataset}
\end{tabular}
\end{center}
%$$
%       \textrm{Rel. } L^2(\bar{\mathbf{u}}^{\sz},{\mathbf{u}}^{\sz}) = \sqrt{ \frac{1}{ {N_B}}  \sum_{i}^{N_B} \Bigg( \cfrac{ \big|\bar{\mathbf{u}}^{\sz}_{i,\sitf} - {\mathbf{u}}^{\sz}_{i,\sitf}\big| }{\big|\bar{\mathbf{u}}^{\sz}_{i,\sitf}| + |{\mathbf{u}}^{\sz}_{i,\sitf}\big|} \Bigg)^2   }
%$$
\end{wrapfigure} 
%\vspace*{-22mm}

\subsection{Scaling and Noise}
A first set of experiments evaluates the performance of neural operators under varying system size and noise. Specifically, we report prediction accuracy of each model on the fine-grained scale as the symmetric~\cite{hyndman2006another,makridakis2020m4,van2023all} Relative $L^2$ error~\cite{li2020neural,CViT2024} $$ \textrm{Rel. } L^2 = \sqrt{ \frac{1}{ {N_B}}  \sum_{i}^{N_B} \Bigg( \cfrac{ \big|\bar{\mathbf{u}}^{\sz}_{i,\sitf} - {\mathbf{u}}^{\sz}_{i,\sitf}\big| }{\big|\bar{\mathbf{u}}^{\sz}_{i,\sitf}| + |{\mathbf{u}}^{\sz}_{i,\sitf}\big|} \Bigg)^2   }
$$ for several Kuramoto models with varying system size and noise in Table \ref{tab:scaling}. This allows us to examine which choice of architecture and parameter allocation is most effective under realistic conditions. Namely, the first three columns of Table \ref{tab:scaling} show results of solving Kuramoto models of varying size, and the last two columns show results of increasing noise on input and output training data from moderate (2\%) to high (10\%). \\ 

\begin{figure}[H]

\centering
\hspace*{-20mm}
\includegraphics[width=.46\textwidth]{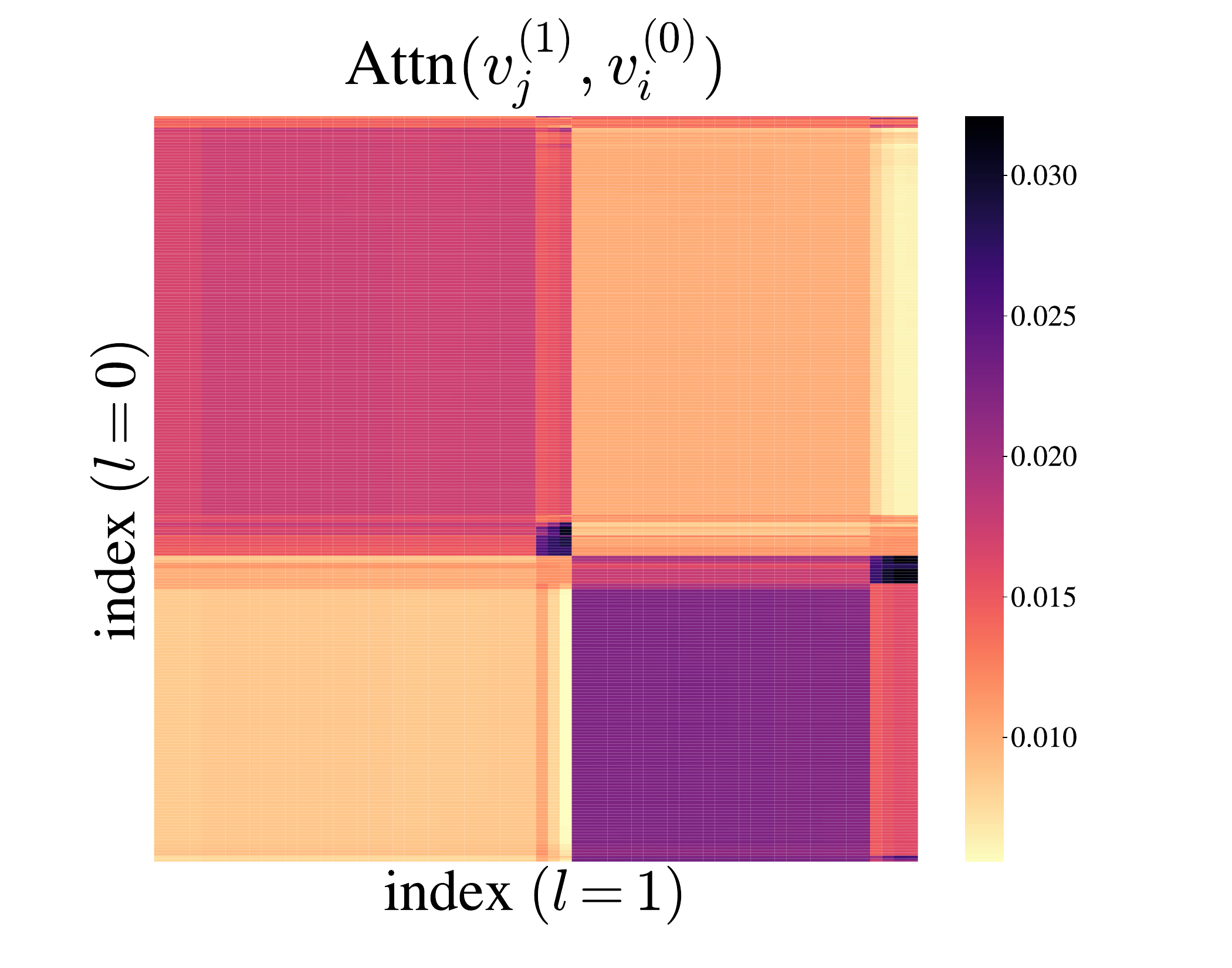}\hspace*{-9mm}
\includegraphics[width=.46\textwidth]{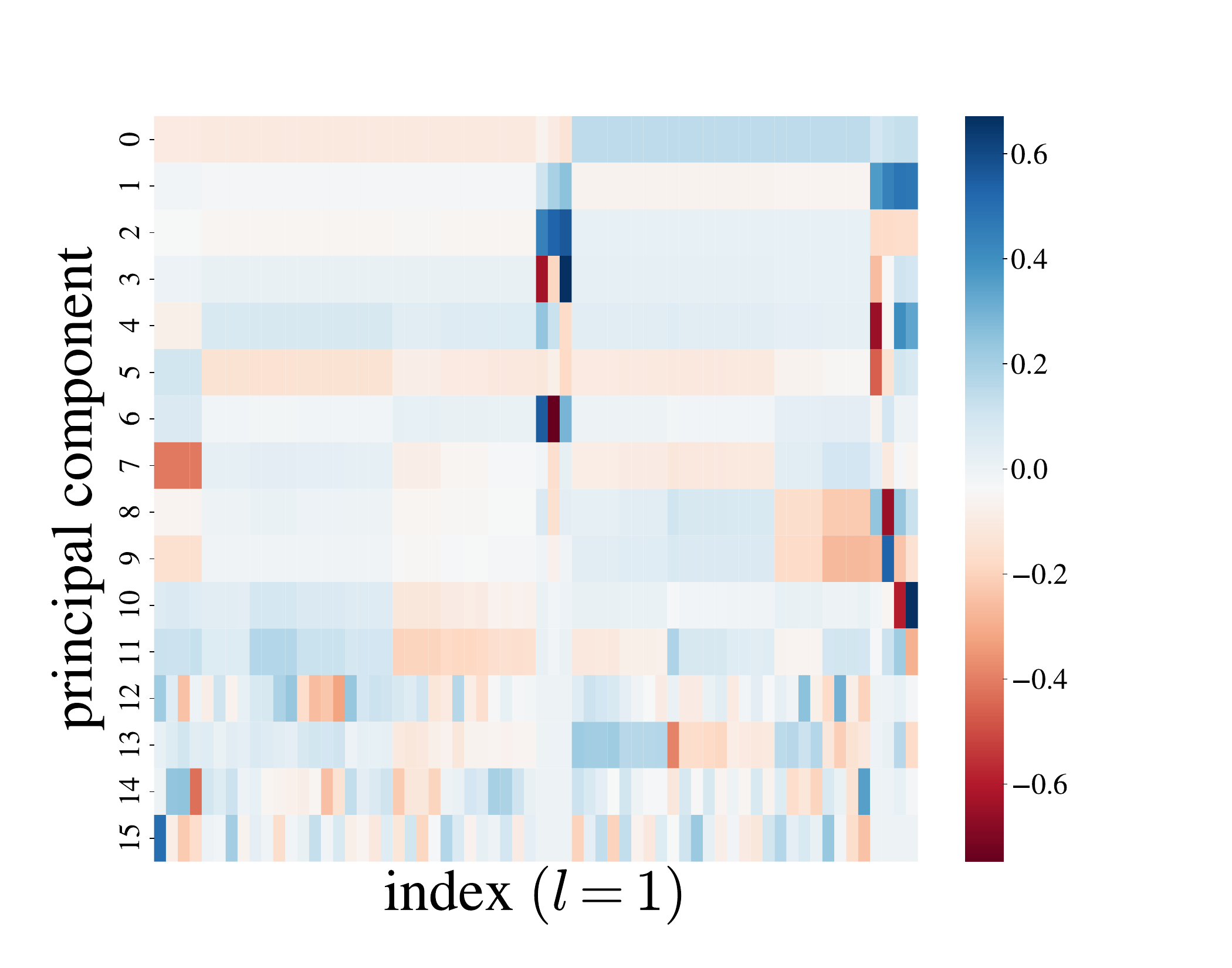}\hspace*{-9mm}
\includegraphics[width=.45\textwidth]{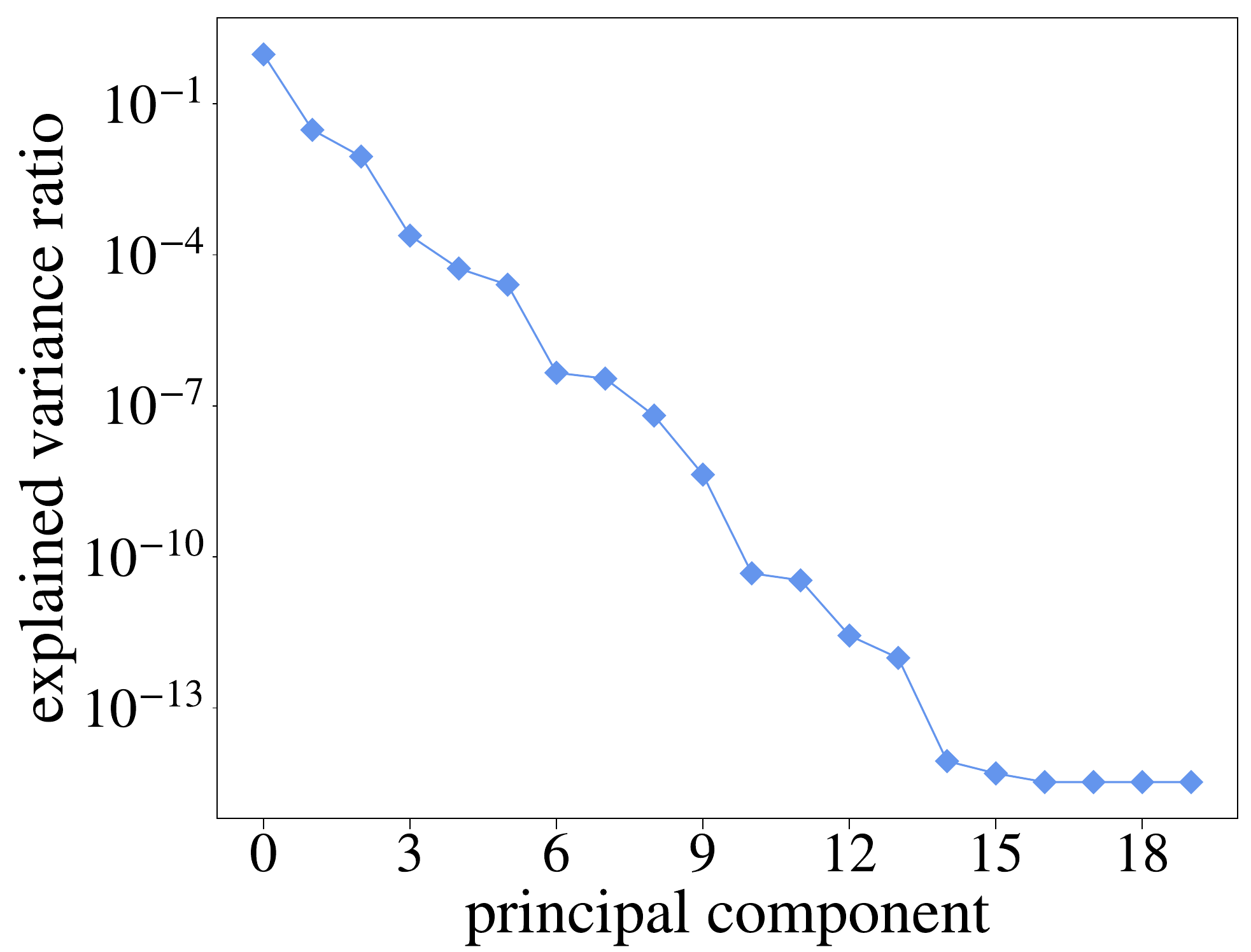}

\caption{\textbf{Left:} Average attention score from coarse-grained ($l=1$) to fine-grained ($l=0$) nodes in the first ROMA transformer block (KM-3M). \textbf{Center:} First 16 principle components of the attention scores. \textbf{Right:} Explained variance ratios of the first 16 principle components.}
\label{fig:attention}

\end{figure}

To illustrate how the ROMA architecture uses multiscale structure to enhance predictions on the fine-grained scale, we show the attention weights of the first attention block from the coarse-grained scale to the fine-grained scale in Fig. \ref{fig:attention} (Left), which are clustered to show functional units resulting from the training process. In Fig. \ref{fig:attention} (Center), the first 16 principal components of the attention weights are shown, with the explained variance ratio of each component shown in Fig. \ref{fig:attention} (Right). \\

\begin{center}

\captionof{table}{Performance of 1-step forecast on fine-grained scale, $\textrm{Rel. } L^2(\bar{\mathbf{u}}^{\sz},{\mathbf{u}}^{\sz})$. }
\begin{tabular}{lccccc}
\toprule  \multirow{2}{*}{ \ } & \multicolumn{4}{c}{ \textbf{Dataset} } \\
\cline{ 2-5 }  { \textbf{Model} \ \ \ \ \ \   } & \textbf{KM-38k} & \textbf{KM-314k} & \textbf{KM-3M} & \textbf{KM-3M}  & \textbf{\# Params}   \\ 
\hline \vspace*{-3.5mm} \\ 
GP    &${2.16 \!\times\! 10^{-3}}$  &$2.44 \!\times\! 10^{-3}$          & $2.35 \!\times\! 10^{-3}$       &$6.98 \!\times\! 10^{-3}$         & 226k\\
DON       & $7.90 \!\times\! 10^{-2}$          &$9.46\!\times\! 10^{-2}$           & $8.33\!\times\! 10^{-2}$        & $8.33 \!\times\! 10^{-2}$        & 264M\\
DON-MP    &$\mathbf{1.58\!\times\!10^{-3}}$             &$\mathbf{1.88 \!\times\! 10^{-3}}$ & $1.61 \!\times\! 10^{-3}$       & ${3.27 \!\times\! 10^{-3}}$        & 264M\\
NOMAD-MP  & ${1.68 \!\times 10^{-3}}$            &$2.29 \!\times\! 10^{-3}$          & $4.28 \!\times\! 10^{-3}$       & ${3.26 \!\times\! 10^{-3}}$        & 264M\\
ROMA      & $2.14 \!\times 10^{-3}$            & $2.46\!\times\!10^{-3}$           &$\mathbf{1.46\!\times\!10^{-3}}$ & $\mathbf{3.13\!\times\!10^{-3}}$ & 264M\\
\hline
std($\eta$)&  2 \%                             & 2 \%                              &      2 \%                        &    10 \%  \vspace*{0mm} 
%\multicolumn{2}{c}{$^*$In sample training/testing \hspace*{2mm}} &&&
\label{tab:scaling}
\end{tabular}
\vspace*{5mm}
\end{center}

\nopagebreak
Upon inspection of the multiscale attention weights in Fig.  \ref{fig:attention}, we note that the attention mechanism segments the coarse-grained nodes into functional units of varying size. The components with the highest explained variance have larger functional units containing dozens of nodes, while lower components have functional units of 3-4 nodes. These smaller functional units could be interpreted as high-frequency refinements to the low-frequency attentional units of the first few components.  \\
\nopagebreak
\vspace*{-4mm}

\newpage

\subsection{Effective Dynamics}
In order to understand the interplay between renormalization, effective dynamics, and forecasting performance, we benchmark ROMA against several physics-informed baselines, which have effective dynamics parameterized by a base field in addition to message passing capabilities of DON-MP and NOMAD-MP. In particular, DON-MP-PI has the same architecture as ROMA besides renormalization and multiscale attention, thereby isolating the impact of these elements on performance.  Overall, we see that in all message passing models, successful minimization the PDE residual 
\begin{equation}
\begin{aligned}
    \textrm{RMSE}(\partial_t\bar{\sigma}, \, \mathcal{G}_{\textrm{D}}) = \sqrt{ \frac{1}{N_B}\sum_{i}^{N_B} \bigg|\frac{\partial \bar{\sigma}}{\partial t} - \mathcal{G}_{\textrm{D}} \bigg|^2  } \vspace*{-2mm}
\end{aligned}
\end{equation}
generally corresponds to smaller errors in the forecasting task. However, this  coupling between the forecasting and PDE tasks appears to be architecture dependent, and the PDE residual in some cases is minimized without positive transfer to the forecasting task. This is clearly seen in the DON-PI and NOMAD-MP-PI performance for the KM-3M example, and to a lesser extent in the DON-MP-PI performance for BD-3M. In both examples, the multiscale representations learned by the transformer encoder in ROMA appear to facilitate greater positive transfer between the forecasting and PDE tasks.   \\

\begin{center}
\captionof{table}{Performance of 1-step forecast and PDE residual : $\textrm{Rel. } L^2(\bar{\mathbf{u}}^{\sz},{\mathbf{u}}^{\sz})$ and $\textrm{RMSE}(\partial_t\bar{\sigma}, \, \mathcal{G}_{\textrm{D}})$. The number of parameters is 264M and 1B for KM-3M and BD-3M, respectively. }
\begin{tabular}{lcccccc}

\toprule \vspace*{-4mm} \\ % \multirow{3}{*}{  } & & \multicolumn{3}{c}{ \textbf{Data Set}} \\
    & \multicolumn{2}{c}{ \textbf{KM-3M} }  &\multicolumn{4}{c}{ \textbf{BD-3M}}  \\
 \cline{ 2-3 }  \cline{ 5-6 } \multicolumn{1}{l}{ \textbf{Model} \ \ \ \ } & \underline{Rel. $L^2_{}$} & \underline{RMSE$_{\textrm{PDE}}$} &  & \underline{Rel. $L^2_{}$} & \underline{RMSE$_{\textrm{PDE}}$} &  \vspace{1mm}  \\ \hline \vspace*{-3.5mm} \\ 

DON-PI & $8.33 \!\times\! 10^{-2}$ & $\mathbf{8.48 \!\times\! 10^{-6}}$  &      & $2.88 \!\times\! 10^{-2}$ & $1.03 \!\times\! 10^{-3}$  &  \\
%DON-MP & $1.61 \!\times\! 10^{-3}$ & - & $\mathbf{1.63 \!\times\! 10^{-4}}$ & -  \\
DON-MP-PI & $2.27 \!\times\! 10^{-3}$ & $1.05 \!\times\! 10^{-3}$ &      & ${9.85 \!\times\! 10^{-3}}$ & $\mathbf{2.48 \!\times\! 10^{-9}}$ &  \\
%NOMAD-MP & $4.28 \!\times\! 10^{-3}$ & - & $1.37 \!\times\! 10^{-3}$ & -  \\
NOMAD-MP-PI & $2.68 \!\times\! 10^{-3}$ & $1.22 \!\times\! 10^{-4}$     &   & $3.06 \!\times\! 10^{-2}$ & ${2.74 \!\times\! 10^{-8}}$ &  \\
ROMA & $\mathbf{1.46 \!\times\! 10^{-3}}$ & $3.12 \!\times\! 10^{-5}$  &   & $\mathbf{9.81 \!\times\! 10^{-3}}$ & $6.32 \!\times\! 10^{-9}$ &  \\

\hline

\label{tab:effective}
\end{tabular}
\end{center}

\vspace*{-2mm}
\subsection{ROMA Scaling}

Three configurations for ROMA variants are shown in Table \ref{tab:roma_configs}. Namely, we investigate the performance of ROMA under increasing model size. For both sets of experiments, model performance was most influenced by the embedding dimension in the transformer encoder and the dimension of the output basis $p$, with little performance benefit beyond 7 layers and 8 attention heads. This scaling specifically in transformer width is likely due to the product network architecture of DeepONet~\cite{lu2021learning,jin2022mionet}, whereas scaling is largely insensitive to shape for standard transformer models~\cite{kaplan2020scaling}.\\

\begin{center}
\captionof{table}{Details of ROMA model variants for KM-3M/BD-3M.}
\begin{tabular}{lccccc}

\toprule \vspace*{-3.5mm} \\
\multicolumn{1}{l}{ \textbf{Model} \ \ \ \ }  &  \textbf{Encoder Layers} & \textbf{Embedding dim} & \textbf{MLP width} & \textbf{Heads} & \textbf{Basis dim $p$}    \\
 \midrule \vspace*{-3.5mm} \\ 

ROMA-B & 7 & 1024 & 4096 & 8 & 2048 \\
ROMA-L & 7 & 2048 & 8192 & 8 & 2048  \\
ROMA-H & 7 & 2560 & 10240 & 8 & 2560/2048 \vspace*{0mm}  \\ 
\bottomrule

\label{tab:roma_configs}
\end{tabular}
\end{center}

\begin{figure}[H]
        \centering
        \hspace*{-12mm}
        \begin{subfigure}[b]{0.48\textwidth}
            \centering
            %\hspace*{-16mm}
              %\includegraphics[width=1.2\textwidth,center]{image}
            \includegraphics[width=1.2\textwidth]{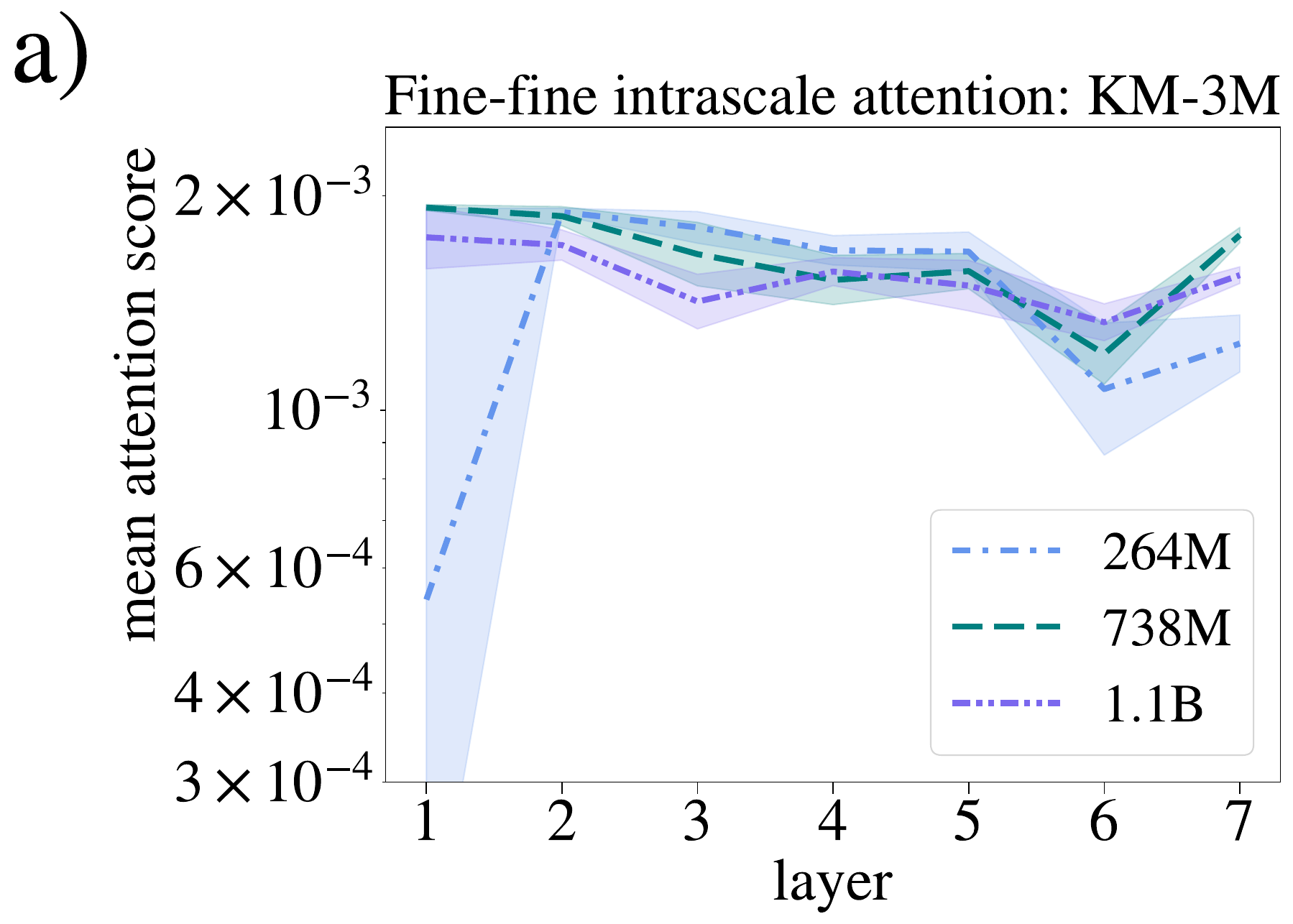}   
            \label{fig:mean and std of net14}
        \end{subfigure}
        \hfill 
        \begin{subfigure}[b]{0.48\textwidth}  
            \centering 
            \hspace*{-0mm}
            \includegraphics[width=1.2\textwidth]{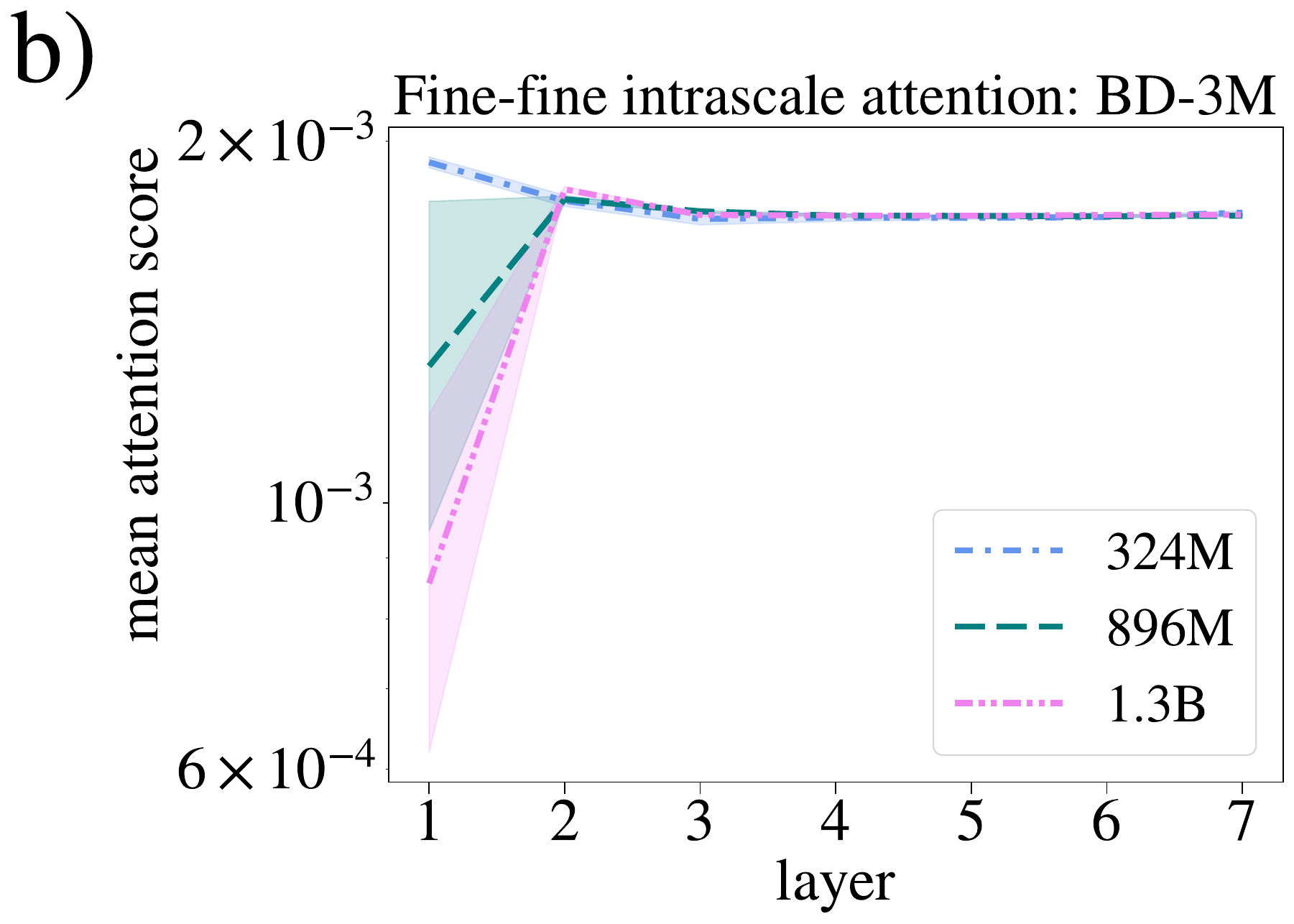} 
            \label{fig:mean and std of net24}
        \end{subfigure}
        \vspace*{-8mm}  \vskip\baselineskip
        \hspace*{-12mm}
        \begin{subfigure}[b]{0.48\textwidth}   
            \centering
            \includegraphics[width=1.2\textwidth]{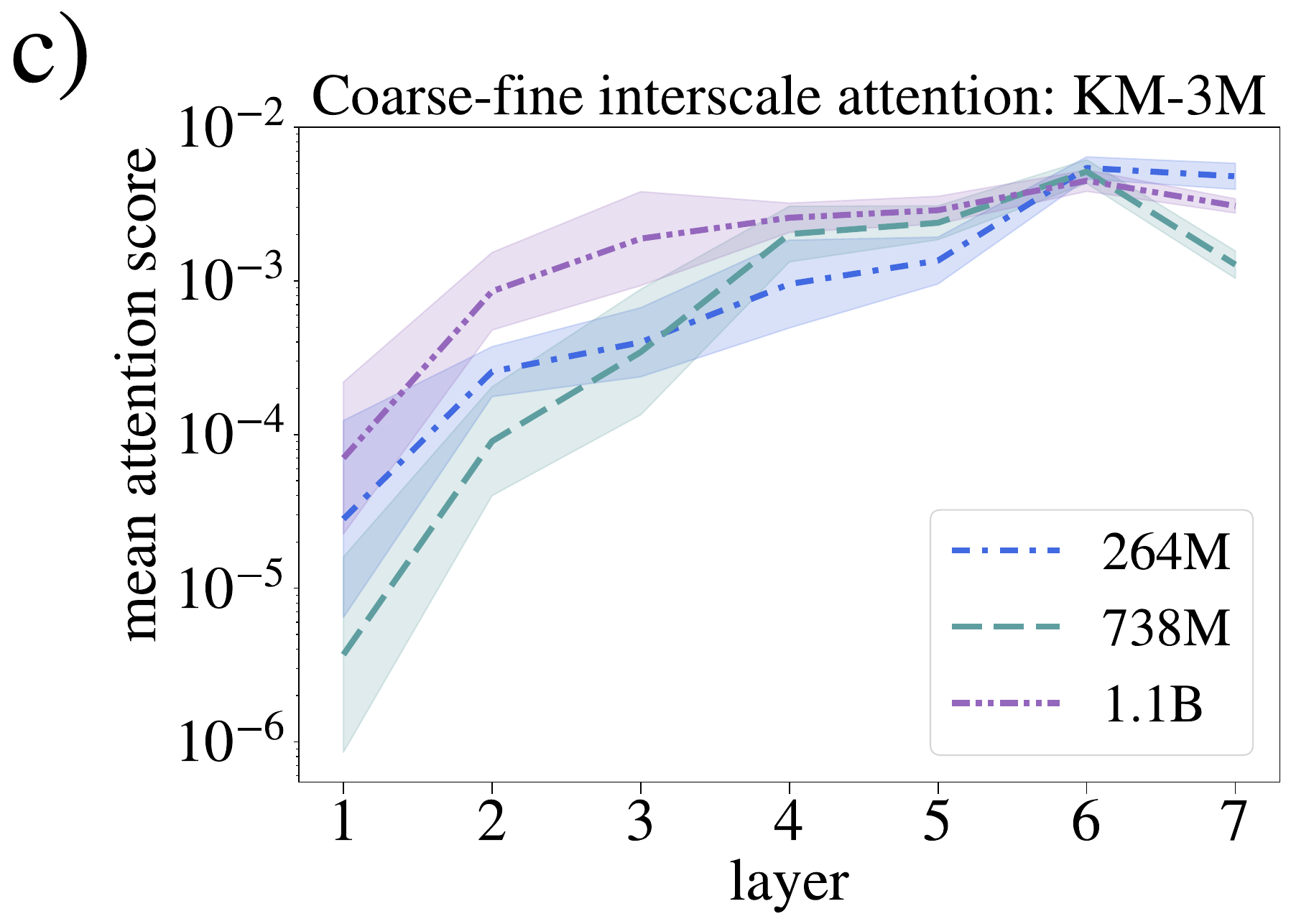} 
            \label{fig:mean and std of net34}
        \end{subfigure}
        \hfill
        \begin{subfigure}[b]{0.48\textwidth}   
            \centering 
            \hspace*{-0mm}
            \includegraphics[width=1.2\textwidth]{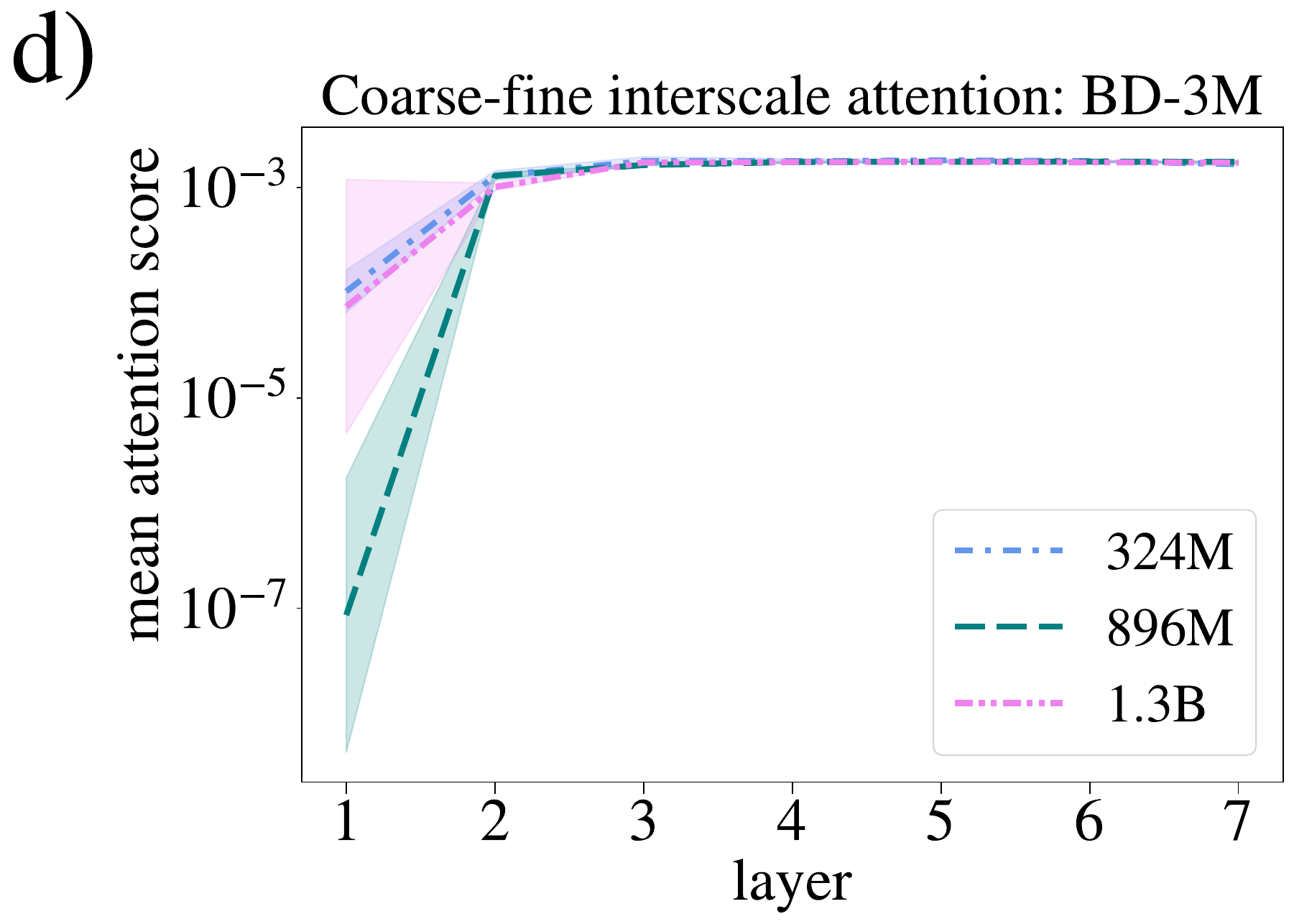}   
            \label{fig:mean and std of net44}
        \end{subfigure}
                \vspace*{-8mm}  \vskip\baselineskip
            \hspace*{-12mm} 
        \begin{subfigure}[b]{0.48\textwidth}   
            \centering
            \includegraphics[width=1.2\textwidth]{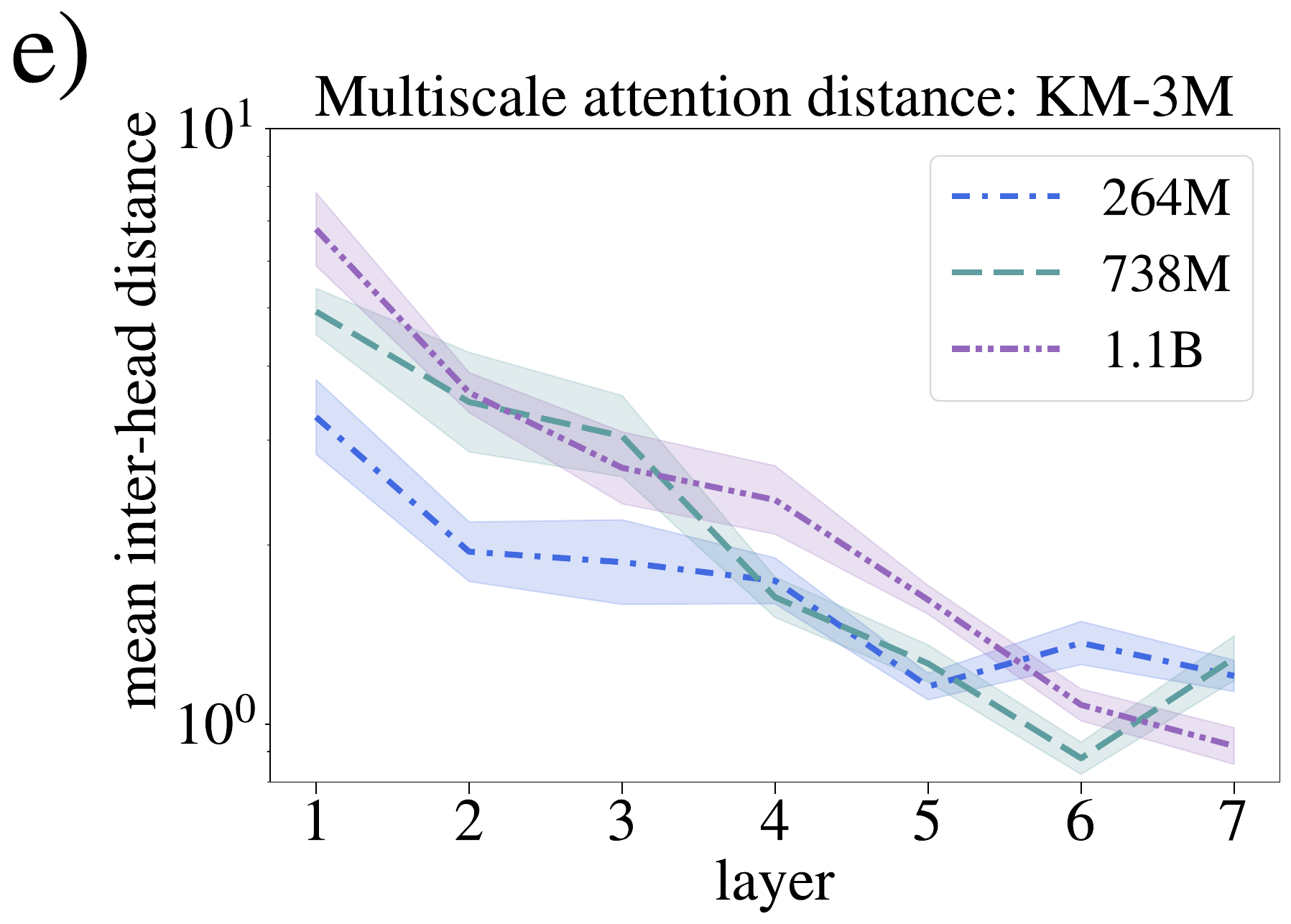} 
            \label{fig:mean and std of net54}
        \end{subfigure}
        \hfill
        \begin{subfigure}[b]{0.48\textwidth}   
            \centering 
            \hspace*{-0mm}
            \includegraphics[width=1.2\textwidth]{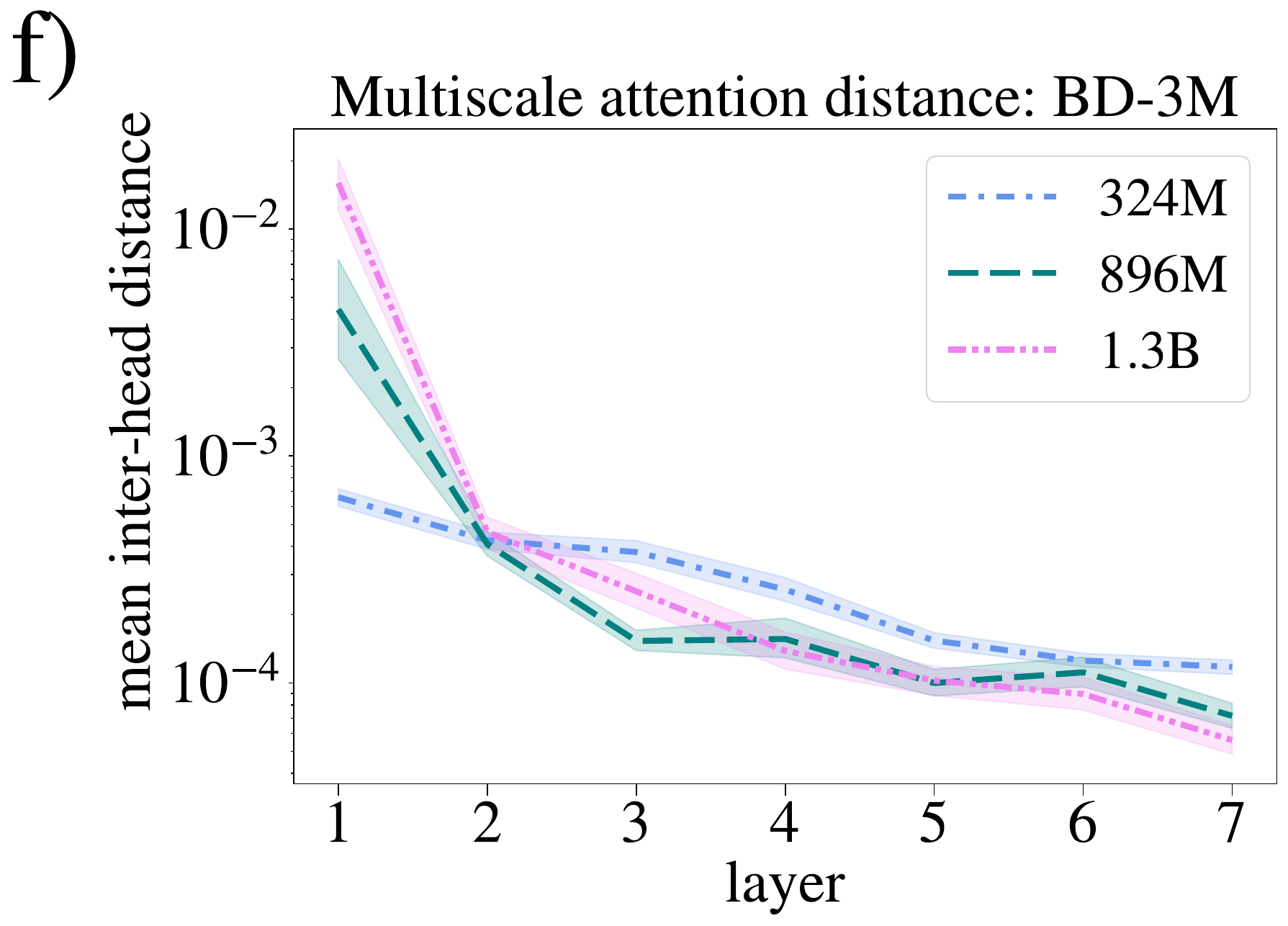}   
            \label{fig:mean and std of net64}
        \end{subfigure}
        \caption[ The average and standard deviation of critical parameters ]
        {\small Headwise attention statistics: (a-b)  per-head fine to fine attention, (c-d) per-head coarse to fine attention, and (e-f) root mean squared (Frobenius) inter-head attention distance.} 
        \label{fig:mean and std of nets}
\end{figure}

We report layer-wise attention head statistics for the transformer encoder of each ROMA variant in Figure \ref{fig:mean and std of nets}. Despite having identical graph structure in the KM-3M and BD-3M experiments, the statistics of the attention mechanism differ greatly between trained models. In the BD-3M examples, the behavior of the attention mechanism is similar to ViT~\cite{dosovitskiy2020image}, with attention heads in the first layers attending more broadly before converging in later layers. In the KM-3M examples, not only is the diversity of attention heads much greater throughout all layers (Figure \ref{fig:mean and std of nets}e/f), but the utilization of the multiscale attention mechanism (Figure \ref{fig:mean and std of nets}c/d) is both greater overall and better distributed throughout layers. These differences potentially indicate that it is not merely the graph structure giving rise to multiscale interactions, but the interplay between the dynamics and graph structure. Namely, it is known that the Kuramoto model near the critical point gives rise to synchronization across scales, producing multiscale interactions beyond structure alone. The presence of these multiscale interactions may explain the much greater performance benefit of ROMA over baselines in the KM-3M experiments in Tables \ref{tab:scaling}/\ref{tab:effective} whereas only a small benefit is realized for the less pronounced multiscale interactions of the diffusive dynamics in the BD-3M experiments.\\

%\begin{center}
%\captionof{table}{Performance scaling of ROMA for KM-3M and BD-3M. }
%\label{tab:roma_powerlaws}
%\begin{tabular}{c|c|c|c}
%\toprule
%\textbf{Dataset} & \textbf{Loss} & $\alpha_N$ & $N_c$   \\ \hline 
%KM-3M   & Rel. $L^2_{}$             &  $\ \ 0.0222 \pm 0.0006$       & $ \ \ 2.46 \times 10^{13}  \pm 1.43\times 10^{13} \ \ \ $       \\  \hline 
%KM-3M   & RMSE$_{\textrm{PDE}}$     &    $\ \ 0.2496 \pm 0.1990  $   &   $ 2.98\times 10^9 \ \pm 4.82\times 10^{10} \ $      \\ \hline
%BD-3M   & Rel. $L^2_{}$             &  $ \ \ 0.0564 \pm 0.0069$       &   $6.31\times 10^8 \ \pm 1.58\times 10^9 \ \ $     \\ \hline
%BD-3M   & RMSE$_{\textrm{PDE}}$     &    $\ \ 1.4599 \pm 0.5674  $  &    $ 4.52\times 10^8  \  \pm 8.29\times 10^9 \ \ $    \\ \hline 
%\end{tabular}
%\end{center}
 ${}$ \\

%\vspace*{-12mm}

%\newpage
%\vspace*{-4mm}
\captionof{table}{Performance scaling of ROMA for KM-3M and BD-3M. }
\hspace*{-10mm} \nopagebreak
\begin{tabular}{lccc}
\toprule \vspace*{-3mm} \\
  & \multicolumn{2}{c}{ \textbf{KM-3M} }  \vspace*{0mm}  \\
 \cline{ 2-3 }  \multicolumn{1}{l}{ \textbf{Model} \ \ \ \ }  & \underline{Rel. $L^2_{}$} & \underline{RMSE$_{\textrm{PDE}}$} &  \textbf{\# Params}   \\ \midrule \vspace*{-3.5mm} \\ 

ROMA-B & $1.29 \!\times\! 10^{-3}$ & ${1.75 \!\times\! 10^{-5}}$ & 264 M   \\
%DON-MP & $1.61 \!\times\! 10^{-3}$ & - & $\mathbf{1.63 \!\times\! 10^{-4}}$ & -  \\
ROMA-L & $1.26 \!\times\! 10^{-3}$ & ${1.67 \!\times\! 10^{-5}}$ & 738 M  \\
%NOMAD-MP & $4.28 \!\times\! 10^{-3}$ & - & $1.37 \!\times\! 10^{-3}$ & -  \\
ROMA-H & $\mathbf{1.25 \!\times\! 10^{-3}}$ & $\mathbf{1.14 \!\times\! 10^{-5}}$ & 1.1 B   \\

\bottomrule

\end{tabular}
\quad
\begin{tabular}{lccc}

\toprule \vspace*{-3mm} \\
  & \multicolumn{2}{c}{ \textbf{BD-3M} }  \vspace*{0mm}  \\
 \cline{ 2-3 } \multicolumn{1}{l}{ \textbf{Model} \ \ \ \ } & \underline{Rel. $L^2_{}$} & \underline{RMSE$_{\textrm{PDE}}$} &  \textbf{\# Params}   \\ \midrule \vspace*{-3.5mm} \\ 

ROMA-B & $1.04 \!\times\! 10^{-2}$ & ${1.84 \!\times\! 10^{-8}}$ & 324 M   \\
%DON-MP & $1.61 \!\times\! 10^{-3}$ & - & $\mathbf{1.63 \!\times\! 10^{-4}}$ & -  \\
ROMA-L & ${9.75 \!\times\! 10^{-3}}$ & $\mathbf{2.34 \!\times\! 10^{-9}}$ & 896 M  \\
%NOMAD-MP & $4.28 \!\times\! 10^{-3}$ & - & $1.37 \!\times\! 10^{-3}$ & -  \\
ROMA-H & $\mathbf{9.64 \!\times\! 10^{-3}}$ & ${2.99 \!\times\! 10^{-9}}$ & 1.3 B   \\

\bottomrule

\end{tabular}

\vspace*{-12.mm} ${}^{}$ \\

\vspace*{-6mm}
In both experiments, performance in both the forecasting task and PDE task appears to scale directly with model size. Performance between tasks appears to remain coupled under architecture scaling, with the exception of ROMA-H in the BD-3M experiment having decreased PDE performance relative to ROMA-L.  To examine scaling relations more closely, we determine power laws for loss functions in the number of model parameters $P$~\cite{kaplan2020scaling,bahri2024explaining} \vspace*{-2mm}
\begin{equation}
    L(P) = \bigg( \frac{P_c}{P}\bigg)^{\alpha_P}
\end{equation}
for both the forecasting task and PDE task of each experiment in Table \ref{tab:roma_powerlaws}. \\
\begin{center}
\captionof{table}{Power law parameters for KM-3M and BD-3M: forecasting loss is  Rel. $L^2$ and PDE loss is RMSE$_{\textrm{PDE}}$. Additional power law fits from~\cite{CViT2024} are shown for incompressible/compressible Navier-Stokes (NS/CNS), Diffusion-Reaction (DR), and shallow-water equation (SWE) experiments.  }
\label{tab:roma_powerlaws}
\begin{tabular}{ccccc}

\textbf{Model} &\textbf{Dataset} & \textbf{Loss} & $\alpha_P$ & $P_c$   \\   \toprule
\multirow{ 2}{*}{ROMA}&\multirow{ 2}{*}{KM-3M} & Rel. $L^2_{}$     &  $\ \ 0.022 $       & $ \ \ 2.46 \times 10^{13}  $   \\ \cline{3-5} 
&     & RMSE$_{\textrm{PDE}}$     &    $\ \ 0.249 $   &   $ 2.98\times 10^9  $      \\ \cline{1-5} 
\multirow{ 2}{*}{ROMA} & \multirow{ 2}{*}{BD-3M}  & Rel. $L^2_{}$             &  $ \ \ 0.056 $       &   $6.31\times 10^8 $     \\ \cline{3-5}
&     & RMSE$_{\textrm{PDE}}$     &    $\ \ 1.459 $  &    $ 4.52\times 10^8 $    \\ \midrule \midrule
CViT&  NS   & Rel. $L^2_{}$     &  $\ \ 0.240 $       & $ \ \ 3.34 \times 10^{9}  $   \\ \cline{1-5} 
CViT&  CNS   & Rel. $L^2_{}$     &  $\ \ 0.379 $       & $ \ \ 1.81 \times 10^{8}  $   \\ \cline{1-5}
CViT&  DR   & Rel. $L^2_{}$     &  $\ \ 0.269 $       & $ \ \ 2.70 \times 10^{8}  $   \\ \cline{1-5}
CViT&  SW   & Rel. $L^2_{}$     &  $\ \ 0.535 $       & $ \ \ 2.04 \times 10^{8}  $   \\ \cline{1-5}
     
\end{tabular}
\end{center}
${}$ \\

In the examples shown in Table \ref{tab:roma_powerlaws}, power law exponents $\alpha_P$ span the entire range of the typical values in~\cite{bahri2024explaining}, $0 < \alpha < 1$. In particular, the exponent for the BD-3M experiments, $\alpha_P = 0.056$, is in the range of typical values for LLMs found in~\cite{kaplan2020scaling}, and moreover, the statistics of the attention heads in the BD-3M experiments (Figure \ref{fig:mean and std of nets}) are similar to those obtained by the ViT architecture~\cite{dosovitskiy2020image}.  In contrast, the forecasting loss exponent for KM-3M, $\alpha_P = 0.022$, is well below typical values for transformer models reported in~\cite{kaplan2020scaling} or~\cite{bahri2024explaining}, indicating weak scaling in model size. For both ROMA examples, the power law exponents for the PDE loss are an order of magnitude greater than their forecasting loss counterparts. \\

Both ROMA examples appear to scale in a different regime than CViT models, where all exponents fall within the resolution-limited scaling regime of random feature models in~\cite{bahri2024explaining}, $0.34 < \alpha_P < 1.31$. These differences potentially arise from the hierarchality of dynamics in each case, which are dictated by two main factors:  \vspace*{-2.2mm}\\

\begin{enumerate}[itemsep=1.3mm]
    \item[(1)]  topological heterogeneity;
    \item[(2)]  dynamical effects.
\end{enumerate}
${}$ \vspace*{-2.2mm}\\
The first factor, topological heterogeneity, divides the ROMA and CViT examples into different scaling regimes, separated by an order of magnitude. In particular, both the KM-3M and BD-3M examples have dynamics dictated largely by hierarchical graph structure (Table \ref{tab:dataset}), with further hierarchality arising in the KM-3M example due to synchronization at various scales. In the CViT experiments, the topology in each case is a 2D regular lattice, and any differences in hierarchality are due to dynamical effects. Potentially, the source of these dynamical effects can be explained in terms of graph measures derived from diffusion dynamics~\cite{reuveni2010anomalies,lacasa2013correlation,arnaudon2020scale,peach2019semi,peach2022relative}, which jointly quantify both (1) and (2). Broader implications of these results are: \vspace*{-1mm}\\
\begin{enumerate}[itemsep=1.4mm]
    \item In applications with a high degree of topological heterogeneity, model architecture has a much greater impact than model size.
    \item Effective dynamics scale much more strongly than forecasting accuracy, potentially enabling increased forecasting accuracy when dynamics and forecasting are sufficiently coupled (e.g., through multiscale interactions).
    \item Hierarchicality arising from dynamical effects can potentially be understood in terms of diffusive quantities such as the graph laplacian and characteristic length scale.
    \item In systems with unknown dynamics, power law scaling may differentiate systems based on their degree of hierarchality.
\end{enumerate}
\newpage

\subsection{Positional Embedding}
We investigate several positional embeddings for modeling multiscale interactions with the attention mechanism. Namely, we perform ablations on three positional embedding terms added to the function space prior to the self-attention blocks: 
\begin{equation}
\label{eq:pe}
    \hat{v}\big(\bu\big) = v\big(\bu\big) + \underbrace{\tilde{\mathbf{e}}(\bx)}_{\textrm{context}} + \underbrace{\lambda_s \, \mathbf{e}^{\sl}}_{\textrm{scale}} + \underbrace{\lambda^{\sl}\mathbf{e}_{i}}_{\textrm{index}}.   \vspace*{0mm} \\
\end{equation}
The first term incorporates context-specific information from the multiscale embeddings $\bx$ 
\begin{equation}
    \tilde{\mathbf{e}}(\bx) = \textrm{LN}(\textrm{MLP}(\bx)),
\end{equation}
where LN is a LayerNorm\cite{lei2016layer}, and the last two terms specify the position within the multiscale structure as trainable embeddings with two terms
\begin{equation}
        \underbrace{\bar{\mathbf{e}}}_{\textrm{multiscale}} = \underbrace{\lambda_s \, \mathbf{e}^{\sl}}_{\textrm{scale}} + \underbrace{\lambda^{\sl}\mathbf{e}_{i}}_{\textrm{index}}, 
\end{equation}
where embeddings are initialized as $\mathbf{e}^{\sl},\mathbf{e}_{i} \sim \mathcal{N}(0,1)$. Formally, the scale and index embeddings are biases independent of ${\bx}$.
\noindent We compare several parameterizations ranging from a standard index (ViT) positional encoding~\cite{vaswani2017attention} to full context and multiscale positional embeddings in Table \ref{tab:pe}:

\begin{enumerate}[topsep=9pt,itemsep=4pt, wide, labelwidth=!, labelindent=52pt]
    \item[(1)] \underline{index (ViT)}: \hspace*{12.4mm} $\lambda^{\sl}  = 1, \ \ \lambda_s = 0, \ \ {\tilde{\mathbf{e}}(\bx)} = 0$
    \item[(2)] \underline{multiscale}: \hspace*{15.4mm} $\lambda^{\sl}  = 1, \ \ \lambda_s = 1, \ \ {\tilde{\mathbf{e}}(\bx)} = 0$
    \item[(3)] \underline{context}: \hspace*{19.4mm} $\lambda^{\sl}  = 0, \ \ \lambda_s = 0, \ \ {\tilde{\mathbf{e}}(\bx)} \neq 0$
    \item[(4)] \underline{index/context}: \hspace*{0mm} \hspace*{8.4mm} $\lambda^{\sl}  = 1, \ \ \lambda_s = 0, \ \ {\tilde{\mathbf{e}}(\bx)} \neq 0$
    \item[(5)] \underline{scale/context}: \hspace*{0mm} \hspace*{9.6mm} $\lambda^{\sl}  = 0, \ \ \lambda_s = 1, \ \ {\tilde{\mathbf{e}}(\bx)} \neq 0$
    \item[(6)] \underline{multiscale/context}: \hspace*{0mm} \hspace*{1.6mm} $\lambda^{\sl}  = 1, \ \ \lambda_s = 1, \ \ {\tilde{\mathbf{e}}(\bx)} \neq 0$
\end{enumerate}
\vspace*{4mm} 

For both experiments in Table \ref{tab:pe}, the most effective parameterization incorporates a scale embedding, with differing methods of specifying node identity. In the BD-3M example, the best performance is achieved with a learnable multiscale embedding, which simply adds scale information to a ViT style index embedding. In the KM-3M example, specifying node identity based on context rather than index embedding achieves the best performance. Implications of these results are: (i) node identity can be specified with context or index embeddings, but using both may cause conflict, i.e., scale/context performs better than index/context and multiscale/context in both experiments; (ii) node scale is best specified as trainable embeddings, and cannot be fully replaced by context embeddings, i.e. scale/context performs better than context in both experiments; (iii) context embeddings have somewhat overlapping function in specifying both node identity and scale. In some instances this overlap is complementary but in others conflicting, and detailed experiments need to be conducted to determine the best positional embedding for a given application.

\vspace*{2mm}
\begin{minipage}{\textwidth}
   \hspace*{-8mm} 
  \begin{minipage}[b]{0.51\textwidth}
    
\captionof{table}{Several settings of the positional embeddings for the transformer encoder in the ROMA architecture, and corresponding forecast performance $\textrm{Rel. } L^2(\bar{\mathbf{u}}^{\sz},{\mathbf{u}}^{\sz})$. }
\label{tab:pe}
\vspace*{6mm}
\begin{tabular}{lcccc}

\toprule  %\multirow{2}{*}{  } & \multicolumn{4}{c}{ \textbf{Data Set}} \\
\multicolumn{1}{c}{ \textbf{Setting} \ \ \ \ }  & \multicolumn{2}{c}{ \textrm{KM-3M} } & \multicolumn{2}{c}{ \textrm{BD-3M} } \vspace*{0mm} \\
\midrule \vspace*{-3.5mm} \\
index (ViT) & \multicolumn{2}{c}{ ${1.76\!\times\!10^{-3}}$ } & \multicolumn{2}{c}{ $1.19\!\times\!10^{-2}$ }   \\
multiscale  & \multicolumn{2}{c}{ $1.79\!\times\!10^{-3}$ } & \multicolumn{2}{c}{ $\mathbf{1.08\!\times\!10^{-2}}$ }   \\
context  & \multicolumn{2}{c}{ $1.43\!\times\!10^{-3}$ } & \multicolumn{2}{c}{ ${1.29\!\times\!10^{-2}}$ }   \\
index/context  & \multicolumn{2}{c}{ $1.52\!\times\!10^{-3}$ } & \multicolumn{2}{c}{ ${1.61\!\times\!10^{-2}}$ }   \\
scale/context  & \multicolumn{2}{c}{ $\mathbf{1.41\!\times\!10^{-3}}$ } & \multicolumn{2}{c}{ ${1.23\!\times\!10^{-2}}$ }   \\
multiscale/context  & \multicolumn{2}{c}{ ${1.46\!\times\!10^{-3}}$ } & \multicolumn{2}{c}{ ${1.25\!\times\!10^{-2}}$ }   \\
%partial index$^\dagger$
%& \multicolumn{2}{c}{ $\mathbf{1.91\!\times\!10^{-6}}$ } & \multicolumn{2}{c}{ ${9.59\!\times\!10^{-4}}$ }    \\ 
\bottomrule \vspace*{-3mm}\\
%\multicolumn{3}{c}{$^{\dagger}$ best of $\alpha \in [1/4, \, 1/2 \, ,1]$ \ \ \ \ \  } \\
 & & & & \\
 & & & & \\
 & & & & \\
 & & & & \\
 & & & & \\
 & & & & \\
 & & & & \\
 & & & & \\
 & & & & \\

\label{tab:ablations}
\end{tabular}
  \end{minipage}
  \hfill
  \begin{minipage}[t]{0.49\textwidth}
    \centering
    \includegraphics[width=1.06\textwidth]{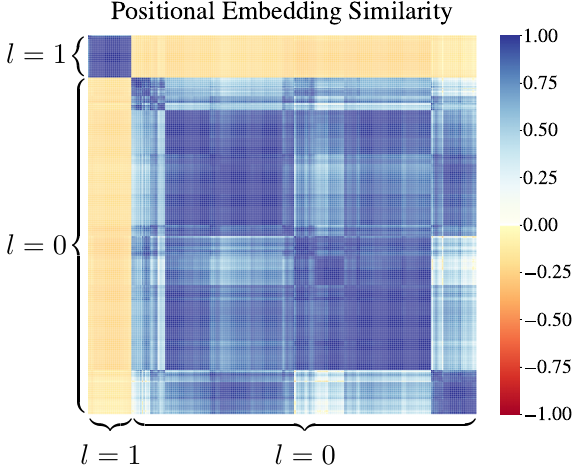}\hspace*{-8.5mm}
\captionof{figure}{Positional embedding similarity of scale/context embedding for KM-3M.} %Positional embeddings at levels $l=0$ and $l=1$ become anticorrelated while nodes within each scale become highly correlated or almost completely uncorrelated.}
\label{fig:pe}

\end{minipage}
\end{minipage}
\vspace*{-32mm}

\section{Discussion}

Encoding the dynamical regularities of complex systems as effective dynamics can potentially give insight into a number of out-of-equilibrium systems that have previously eluded compact dynamical description. To this end, we have demonstrated how the interplay between structure and dynamics of complex systems can play a central role in designing the next generation of neural operators for learning effective dynamics of large-scale complex systems. In particular, we emphasize that message passing is essential in learning coupled dynamics in a generalizable and scalable fashion, with learnable renormalization and multiscale interactions providing additional performance, interpretability, and physical realism in neural operators. \\

Additionally, we have demonstrated several specific approaches to leverage message passing architectures for learning multiscale interactions. The first is the renormalization precoder, which learns multiscale representations that condition both the transformer encoder and dynamical operators based on a local subgraph, enabling generalization to unseen subgraphs. The extensive experiments with positional embedding schemes show specifically how one can incorporate information from multiscale representations into the transformer architecture as context embeddings, enhancing physics-informed learning and forecasting performance. \\

Though we have illustrated several approaches for co-learning of structure and dynamics, presumably there are additional techniques that can further improve performance. Firstly, the reliance on MLPs as the base field architecture is likely suboptimal, with architectures such as KANs~\cite{liu2024kan} and LNOs~\cite{cao2024laplace} offering representation alternatives. In particular, LNO has been demonstrated to significantly improve performance on systems with aperiodic and transient dynamics, which are widely observed for complex systems. KANs with spectral representation of the activation functions can provide more compact and perhaps more explainable representations and can be used in the context of DeepOnet either in the trunk or in the branch or both ~\cite{shukla2024comprehensive}. \\

In our experiments, we use a single renormalization step for all ROMA models. This alone proved sufficient for learning effective dynamics and enhancing forecasting performance, but there may be further benefit from additional coarse-graining steps. However, the current formulation of ROMA based on DiffPool~\cite{ying2018hierarchical} uses a separate graph network for each coarse-graining step, the size of which is comparable to the transformer encoder and base fields. The current formulation is therefore not suitable for iterated coarse-graining, since scaling in the number of iterations is expected to be poor. One solution is to use a single coarse-graining network at all scales that is modulated by a scale-aware conditioning mechanism. Since network structure under renormalization is expected to be scale-invariant~\cite{boguna2021network}, using a single scale-aware coarse-graining network may not only be more parameter efficient, but also enforce scale invariance by a soft mechanism. \\

Lastly, we only explored two geometric models for the graph network precoder: Euclidean space and the Poincar\'e disk. While the Poincar\'e disk with constant negative curvature proved effective within the ROMA framework, it is known that curvature can vary across real world networks. More flexible models of negatively curved space may therefore be desirable for representing variations in local curvature over networks. One solution is to simply let curvature be a function of the local subgraph, i.e., as part of the precoder step. A second option is to let the manifold be a function of the local subgraph via Riemannian manifold learning~\cite{lebanon2012learning, hauberg2012geometric, arvanitidis2016locally,scarvelis2022riemannian,qiu2024estimating,diepeveen2024score}. In either approach, connecting the learned metrics at each scale through a Ricci flow~\cite{chow2004ricci,gracyk2024ricci} may be desirable as an inductive bias, improving scalability and physical plausibility. \\

\section{Methods}
\label{methods}
\subsection{Subgraph sampling}
Effective sampling of data in a batch-wise fashion is essential to training  AI models on large-scale complex systems. The additional graph structure obtained in subgraph sampling methods requires special consideration to achieve scalability, generalization, and efficiency of training. The most important considerations for subgraph sampling are controlling both the size and composition of computational neighborhoods in successive applications of graph operators in order to control variance, avoid  neighborhood explosion, and prevent oversmoothing~\cite{hamilton2017inductive,ying2018graph,zeng2019graphsaint}. This is achieved by systematically sampling the computational neighborhood of each node subsampling neighbors in each message passing layer \cite{hamilton2017inductive,ying2018graph}, or alternatively, by selecting a subgraph with suitable computational neighborhoods \cite{zeng2019graphsaint}.
We employ the latter approach and use the GraphSAINT random walk sampling algorithm as it is has a simple, modular implementation that can easily be composed with any graph operator. Additionally, the subgraphs sampled by the GraphSAINT algorithm are typically well connected, thereby capturing higher-order graph structures with high efficiency relative to the subgraph size. \\

In  learning effective dynamics of complex systems, we consider topologically and temporally localized samples comprising the sampled subgraph of fixed batch dimension $N_{B}$: $A^{\sz} = A^{\sz}_t$, and length $m$ dynamical histories at time $t$ for each of the $i$th nodes belonging to a  subgraph $A^{\sz}$
\begin{equation}
    \ub_{i}^{\sz} = \big[\ub_{i,t}^{\sz} \,, \ldots , \ub_{i,t-k\delta t}^{\sz} \,, \ldots  \big] \in \mathbb{R}^{1\times m}, \hspace{2cm} k = 1, \ldots, m.
\end{equation}
We denote the history of subgraph $A^{\sz}$ at time $t$ as
\begin{equation}
    \ub^{\sz} = \big[\ub_{0}^{\sz} \,, \ldots , \ub_{N_B}^{\sz}  \big] \in \mathbb{R}^{N_B\times m}
\end{equation}
where we suppress the node and temporal indices for notational convenience.

\subsection{Multiscale conditioning}

We develop a multiscale conditioning mechanism that enables in-context learning of multiscale structure. The multiscale conditioning mechanism learns local, in-context representations of sampled subgraphs and dynamical histories $( \mathbf{u}^{\sz},A^{\sz})$. These representations are subsequently used to compute coarse-graining operators and define a coupling mechanism between coarse-grained scales. \\

\subsubsection{Renormalization Precoder}

We define the neural renormalization procedure used in the precoder based on graph learning analogs of GR and LR. Specifically, we substitute components of GR and LR procedures for graph learning methods as follows:

\begin{enumerate}[topsep=7pt,itemsep=6pt,partopsep=7pt]
    \item[1.] Manifold embedding (GR)~\cite{krioukov2010hyperbolic,garcia2018multiscale,boguna2021network,allard2023geometric} $\rightarrow$ Hyperbolic graph networks with manifold $\mathcal{M}$~\cite{chami2019hyperbolic,liu2019hyperbolic}
    \item[2.] Coarse-graining (GR)~\cite{krioukov2010hyperbolic,garcia2018multiscale,boguna2021network,allard2023geometric} $\rightarrow$ Graph Pooling with manifold $\mathcal{M}$~\cite{ying2018hierarchical,lee2019self}
    \item[2.] Laplacian renormalization (LR)~\cite{villegas2023laplacian} $\rightarrow$ Laplacian eigenvector positional encoding~\cite{belkin2003,LapEig}
\end{enumerate}
We assume that the full graph structure is known in advance, and compute the top-$k$ Laplacian eigenvectors offline and for use as positional embeddings. We denote this positional embedding as 
\begin{equation}
    \mathbf{e}^{\sz} = \textrm{PE}(A^{\sz}) \in \mathbb{R}^{N_B\times n}
\end{equation}
which we append to each training history as 
\begin{equation}
    \hat{\ub}_{}^{\sz} = \big[\ub_{}^{\sz}; \mathbf{e}^{\sz} \big] \in \mathbb{R}^{N_B\times (m+n)}
\end{equation}
%We also consider node2vec\cite{node2vec} as both a complement and alternative to LE positional embedding in ablation experiments. 
The positionally encoded dynamical history is then mapped to a manifold $\mathcal{M}$ by a hyperbolic precoder
\begin{equation}
    \xb^{\sz} = \textrm{MPNN}^{\mathcal{M}}_{\textrm{pre}}\big(\hat{\ub}_{}^{\sz}, A^{\sz}\big) \in \mathcal{M}
\end{equation}
where we specifically consider the \Poincare  ball $\mathcal{M} = \mathbb{D}^d$, but one can also consider the isomorphic Hyperboloid model $\mathbb{H}^d$ for applications where large curvatures or learning rates are desirable, since these can produce instabilities while learning \Poincare embeddings~\cite{chami2019hyperbolic}.\\

We then compute the $l$th coarse-grained layer using a graph pooling~\cite{ying2018hierarchical} approach with separate networks for the coarse-graining matrix $S^{\sl}$ and coarse-grained embeddings:
\begin{align}
        S^{\sl} &= \textrm{MPNN}^{\mathcal{M};(l)}_{\textrm{CG}}\big(\xb^{\sl}, A^{\sl}\big) \\
        \zb^{\sl} &= \textrm{MPNN}^{\mathcal{M};\sl}_{\textrm{emb}}\big(\xb^{\sl}, A^{\sl}\big) 
\end{align}
where $\zb^{\sl}$ are concurrently computed embeddings that transform linearly according to $S^{\sl}$
\begin{align}
    A^{\slo} &= S^{\sl T} A^{\sl} S^{\sl} \\
    \xb^{\slo} &= S^{\sl T} \zb^{\sl} \\
    \ub^{\slo} &= S^{\sl T} \ub^{\sl} 
\end{align}
and the graph at layer $A^{\sl}$ transforms bilinearly into the next coarse-grained layer. In practice this procedure maps a graph of batch size $N_B$ to smaller graphs of predetermined size, e.g., $N_B\rightarrow N_1 \rightarrow \mydots \rightarrow N_L$, which imply assignment matrices
\begin{align}
    S^{(0)} &\in \mathbb{R}^{N_B \times N_1}, \\
    \ldots \notag \\
    S^{(L\text{-}1)} &\in \mathbb{R}^{N_{L\text{-}1} \times N_{L}}.
\end{align}
Each MPNN uses a common message passing rule specified in Appendix \ref{ROMA_MP}. \\

The multiscale batch is defined by concatenating the embeddings to obtain the composite feature vector
 \begin{equation}
 \begin{aligned}
          \bar{\mathbf{x}} &= \bigg[ \mathbf{x}^{\sz} \, , \, \mathbf{x}^{(1)} , \, \mydots \, , \mathbf{x}^{\sL} \bigg] \\
            &= \bigg[\mathbf{x}^{\sz} , \, S^{(0)}\mathbf{z}^{(0)} ,\, \mydots \, , \, S^{(L-1)}\mathbf{z}^{(L-1)} \bigg] \\
            %\bigg[ [ \mathbf{x}_{0}^{\sz}, \, \mydots \, ,\mathbf{x}_{N_B}^{\sz}]^T; \, \mydots \, ; \, [\mathbf{x}_{0}^{\sL}, \, \mydots \, ,\mathbf{x}_{N_L}^{\sL}]^T   \bigg] \\ 
          %& = 
 \end{aligned}
\end{equation}
and dynamical history
\begin{equation}
\begin{aligned}
        \bar{\mathbf{u}} &=  \bigg[ \mathbf{u}^{\sz} , \, \mathbf{u}^{(1)} , \, \mydots \, , \mathbf{u}^{\sL} \bigg] \\
            &= \bigg[\mathbf{u}^{\sz} , \, S^{(0)}\mathbf{u}^{(0)} ,\, \mydots \, , S^{(L-1)}\mathbf{u}^{(L-1)} \bigg] \\
            %\bigg[ [ \mathbf{s}_{0}^{\sz}, \, \mydots \, ,\mathbf{s}_{N_B}^{\sz}]^T; \, \mydots \, ; \, [\mathbf{s}_{0}^{\sL}, \, \mydots \, ,\mathbf{s}_{N_L}^{\sL}]^T   \bigg] \\ 
        %& =
\end{aligned}
\end{equation}
which produce multiscale features $\bar{\mathbf{x}} \in \mathbb{R}^{N_{B}' \times m}$ and $\bar{\mathbf{u}} \in \mathbb{R}^{N_{B}' \times m}$ with multiscale batch dimension $ N_{B}' = N_B + \mydots + N_L$. \\

\subsubsection{Transformer encoder} \label{transformer}
The self-attention mechanism \cite{bahdanau2014neural,vaswani2017attention,dosovitskiy2020image,LOCA2022,CViT2024} provides a direct way to exchange information between batched data using message passing, rather than indirectly through parameter updates. Here we discuss an approach to modeling multiscale interactions in neural operators using the self-attention mechanism. \\

For the attention mechanism to learn relations between nodes across scales, one must specify the identity of nodes that are considered distinct in the multiscale structure. This is typically accomplished by incorporating positional information prior to the attention blocks, since by default the attention mechanism is agnostic to the index of incoming nodes (formally, nodes are permutation invariant). For example, if we hypothesize that nodes at different scales ($l_1\neq l_2$) are not interchangeable, and supernodes within each coarse-grained scale ($l>0$) are not interchangeable, we must specify their identity with unique positional information. To incorporate each of these conditions we use an absolute positional embedding specifying the position within the multiscale batch
\begin{equation}
    \bar{\mathbf{e}} = \lambda_s \, \mathbf{e}^{\sl} + \lambda^{\sl}\mathbf{e}_{i}, \hspace{2cm} l = 1,\ldots,L, \ \ i = 1,\ldots, N_{B'}
\end{equation}
where  $\lambda_s$ and $\lambda^{\sl}$ are fixed parameters controlling the variance of the index and scale embeddings $\mathbf{e}^{\sl},\mathbf{e}_{i} \sim \mathcal{N}(0,1)$. \\

Additionally, we can inject context specific information from the multiscale embedding $\bar{\mathbf{x}}$ as
\begin{equation}
    \tilde{\mathbf{e}}(\bx) = \textrm{LN}(\textrm{MLP}(\bx)) 
\end{equation}
which we add to the dynamical history after projecting to a function space $v\in \mathcal{V}$ and prior to the self-attention blocks
\begin{equation}
       \hat{v}\big(\bu\big) = v\big(\bu\big) + \tilde{\mathbf{e}}(\bx) + \bar{\mathbf{e}} 
 \ \in \mathbb{R}^{N_{B}' \times m\times q }  \vspace*{4mm}
\end{equation} 
where $q$ is the number functions $v\in \mathcal{V}$ sampled in each forward pass. Alternatively, one could add positional embeddings of smaller dimension prior to the function space as 
\begin{equation}
\begin{aligned}
       \hat{v}\big(\bu\big) &= v\big(\bu + \tilde{\mathbf{e}}(\bx) + \bar{\mathbf{e}} \big)  \in \mathbb{R}^{N_{B}' \times m\times q } \\ \vspace*{2mm}
\end{aligned}
\end{equation}
which only requires embeddings of dimension $\tilde{\mathbf{e}}(\bx), \, \bar{\mathbf{e}}   \in \ \mathbb{R}^{N_{B}' \times m}$.
However, we find that positional embedding in the function space is much more effective, justifying the larger embeddings required by this approach. \\

With the identity and context of each node in the multiscale structure specified, we set $\bar{\mathbf{b}}_0 = \hat{v}\big(\bu\big)$ and learn coupled multiscale representations with pre-norm style transformer blocks \cite{dosovitskiy2020image,CViT2024}
\begin{equation}
    \begin{aligned}
        \bar{\mathbf{b}}_k' &= \bar{\mathbf{b}}_{k-1} + \textrm{MHA}\big(\textrm{LN}(\bar{\mathbf{b}}_{k-1}), \, \textrm{LN}(\bar{\mathbf{b}}_{k-1}), \, \textrm{LN}(\bar{\mathbf{b}}_{k-1}) \big)  \\
        \bar{\mathbf{b}}_k &= \bar{\mathbf{b}}_{k}' + \textrm{MLP}\big(\textrm{LN}(\bar{\mathbf{b}}_{k}') \big).
    \end{aligned}
\end{equation}
We denote the final layer as $\bar{\mathbf{b}}  \in \mathbb{R}^{N_B' \times d \times p}$, which serves as a multiscale conditioning mechanism in each of the message passing neural operators, and is analogous to the dimension $p$ basis computed by the branch network in the standard DeepONet architecture. 

\subsection{Operators}
We approach the problem of learning effective dynamics based on local subgraphs by learning locally coupled, emergent PDEs~\cite{Kemeth2022} of a latent field $\bar{\sigma}(t,\yb)$. This is achieved by first computing an operator $\mathcal{G}_{\textrm{P}}$ for the field at time $t + \delta t$, or propagator, as well as a separate operator $\mathcal{G}_{\textrm{D}}$ specifying the the effective dynamics of the propagator at $t + \delta t$. \\

\noindent The propagator can be written as
\begin{equation}
\begin{aligned}
        \bar{\sigma}(t,\yb) &= \mathcal{G}_{\textrm{P}}(v)\big(t,\yb;\bar{\mathbf{b}},\bar{\mathbf{x}}\big) \\
        &= \sum_{i=1}^p \bar{\mathbf{b}}_i \odot \bar{\mathbf{t}}_i  \in \mathbb{R}^{N'_{B} \times d}
\end{aligned}
\end{equation}
where $\odot$ is the Hadamard product. The trunk net takes as input encoded coordinates \begin{equation}
    (t,\mathbf{y}) = \mathbf{v} \mapsto \gamma(\mathbf{v}) = \begin{bmatrix}
        \textrm{cos}(\mathbf{B} \mathbf{v}) \\
        \textrm{sin}(\mathbf{B} \mathbf{v}) \\
    \end{bmatrix}
\end{equation}
where  $\mathbf{B} \in \mathbb{R}^{(1+d)\times m}$, and $\mathbf{B} \sim \mathcal{N}(0,\epsilon^2)$ where $\epsilon$ is a hyperparameter. This approach to coordinate representation  mitigates spectral bias~\cite{wang2023expert}, which is known to hinder physics-informed machine learning models~\cite{raissi2018deep,raissi2019physics}. The multiscale embedding $\bar{\mathbf{x}}$ can be concatenated as extra features~\cite{lu2022fair} to modulate the coordinate representation 
\begin{equation} \label{eq:coord}
    \bar{\mathbf{v}} = [\gamma(\mathbf{v}); \, \bar{\mathbf{x}}]
\end{equation}
which is used to compute the trunk net
\begin{equation}
    \bar{\mathbf{t}} = \textrm{MLP}\big(\bar{\mathbf{v}}\big) \in \mathbb{R}^{N_B' \times d \times p}.
\end{equation}
We set $\mathbf{y} = \mathbf{0}$ for each node and let $\bar{\mathbf{x}}$ modulate the spatial derivatives with respect to $\mathbf{y}$, rather than having the coordinates $\mathbf{y}$ encode both contextual representations and coordinates for spatial derivatives~\cite{Kemeth2022}. Importantly, this allows us to use a smaller dimension $d$ for the propagator field and its gradients $\nabla \bar{\sigma}$, $\nabla^2 \bar{\sigma}$, which are known to incur a significant computational cost due to successive applications of automatic differentiation (AD)~\cite{yu2022gradient}.

\subsubsection{Effective Dynamics}
The second operator $\mathcal{G}_{\textrm{D}}$ learns effective dynamics of the propagator at time $t+\delta t$, and takes as input the field $\bar{\sigma}$ as well as spatial derivatives with respect to the coordinate $\mathbf{y}$ ($\nabla \bar{\sigma}$, $\nabla^2 \bar{\sigma}, \ldots$). We consider two forms of effective dynamics: fully emergent, black-box dynamics where the right hand side is the operator
\begin{equation}
    \cfrac{\partial \bar{\sigma}}{\partial t} 
\simeq \mathcal{G}_{\textrm{D}}(v)\big(t, \mathbf{y};\bar{\mathbf{b}},\bar{\xb}, \bar{\sigma}, \nabla \bar{\sigma}, \nabla^2 \bar{\sigma} \big)
\end{equation}
and semi-supervised, gray-box dynamics of the form
\begin{equation}
    \cfrac{\partial \bar{\sigma}}{\partial t} 
\simeq \sum_{k=1}^m \mathcal{G}^k_{\textrm{D}}(v)\big(t, \mathbf{y};\bar{\mathbf{b}},\bar{\xb}\big) \mathbf{f}^k \big( \bar{\sigma}, \nabla \bar{\sigma}, \nabla^2 \bar{\sigma} \big)
\end{equation}
where each of the $m$ operators has a separate trunk net $\bar{\mathbf{t}}^k_i$
\begin{equation}
   \mathcal{G}^k_{\textrm{D}}(v)\big(t, \mathbf{y};\bar{\mathbf{b}},\bar{\xb}\big) = \sum_{i=1}^p \bar{\mathbf{b}}_i \odot \bar{\mathbf{t}}^k_i  \in \mathbb{R}^{N'_{B} \times d}.
\end{equation}
For example, we can write a Burgers equation with dynamic coefficients  
\begin{equation}
            \cfrac{\partial \bar{\sigma}}{\partial t} \simeq - \bar{F} \, (\bar{\sigma} \cdot \nabla) \bar{\sigma} + \bar{\nu} \, \nabla^2 \bar{\sigma} 
\end{equation}
where we have operator valued coefficients
\begin{equation}
    \begin{aligned}
         \mathcal{G}^0_{\textrm{D}}(v)\big(t, \mathbf{y};\bar{\mathbf{b}},\bar{\xb}\big) &= \bar{F} ,  \\
        \mathcal{G}^1_{\textrm{D}}(v)\big(t, \mathbf{y};\bar{\mathbf{b}},\bar{\xb}\big) &= \bar{\nu} ,  \\
    \end{aligned}
\end{equation}
and functional terms
\begin{equation}
    \begin{aligned}        
         \mathbf{f}^0\big( \bar{\sigma}, \nabla \bar{\sigma}, \nabla^2 \bar{\sigma} \big) &=(\bar{\sigma} \cdot \nabla) \bar{\sigma}, \\
         \mathbf{f}^1\big( \bar{\sigma}, \nabla \bar{\sigma}, \nabla^2 \bar{\sigma} \big) &= \nabla^2 \bar{\sigma} .  \\
    \end{aligned}
\end{equation}

%where  $\mathbf{B} \in \mathbb{R}^{(1+d)\times m}$, and $\mathbf{B} \sim \mathcal{N}(0,\epsilon^2)$ where $\epsilon$ is a hyperparameter. This approach to coordinate representation  mitigates spectral bias\cite{wang2023expert}, which is known to hinder physics-informed machine learning models\cite{raissi2018deep,raissi2019physics}.

\subsection{Projection}

The final operation is a \vspace*{1mm} projection $\mathcal{P}: \mathbb{R}^{N'_{B} \times d} \rightarrow \mathbb{R}^{ N'_{B}}$ that maps the propagator back to the original data space
\begin{equation}
    \bar{\mathbf{u}}_{\sitf} = \mathcal{P}\big(\bar{\sigma}\big) = \bigg[ \mathcal{P}\big({\sigma^{\sz}}\big), \, \mathcal{P}\big({S^{(0)}\sigma^{\sz}}\big), \, \mydots \, , \mathcal{P}\big({S^{(L-1)}\sigma^{(L-1)}}\big) \bigg] \in \mathbb{R}^{N'_B}
\end{equation}
which forecasts the subgraph at $t+\delta t$
\begin{equation}
    {\mathbf{u}}_{\sitf} = \bigg[ {\mathbf{u}^{\sz}_{\sitf}}, \, {S^{(0)}\mathbf{u}^{\sz}_{\sitf}}, \, \mydots \, ,{S^{(L-1)}\mathbf{u}^{(L-1)}_{\sitf}} \bigg]  \in \mathbb{R}^{N'_B}
\end{equation}
where we note that the bar notation $\bar{\mathbf{u}}_{\sitf}$ here implies a multiscale estimator of ${\mathbf{u}}_{\sitf}$.
Assuming the input-output data is normalized in the range $[0,1]$, we can apply a non-negative function such as a norm $\mathcal{P}(\cdot) = ||\cdot||_2$ or a learnable function such as an MLP with a sigmoid as the final activation.

\subsection{Loss}
We decompose the total loss function into two terms that encapsulate the dynamics and the multiscale geometric structure:
\begin{equation}
    \mathcal{L}_{\textrm{total}} = \mathcal{L}_{\textrm{dyn.}} + \mathcal{L}_{\textrm{MS}} .
\end{equation}
The dynamics loss is a weighted sum of the prediction error (data), effective dynamics (PDE), and gradient regularization of the effective dynamics (gPDE)
\begin{equation}
\begin{aligned}
    \mathcal{L}_{\textrm{dyn.}} &= w_{\textrm{data}}\,\mathcal{L}_{\textrm{data}} + w_{\textrm{PDE}}\,\mathcal{L}_{\textrm{PDE}} + w_{\textrm{gPDE}}\,\mathcal{L}_{\textrm{gPDE}} 
\end{aligned}
\end{equation}
where the data loss is the symmetric squared percentage error
\begin{equation}
\begin{aligned}
    \mathcal{L}_{\textrm{data}} &= \textrm{sMSPE}(\bar{\mathbf{u}},{\mathbf{u}}) = \frac{1}{N_B}\sum_{i}^{N_B} \bigg( \cfrac{|\bar{\mathbf{u}}_{i,\sitf} - {\mathbf{u}}_{i,\sitf}| }{|\bar{\mathbf{u}}_{i,\sitf}| + |{\mathbf{u}}_{i,\sitf}| + \epsilon } \, \bigg)^2 \\
\end{aligned}
\end{equation}
where $\epsilon$ controls the effective weights of errors at different scales\footnote{For example, a small $\epsilon=10^{-7}$ can prevent errors at very small values of $(\bar{\mathbf{u}},{\mathbf{u}})$ from being overrepresented.}. The PDE loss is the square of the residuals
\begin{equation}
\begin{aligned}
    \mathcal{L}_{\textrm{PDE}} &= \textrm{MSE}(\partial_t\bar{\sigma}, \, \mathcal{G}_{\textrm{D}}) = \frac{1}{N_B}\sum_{i}^{N_B} \bigg|\frac{\partial \bar{\sigma}}{\partial t} - \mathcal{G}_{\textrm{D}} \bigg|^2 = \frac{1}{N_B}\sum_{i}^{N_B} |{f}_i|^2 \\
\end{aligned}
\end{equation}
and we employ gradient regularization of the PDE residuals $f_i$ \cite{yu2022gradient}
\begin{equation} \label{eq:gpinn}
\begin{aligned}
    \mathcal{L}_{\textrm{gPDE}} &= \frac{1}{N_B} \sum_{i}^{N_B} \big| \nabla_{\mathbf{v}} f_i \big|  ^2  \\
\end{aligned}
\end{equation}
where $\mathbf{v} = (t,\mathbf{y})$. Gradient regularization is particularly beneficial for ill-posed inverse problems, and may therefore provide useful constraints for solving many-body systems. The cost of additional applications of AD required by \ref{eq:gpinn} is significant and scales as $\mathcal{O}(d^3)$, and hence the coordinate representation \ref{eq:coord} used in the trunk net serves to minimize this cost by minimizing the latent dimension $d$. \\

\noindent We employ a multiscale loss
\begin{equation}
\begin{aligned}
    \mathcal{L}_{\textrm{MS}} &= w_{\textrm{S}}\,\mathcal{L}_{{S}} + w_{\textrm{A}}\,\mathcal{L}_{{A}} + w_{\textrm{LP}}\,\mathcal{L}_{\textrm{LP}} \\
\end{aligned}
\end{equation}
where the first two terms index the convergence of the multiscale structure $S^{\sl}$ and $A^{\sl}$,
and the link prediction term measures reconstruction error of the initial graph structure based on embedding distances over the latent manifold $\mathcal{M}$. The multiscale convergence is measured by the entropies
\begin{equation}
\begin{aligned}
    \mathcal{L}_{{S}} &=  -\sum_{l,m,n} S_{mn}^{\sl} \textrm{log}(S_{mn}^{\sl}) \\  
    \mathcal{L}_{{A}} &=  -\sum_{l,m,n} A_{mn}^{\sl} \textrm{log}(A_{mn}^{\sl}) \\ 
\end{aligned}
\end{equation}
and the link prediction task is 
\begin{equation}
\begin{aligned}
    \mathcal{L}_{\textrm{LP}} &= \sum_{l,i,j} \big| A^{\sl}_{ij} - p_{ij}\big|^2, \\
    p_{ij} &= \cfrac{1}{ e^{(d_{\mathcal{M}}(\bar{\mathbf{x}}_{i}, \bar{\mathbf{x}}_{j}) - r)/t} + 1}, 
\end{aligned}
\end{equation}
${}$

where $p_{ij}$ follows a Fermi-Dirac distribution~\cite{krioukov2010hyperbolic,chami2019hyperbolic,nickel2017poincare} with learnable parameters $r$ and $t$. The link prediction task serves to regularize latent embeddings such that they preserve graph structure~\cite{chami2019hyperbolic}, and also preventing convergence to spurious minima early in training~\cite{ying2018hierarchical}. \\

The choice of weights for the multiscale loss will dictate how the ROMA mechanism is used. In particular, we examine two regimes: \\

\begin{enumerate}
    \item \underline{renormalized}: $w_S, w_A, w_{\textrm{LP}}\sim 1$, which converges to a near binary coarse-grained assignment and adjacency matrices.
    \item \underline{statistical}: $w_S, w_A, w_{\textrm{LP}}\ll 1$, which acts as a weight decay on the coarse-graining assignment, and in general will result in distributed assignment across multiple coarse-grained nodes, rather than binary assignment of each node to a single coarse-grained node.
\end{enumerate}
${}$ \\
We find that the former regime is more effective in the KM-3M example, while the latter is more effective in the BD-3M example. This is consistent with other results such as the power law scaling and attention head statistics for each example, where we see a much more pronounced multiscale structure in the best performing configuration for KM-3M than BD-3M. \\

Lastly, We note that coupling between individual terms in $\mathcal{L}_{\textrm{dyn.}}$ and $\mathcal{L}_{\textrm{MS}}$ is achieved through the attention mechanism of the branch net $\bar{\mathbf{b}}$ and the multiscale embedding $\bar{\mathbf{x}}$ in both operators 
\begin{equation}
    \begin{aligned}
        \bar{\sigma}(t,\yb) &= \mathcal{G}_{\textrm{P}}(v)\big(t,\yb;\bar{\mathbf{b}},\bar{\mathbf{x}}\big) \\
        \cfrac{\partial \bar{\sigma}}{\partial t} 
        &\simeq \mathcal{G}_{\textrm{D}}(v)\big(t, \mathbf{y};\bar{\mathbf{b}},\bar{\mathbf{x}}, \bar{\sigma}, \nabla \bar{\sigma}, \nabla^2 \bar{\sigma} \big)
    \end{aligned}
\end{equation}
facilitating positive transfer between tasks that may otherwise compete. 

\subsection*{Acknowledgments}
This material is based upon work supported by US Air Force Office of Scientific Research awards FA9550-20-1-0382 and FA9550-20-1-0383. N.F.J is additionally supported by The John Templeton Foundation.
\newpage

\bibliography{bibli}% common bib file


\begin{thebibliography}{106}
\ifx \bisbn   \undefined \def \bisbn  #1{ISBN #1}\fi
\ifx \binits  \undefined \def \binits#1{#1}\fi
\ifx \bauthor  \undefined \def \bauthor#1{#1}\fi
\ifx \batitle  \undefined \def \batitle#1{#1}\fi
\ifx \bjtitle  \undefined \def \bjtitle#1{#1}\fi
\ifx \bvolume  \undefined \def \bvolume#1{\textbf{#1}}\fi
\ifx \byear  \undefined \def \byear#1{#1}\fi
\ifx \bissue  \undefined \def \bissue#1{#1}\fi
\ifx \bfpage  \undefined \def \bfpage#1{#1}\fi
\ifx \blpage  \undefined \def \blpage #1{#1}\fi
\ifx \burl  \undefined \def \burl#1{\textsf{#1}}\fi
\ifx \doiurl  \undefined \def \doiurl#1{\url{https://doi.org/#1}}\fi
\ifx \betal  \undefined \def \betal{\textit{et al.}}\fi
\ifx \binstitute  \undefined \def \binstitute#1{#1}\fi
\ifx \binstitutionaled  \undefined \def \binstitutionaled#1{#1}\fi
\ifx \bctitle  \undefined \def \bctitle#1{#1}\fi
\ifx \beditor  \undefined \def \beditor#1{#1}\fi
\ifx \bpublisher  \undefined \def \bpublisher#1{#1}\fi
\ifx \bbtitle  \undefined \def \bbtitle#1{#1}\fi
\ifx \bedition  \undefined \def \bedition#1{#1}\fi
\ifx \bseriesno  \undefined \def \bseriesno#1{#1}\fi
\ifx \blocation  \undefined \def \blocation#1{#1}\fi
\ifx \bsertitle  \undefined \def \bsertitle#1{#1}\fi
\ifx \bsnm \undefined \def \bsnm#1{#1}\fi
\ifx \bsuffix \undefined \def \bsuffix#1{#1}\fi
\ifx \bparticle \undefined \def \bparticle#1{#1}\fi
\ifx \barticle \undefined \def \barticle#1{#1}\fi
\bibcommenthead
\ifx \bconfdate \undefined \def \bconfdate #1{#1}\fi
\ifx \botherref \undefined \def \botherref #1{#1}\fi
\ifx \url \undefined \def \url#1{\textsf{#1}}\fi
\ifx \bchapter \undefined \def \bchapter#1{#1}\fi
\ifx \bbook \undefined \def \bbook#1{#1}\fi
\ifx \bcomment \undefined \def \bcomment#1{#1}\fi
\ifx \oauthor \undefined \def \oauthor#1{#1}\fi
\ifx \citeauthoryear \undefined \def \citeauthoryear#1{#1}\fi
\ifx \endbibitem  \undefined \def \endbibitem {}\fi
\ifx \bconflocation  \undefined \def \bconflocation#1{#1}\fi
\ifx \arxivurl  \undefined \def \arxivurl#1{\textsf{#1}}\fi
\csname PreBibitemsHook\endcsname

%%% 1
\bibitem[\protect\citeauthoryear{Kemeth et~al.}{2022}]{Kemeth2022}
\begin{barticle}
\bauthor{\bsnm{Kemeth}, \binits{F.P.}},
\bauthor{\bsnm{Bertalan}, \binits{T.}},
\bauthor{\bsnm{Thiem}, \binits{T.}},
\bauthor{\bsnm{Dietrich}, \binits{F.}},
\bauthor{\bsnm{Moon}, \binits{S.J.}},
\bauthor{\bsnm{Laing}, \binits{C.R.}},
\bauthor{\bsnm{Kevrekidis}, \binits{I.G.}}:
\batitle{Learning emergent partial differential equations in a learned emergent space}.
\bjtitle{Nature Communications}
\bvolume{13}(\bissue{1}),
\bfpage{3318}
(\byear{2022})
\end{barticle}
\endbibitem

%%% 2
\bibitem[\protect\citeauthoryear{De~Florio et~al.}{2024}]{de2024ai}
\begin{barticle}
\bauthor{\bsnm{De~Florio}, \binits{M.}},
\bauthor{\bsnm{Kevrekidis}, \binits{I.G.}},
\bauthor{\bsnm{Karniadakis}, \binits{G.E.}}:
\batitle{Ai-lorenz: A physics-data-driven framework for black-box and gray-box identification of chaotic systems with symbolic regression}.
\bjtitle{Chaos, Solitons \& Fractals}
\bvolume{188},
\bfpage{115538}
(\byear{2024})
\end{barticle}
\endbibitem

%%% 3
\bibitem[\protect\citeauthoryear{Vlachas et~al.}{2022}]{vlachas2022multiscale}
\begin{barticle}
\bauthor{\bsnm{Vlachas}, \binits{P.R.}},
\bauthor{\bsnm{Arampatzis}, \binits{G.}},
\bauthor{\bsnm{Uhler}, \binits{C.}},
\bauthor{\bsnm{Koumoutsakos}, \binits{P.}}:
\batitle{Multiscale simulations of complex systems by learning their effective dynamics}.
\bjtitle{Nature Machine Intelligence}
\bvolume{4}(\bissue{4}),
\bfpage{359}--\blpage{366}
(\byear{2022})
\end{barticle}
\endbibitem

%%% 4
\bibitem[\protect\citeauthoryear{Kaltenbach and Koutsourelakis}{2020}]{kaltenbach2020incorporating}
\begin{barticle}
\bauthor{\bsnm{Kaltenbach}, \binits{S.}},
\bauthor{\bsnm{Koutsourelakis}, \binits{P.-S.}}:
\batitle{Incorporating physical constraints in a deep probabilistic machine learning framework for coarse-graining dynamical systems}.
\bjtitle{Journal of Computational Physics}
\bvolume{419},
\bfpage{109673}
(\byear{2020})
\end{barticle}
\endbibitem

%%% 5
\bibitem[\protect\citeauthoryear{Mao et~al.}{2019}]{mao2019nonlocal}
\begin{barticle}
\bauthor{\bsnm{Mao}, \binits{Z.}},
\bauthor{\bsnm{Li}, \binits{Z.}},
\bauthor{\bsnm{Karniadakis}, \binits{G.E.}}:
\batitle{Nonlocal flocking dynamics: learning the fractional order of pdes from particle simulations}.
\bjtitle{Communications on Applied Mathematics and Computation}
\bvolume{1},
\bfpage{597}--\blpage{619}
(\byear{2019})
\end{barticle}
\endbibitem

%%% 6
\bibitem[\protect\citeauthoryear{Karniadakis et~al.}{2021}]{karniadakis2021physics}
\begin{barticle}
\bauthor{\bsnm{Karniadakis}, \binits{G.E.}},
\bauthor{\bsnm{Kevrekidis}, \binits{I.G.}},
\bauthor{\bsnm{Lu}, \binits{L.}},
\bauthor{\bsnm{Perdikaris}, \binits{P.}},
\bauthor{\bsnm{Wang}, \binits{S.}},
\bauthor{\bsnm{Yang}, \binits{L.}}:
\batitle{Physics-informed machine learning}.
\bjtitle{Nature Reviews Physics}
\bvolume{3}(\bissue{6}),
\bfpage{422}--\blpage{440}
(\byear{2021})
\end{barticle}
\endbibitem

%%% 7
\bibitem[\protect\citeauthoryear{Lu et~al.}{2021}]{lu2021learning}
\begin{barticle}
\bauthor{\bsnm{Lu}, \binits{L.}},
\bauthor{\bsnm{Jin}, \binits{P.}},
\bauthor{\bsnm{Pang}, \binits{G.}},
\bauthor{\bsnm{Zhang}, \binits{Z.}},
\bauthor{\bsnm{Karniadakis}, \binits{G.E.}}:
\batitle{Learning nonlinear operators via deeponet based on the universal approximation theorem of operators}.
\bjtitle{Nature machine intelligence}
\bvolume{3}(\bissue{3}),
\bfpage{218}--\blpage{229}
(\byear{2021})
\end{barticle}
\endbibitem

%%% 8
\bibitem[\protect\citeauthoryear{Wang et~al.}{2022}]{wang2022improved}
\begin{barticle}
\bauthor{\bsnm{Wang}, \binits{S.}},
\bauthor{\bsnm{Wang}, \binits{H.}},
\bauthor{\bsnm{Perdikaris}, \binits{P.}}:
\batitle{Improved architectures and training algorithms for deep operator networks}.
\bjtitle{Journal of Scientific Computing}
\bvolume{92}(\bissue{2}),
\bfpage{35}
(\byear{2022})
\end{barticle}
\endbibitem

%%% 9
\bibitem[\protect\citeauthoryear{Kovachki et~al.}{2023}]{kovachki2023neural}
\begin{barticle}
\bauthor{\bsnm{Kovachki}, \binits{N.}},
\bauthor{\bsnm{Li}, \binits{Z.}},
\bauthor{\bsnm{Liu}, \binits{B.}},
\bauthor{\bsnm{Azizzadenesheli}, \binits{K.}},
\bauthor{\bsnm{Bhattacharya}, \binits{K.}},
\bauthor{\bsnm{Stuart}, \binits{A.}},
\bauthor{\bsnm{Anandkumar}, \binits{A.}}:
\batitle{Neural operator: Learning maps between function spaces with applications to pdes}.
\bjtitle{Journal of Machine Learning Research}
\bvolume{24}(\bissue{89}),
\bfpage{1}--\blpage{97}
(\byear{2023})
\end{barticle}
\endbibitem

%%% 10
\bibitem[\protect\citeauthoryear{Cao et~al.}{2024}]{cao2024laplace}
\begin{barticle}
\bauthor{\bsnm{Cao}, \binits{Q.}},
\bauthor{\bsnm{Goswami}, \binits{S.}},
\bauthor{\bsnm{Karniadakis}, \binits{G.E.}}:
\batitle{Laplace neural operator for solving differential equations}.
\bjtitle{Nature Machine Intelligence}
\bvolume{6}(\bissue{6}),
\bfpage{631}--\blpage{640}
(\byear{2024})
\end{barticle}
\endbibitem

%%% 11
\bibitem[\protect\citeauthoryear{Betzel and Bassett}{2017}]{betzel2017multi}
\begin{barticle}
\bauthor{\bsnm{Betzel}, \binits{R.F.}},
\bauthor{\bsnm{Bassett}, \binits{D.S.}}:
\batitle{Multi-scale brain networks}.
\bjtitle{NeuroImage}
\bvolume{160},
\bfpage{73}--\blpage{83}
(\byear{2017}).
\bcomment{Functional Architecture of the Brain}
\end{barticle}
\endbibitem

%%% 12
\bibitem[\protect\citeauthoryear{Garc{\'\i}a-P{\'e}rez et~al.}{2018}]{garcia2018multiscale}
\begin{barticle}
\bauthor{\bsnm{Garc{\'\i}a-P{\'e}rez}, \binits{G.}},
\bauthor{\bsnm{Bogu{\~n}{\'a}}, \binits{M.}},
\bauthor{\bsnm{Serrano}, \binits{M.{\'A}.}}:
\batitle{Multiscale unfolding of real networks by geometric renormalization}.
\bjtitle{Nature Physics}
\bvolume{14}(\bissue{6}),
\bfpage{583}--\blpage{589}
(\byear{2018})
\end{barticle}
\endbibitem

%%% 13
\bibitem[\protect\citeauthoryear{Garuccio et~al.}{2023}]{garuccio2023multiscale}
\begin{barticle}
\bauthor{\bsnm{Garuccio}, \binits{E.}},
\bauthor{\bsnm{Lalli}, \binits{M.}},
\bauthor{\bsnm{Garlaschelli}, \binits{D.}}:
\batitle{Multiscale network renormalization: scale-invariance without geometry}.
\bjtitle{Physical Review Research}
\bvolume{5}(\bissue{4}),
\bfpage{043101}
(\byear{2023})
\end{barticle}
\endbibitem

%%% 14
\bibitem[\protect\citeauthoryear{Chami et~al.}{2022}]{chami2022machine}
\begin{barticle}
\bauthor{\bsnm{Chami}, \binits{I.}},
\bauthor{\bsnm{Abu-El-Haija}, \binits{S.}},
\bauthor{\bsnm{Perozzi}, \binits{B.}},
\bauthor{\bsnm{R{\'e}}, \binits{C.}},
\bauthor{\bsnm{Murphy}, \binits{K.}}:
\batitle{Machine learning on graphs: A model and comprehensive taxonomy}.
\bjtitle{Journal of Machine Learning Research}
\bvolume{23}(\bissue{89}),
\bfpage{1}--\blpage{64}
(\byear{2022})
\end{barticle}
\endbibitem

%%% 15
\bibitem[\protect\citeauthoryear{Grover and Leskovec}{2016}]{node2vec}
\begin{botherref}
\oauthor{\bsnm{Grover}, \binits{A.}},
\oauthor{\bsnm{Leskovec}, \binits{J.}}:
node2vec: Scalable feature learning for networks.
CoRR
\textbf{abs/1607.00653}
(2016)
\end{botherref}
\endbibitem

%%% 16
\bibitem[\protect\citeauthoryear{Xu}{2021}]{xu2021understanding}
\begin{barticle}
\bauthor{\bsnm{Xu}, \binits{M.}}:
\batitle{Understanding graph embedding methods and their applications}.
\bjtitle{SIAM Review}
\bvolume{63}(\bissue{4}),
\bfpage{825}--\blpage{853}
(\byear{2021})
\end{barticle}
\endbibitem

%%% 17
\bibitem[\protect\citeauthoryear{Kipf and Welling}{2016}]{kipf2016semi}
\begin{botherref}
\oauthor{\bsnm{Kipf}, \binits{T.N.}},
\oauthor{\bsnm{Welling}, \binits{M.}}:
Semi-supervised classification with graph convolutional networks.
arXiv preprint arXiv:1609.02907
(2016)
\end{botherref}
\endbibitem

%%% 18
\bibitem[\protect\citeauthoryear{Veličković et~al.}{2018}]{GAT2018}
\begin{bchapter}
\bauthor{\bsnm{Veličković}, \binits{P.}},
\bauthor{\bsnm{Cucurull}, \binits{G.}},
\bauthor{\bsnm{Casanova}, \binits{A.}},
\bauthor{\bsnm{Romero}, \binits{A.}},
\bauthor{\bsnm{Liò}, \binits{P.}},
\bauthor{\bsnm{Bengio}, \binits{Y.}}:
\bctitle{Graph attention networks}.
In: \bbtitle{International Conference on Learning Representations}
(\byear{2018})
\end{bchapter}
\endbibitem

%%% 19
\bibitem[\protect\citeauthoryear{Gilmer et~al.}{2017}]{MPNN2017}
\begin{barticle}
\bauthor{\bsnm{Gilmer}, \binits{J.}},
\bauthor{\bsnm{Schoenholz}, \binits{S.S.}},
\bauthor{\bsnm{Riley}, \binits{P.F.}},
\bauthor{\bsnm{Vinyals}, \binits{O.}},
\bauthor{\bsnm{Dahl}, \binits{G.E.}}:
\batitle{Neural message passing for quantum chemistry}.
\bjtitle{Proceedings of Machine Learning Research}
\bvolume{70},
\bfpage{1263}--\blpage{1272}
(\byear{2017})
\end{barticle}
\endbibitem

%%% 20
\bibitem[\protect\citeauthoryear{You et~al.}{2022}]{you2022nonlocal}
\begin{barticle}
\bauthor{\bsnm{You}, \binits{H.}},
\bauthor{\bsnm{Yu}, \binits{Y.}},
\bauthor{\bsnm{D'Elia}, \binits{M.}},
\bauthor{\bsnm{Gao}, \binits{T.}},
\bauthor{\bsnm{Silling}, \binits{S.}}:
\batitle{Nonlocal kernel network (nkn): A stable and resolution-independent deep neural network}.
\bjtitle{Journal of Computational Physics}
\bvolume{469},
\bfpage{111536}
(\byear{2022})
\end{barticle}
\endbibitem

%%% 21
\bibitem[\protect\citeauthoryear{Hamilton et~al.}{2017}]{hamilton2017inductive}
\begin{botherref}
\oauthor{\bsnm{Hamilton}, \binits{W.}},
\oauthor{\bsnm{Ying}, \binits{Z.}},
\oauthor{\bsnm{Leskovec}, \binits{J.}}:
Inductive representation learning on large graphs.
Advances in neural information processing systems
\textbf{30}
(2017)
\end{botherref}
\endbibitem

%%% 22
\bibitem[\protect\citeauthoryear{Jin et~al.}{2020}]{jin2020sympnets}
\begin{barticle}
\bauthor{\bsnm{Jin}, \binits{P.}},
\bauthor{\bsnm{Zhang}, \binits{Z.}},
\bauthor{\bsnm{Zhu}, \binits{A.}},
\bauthor{\bsnm{Tang}, \binits{Y.}},
\bauthor{\bsnm{Karniadakis}, \binits{G.E.}}:
\batitle{Sympnets: Intrinsic structure-preserving symplectic networks for identifying hamiltonian systems}.
\bjtitle{Neural Networks}
\bvolume{132},
\bfpage{166}--\blpage{179}
(\byear{2020})
\end{barticle}
\endbibitem

%%% 23
\bibitem[\protect\citeauthoryear{Li et~al.}{2020}]{li2020neural}
\begin{botherref}
\oauthor{\bsnm{Li}, \binits{Z.}},
\oauthor{\bsnm{Kovachki}, \binits{N.}},
\oauthor{\bsnm{Azizzadenesheli}, \binits{K.}},
\oauthor{\bsnm{Liu}, \binits{B.}},
\oauthor{\bsnm{Bhattacharya}, \binits{K.}},
\oauthor{\bsnm{Stuart}, \binits{A.}},
\oauthor{\bsnm{Anandkumar}, \binits{A.}}:
Neural operator: Graph kernel network for partial differential equations.
arXiv preprint arXiv:2003.03485
(2020)
\end{botherref}
\endbibitem

%%% 24
\bibitem[\protect\citeauthoryear{Wang et~al.}{2024}]{CViT2024}
\begin{botherref}
\oauthor{\bsnm{Wang}, \binits{S.}},
\oauthor{\bsnm{Seidman}, \binits{J.H.}},
\oauthor{\bsnm{Sankaran}, \binits{S.}},
\oauthor{\bsnm{Wang}, \binits{H.}},
\oauthor{\bsnm{Pappas}, \binits{G.J.}},
\oauthor{\bsnm{Perdikaris}, \binits{P.}}:
Bridging operator learning and conditioned neural fields: A unifying perspective.
arXiv preprint arXiv:2405.13998
(2024)
\end{botherref}
\endbibitem

%%% 25
\bibitem[\protect\citeauthoryear{Bryutkin et~al.}{2024}]{bryutkin2024hamlet}
\begin{botherref}
\oauthor{\bsnm{Bryutkin}, \binits{A.}},
\oauthor{\bsnm{Huang}, \binits{J.}},
\oauthor{\bsnm{Deng}, \binits{Z.}},
\oauthor{\bsnm{Yang}, \binits{G.}},
\oauthor{\bsnm{Sch{\"o}nlieb}, \binits{C.-B.}},
\oauthor{\bsnm{Aviles-Rivero}, \binits{A.}}:
Hamlet: Graph transformer neural operator for partial differential equations.
arXiv preprint arXiv:2402.03541
(2024)
\end{botherref}
\endbibitem

%%% 26
\bibitem[\protect\citeauthoryear{Ganea et~al.}{2018}]{ganea2018hyperbolic}
\begin{botherref}
\oauthor{\bsnm{Ganea}, \binits{O.}},
\oauthor{\bsnm{B{\'e}cigneul}, \binits{G.}},
\oauthor{\bsnm{Hofmann}, \binits{T.}}:
Hyperbolic neural networks.
Advances in neural information processing systems
\textbf{31}
(2018)
\end{botherref}
\endbibitem

%%% 27
\bibitem[\protect\citeauthoryear{Liu et~al.}{2019}]{liu2019hyperbolic}
\begin{botherref}
\oauthor{\bsnm{Liu}, \binits{Q.}},
\oauthor{\bsnm{Nickel}, \binits{M.}},
\oauthor{\bsnm{Kiela}, \binits{D.}}:
Hyperbolic graph neural networks.
Advances in neural information processing systems
\textbf{32}
(2019)
\end{botherref}
\endbibitem

%%% 28
\bibitem[\protect\citeauthoryear{Chami et~al.}{2019}]{chami2019hyperbolic}
\begin{botherref}
\oauthor{\bsnm{Chami}, \binits{I.}},
\oauthor{\bsnm{Ying}, \binits{Z.}},
\oauthor{\bsnm{R{\'e}}, \binits{C.}},
\oauthor{\bsnm{Leskovec}, \binits{J.}}:
Hyperbolic graph convolutional neural networks.
Advances in neural information processing systems
\textbf{32}
(2019)
\end{botherref}
\endbibitem

%%% 29
\bibitem[\protect\citeauthoryear{Cao}{2021}]{cao2021choose}
\begin{barticle}
\bauthor{\bsnm{Cao}, \binits{S.}}:
\batitle{Choose a transformer: Fourier or galerkin}.
\bjtitle{Advances in neural information processing systems}
\bvolume{34},
\bfpage{24924}--\blpage{24940}
(\byear{2021})
\end{barticle}
\endbibitem

%%% 30
\bibitem[\protect\citeauthoryear{Gell-Mann and Low}{1954}]{gell1954quantum}
\begin{barticle}
\bauthor{\bsnm{Gell-Mann}, \binits{M.}},
\bauthor{\bsnm{Low}, \binits{F.E.}}:
\batitle{Quantum electrodynamics at small distances}.
\bjtitle{Physical Review}
\bvolume{95}(\bissue{5}),
\bfpage{1300}
(\byear{1954})
\end{barticle}
\endbibitem

%%% 31
\bibitem[\protect\citeauthoryear{Wilson}{1971}]{wilson1971renormalization}
\begin{barticle}
\bauthor{\bsnm{Wilson}, \binits{K.G.}}:
\batitle{Renormalization group and strong interactions}.
\bjtitle{Physical Review D}
\bvolume{3}(\bissue{8}),
\bfpage{1818}
(\byear{1971})
\end{barticle}
\endbibitem

%%% 32
\bibitem[\protect\citeauthoryear{Pelissetto and Vicari}{2002}]{pelissetto2002critical}
\begin{barticle}
\bauthor{\bsnm{Pelissetto}, \binits{A.}},
\bauthor{\bsnm{Vicari}, \binits{E.}}:
\batitle{Critical phenomena and renormalization-group theory}.
\bjtitle{Physics Reports}
\bvolume{368}(\bissue{6}),
\bfpage{549}--\blpage{727}
(\byear{2002})
\end{barticle}
\endbibitem

%%% 33
\bibitem[\protect\citeauthoryear{Verstraete et~al.}{2023}]{verstraete2023density}
\begin{barticle}
\bauthor{\bsnm{Verstraete}, \binits{F.}},
\bauthor{\bsnm{Nishino}, \binits{T.}},
\bauthor{\bsnm{Schollw{\"o}ck}, \binits{U.}},
\bauthor{\bsnm{Ba{\~n}uls}, \binits{M.C.}},
\bauthor{\bsnm{Chan}, \binits{G.K.}},
\bauthor{\bsnm{Stoudenmire}, \binits{M.E.}}:
\batitle{Density matrix renormalization group, 30 years on}.
\bjtitle{Nature Reviews Physics}
\bvolume{5}(\bissue{5}),
\bfpage{273}--\blpage{276}
(\byear{2023})
\end{barticle}
\endbibitem

%%% 34
\bibitem[\protect\citeauthoryear{Kraichnan}{1982}]{kraichnan1982hydrodynamic}
\begin{barticle}
\bauthor{\bsnm{Kraichnan}, \binits{R.H.}}:
\batitle{Hydrodynamic turbulence and the renormalization group}.
\bjtitle{Physical Review A}
\bvolume{25}(\bissue{6}),
\bfpage{3281}
(\byear{1982})
\end{barticle}
\endbibitem

%%% 35
\bibitem[\protect\citeauthoryear{Yakhot and Orszag}{1986}]{yakhot1986renormalization}
\begin{barticle}
\bauthor{\bsnm{Yakhot}, \binits{V.}},
\bauthor{\bsnm{Orszag}, \binits{S.A.}}:
\batitle{Renormalization group analysis of turbulence. i. basic theory}.
\bjtitle{Journal of scientific computing}
\bvolume{1}(\bissue{1}),
\bfpage{3}--\blpage{51}
(\byear{1986})
\end{barticle}
\endbibitem

%%% 36
\bibitem[\protect\citeauthoryear{Polyakov}{1993}]{polyakov1993theory}
\begin{barticle}
\bauthor{\bsnm{Polyakov}, \binits{A.M.}}:
\batitle{The theory of turbulence in two dimensions}.
\bjtitle{Nuclear Physics B}
\bvolume{396}(\bissue{2-3}),
\bfpage{367}--\blpage{385}
(\byear{1993})
\end{barticle}
\endbibitem

%%% 37
\bibitem[\protect\citeauthoryear{Zhou}{2010}]{zhou2010renormalization}
\begin{barticle}
\bauthor{\bsnm{Zhou}, \binits{Y.}}:
\batitle{Renormalization group theory for fluid and plasma turbulence}.
\bjtitle{Physics Reports}
\bvolume{488}(\bissue{1}),
\bfpage{1}--\blpage{49}
(\byear{2010})
\end{barticle}
\endbibitem

%%% 38
\bibitem[\protect\citeauthoryear{Krioukov et~al.}{2010}]{krioukov2010hyperbolic}
\begin{barticle}
\bauthor{\bsnm{Krioukov}, \binits{D.}},
\bauthor{\bsnm{Papadopoulos}, \binits{F.}},
\bauthor{\bsnm{Kitsak}, \binits{M.}},
\bauthor{\bsnm{Vahdat}, \binits{A.}},
\bauthor{\bsnm{Bogu\~n\'a}, \binits{M.}}:
\batitle{Hyperbolic geometry of complex networks}.
\bjtitle{Phys. Rev. E}
\bvolume{82},
\bfpage{036106}
(\byear{2010})
\doiurl{10.1103/PhysRevE.82.036106}
\end{barticle}
\endbibitem

%%% 39
\bibitem[\protect\citeauthoryear{Boguna et~al.}{2021}]{boguna2021network}
\begin{barticle}
\bauthor{\bsnm{Boguna}, \binits{M.}},
\bauthor{\bsnm{Bonamassa}, \binits{I.}},
\bauthor{\bsnm{De~Domenico}, \binits{M.}},
\bauthor{\bsnm{Havlin}, \binits{S.}},
\bauthor{\bsnm{Krioukov}, \binits{D.}},
\bauthor{\bsnm{Serrano}, \binits{M.{\'A}.}}:
\batitle{Network geometry}.
\bjtitle{Nature Reviews Physics}
\bvolume{3}(\bissue{2}),
\bfpage{114}--\blpage{135}
(\byear{2021})
\end{barticle}
\endbibitem

%%% 40
\bibitem[\protect\citeauthoryear{Allard et~al.}{2023}]{allard2023geometric}
\begin{botherref}
\oauthor{\bsnm{Allard}, \binits{A.}},
\oauthor{\bsnm{Serrano}, \binits{M.{\'A}.}},
\oauthor{\bsnm{Bogu{\~n}{\'a}}, \binits{M.}}:
Geometric description of clustering in directed networks.
Nature Physics,
1--7
(2023)
\end{botherref}
\endbibitem

%%% 41
\bibitem[\protect\citeauthoryear{Villegas et~al.}{2023}]{villegas2023laplacian}
\begin{barticle}
\bauthor{\bsnm{Villegas}, \binits{P.}},
\bauthor{\bsnm{Gili}, \binits{T.}},
\bauthor{\bsnm{Caldarelli}, \binits{G.}},
\bauthor{\bsnm{Gabrielli}, \binits{A.}}:
\batitle{Laplacian renormalization group for heterogeneous networks}.
\bjtitle{Nature Physics}
\bvolume{19}(\bissue{3}),
\bfpage{445}--\blpage{450}
(\byear{2023})
\end{barticle}
\endbibitem

%%% 42
\bibitem[\protect\citeauthoryear{Caldarelli et~al.}{2024}]{caldarelli2024laplacian}
\begin{botherref}
\oauthor{\bsnm{Caldarelli}, \binits{G.}},
\oauthor{\bsnm{Gabrielli}, \binits{A.}},
\oauthor{\bsnm{Gili}, \binits{T.}},
\oauthor{\bsnm{Villegas}, \binits{P.}}:
Laplacian renormalization group: an introduction to heterogeneous coarse-graining.
Journal of Statistical Mechanics: Theory and Experiment
\textbf{2024}(8)
(2024)
\end{botherref}
\endbibitem

%%% 43
\bibitem[\protect\citeauthoryear{Kaiser and Hilgetag}{2010}]{kaiser2010optimal}
\begin{barticle}
\bauthor{\bsnm{Kaiser}, \binits{M.}},
\bauthor{\bsnm{Hilgetag}, \binits{C.C.}}:
\batitle{Optimal hierarchical modular topologies for producing limited sustained activation of neural networks}.
\bjtitle{Frontiers in neuroinformatics}
\bvolume{4},
\bfpage{713}
(\byear{2010})
\end{barticle}
\endbibitem

%%% 44
\bibitem[\protect\citeauthoryear{{\'O}dor and Kelling}{2019}]{odor2019critical}
\begin{barticle}
\bauthor{\bsnm{{\'O}dor}, \binits{G.}},
\bauthor{\bsnm{Kelling}, \binits{J.}}:
\batitle{Critical synchronization dynamics of the kuramoto model on connectome and small world graphs}.
\bjtitle{Scientific reports}
\bvolume{9}(\bissue{1}),
\bfpage{19621}
(\byear{2019})
\end{barticle}
\endbibitem

%%% 45
\bibitem[\protect\citeauthoryear{D{\"o}rfler and Bullo}{2014}]{dorfler2014synchronization}
\begin{barticle}
\bauthor{\bsnm{D{\"o}rfler}, \binits{F.}},
\bauthor{\bsnm{Bullo}, \binits{F.}}:
\batitle{Synchronization in complex networks of phase oscillators: A survey}.
\bjtitle{Automatica}
\bvolume{50}(\bissue{6}),
\bfpage{1539}--\blpage{1564}
(\byear{2014})
\end{barticle}
\endbibitem

%%% 46
\bibitem[\protect\citeauthoryear{{\'O}dor et~al.}{2021}]{odor2021effect}
\begin{barticle}
\bauthor{\bsnm{{\'O}dor}, \binits{G.}},
\bauthor{\bsnm{Kelling}, \binits{J.}},
\bauthor{\bsnm{Deco}, \binits{G.}}:
\batitle{The effect of noise on the synchronization dynamics of the kuramoto model on a large human connectome graph}.
\bjtitle{Neurocomputing}
\bvolume{461},
\bfpage{696}--\blpage{704}
(\byear{2021})
\end{barticle}
\endbibitem

%%% 47
\bibitem[\protect\citeauthoryear{Anyaeji et~al.}{2021}]{anyaeji2021quantitative}
\begin{barticle}
\bauthor{\bsnm{Anyaeji}, \binits{C.I.}},
\bauthor{\bsnm{Cabral}, \binits{J.}},
\bauthor{\bsnm{Silbersweig}, \binits{D.}}:
\batitle{On a quantitative approach to clinical neuroscience in psychiatry: lessons from the kuramoto model}.
\bjtitle{Harvard Review of Psychiatry}
\bvolume{29}(\bissue{4}),
\bfpage{318}--\blpage{326}
(\byear{2021})
\end{barticle}
\endbibitem

%%% 48
\bibitem[\protect\citeauthoryear{N{\'e}da et~al.}{2000}]{neda2000sound}
\begin{barticle}
\bauthor{\bsnm{N{\'e}da}, \binits{Z.}},
\bauthor{\bsnm{Ravasz}, \binits{E.}},
\bauthor{\bsnm{Brechet}, \binits{Y.}},
\bauthor{\bsnm{Vicsek}, \binits{T.}},
\bauthor{\bsnm{Barab{\'a}si}, \binits{A.-L.}}:
\batitle{The sound of many hands clapping}.
\bjtitle{Nature}
\bvolume{403}(\bissue{6772}),
\bfpage{849}--\blpage{850}
(\byear{2000})
\end{barticle}
\endbibitem

%%% 49
\bibitem[\protect\citeauthoryear{Strogatz et~al.}{2005}]{strogatz2005crowd}
\begin{barticle}
\bauthor{\bsnm{Strogatz}, \binits{S.H.}},
\bauthor{\bsnm{Abrams}, \binits{D.M.}},
\bauthor{\bsnm{McRobie}, \binits{A.}},
\bauthor{\bsnm{Eckhardt}, \binits{B.}},
\bauthor{\bsnm{Ott}, \binits{E.}}:
\batitle{Crowd synchrony on the millennium bridge}.
\bjtitle{Nature}
\bvolume{438}(\bissue{7064}),
\bfpage{43}--\blpage{44}
(\byear{2005})
\end{barticle}
\endbibitem

%%% 50
\bibitem[\protect\citeauthoryear{Kiss et~al.}{2002}]{kiss2002emerging}
\begin{barticle}
\bauthor{\bsnm{Kiss}, \binits{I.Z.}},
\bauthor{\bsnm{Zhai}, \binits{Y.}},
\bauthor{\bsnm{Hudson}, \binits{J.L.}}:
\batitle{Emerging coherence in a population of chemical oscillators}.
\bjtitle{Science}
\bvolume{296}(\bissue{5573}),
\bfpage{1676}--\blpage{1678}
(\byear{2002})
\end{barticle}
\endbibitem

%%% 51
\bibitem[\protect\citeauthoryear{Ludwig and Marquardt}{2013}]{ludwig2013quantum}
\begin{barticle}
\bauthor{\bsnm{Ludwig}, \binits{M.}},
\bauthor{\bsnm{Marquardt}, \binits{F.}}:
\batitle{Quantum many-body dynamics in optomechanical arrays}.
\bjtitle{Physical review letters}
\bvolume{111}(\bissue{7}),
\bfpage{073603}
(\byear{2013})
\end{barticle}
\endbibitem

%%% 52
\bibitem[\protect\citeauthoryear{Strogatz}{2000}]{strogatz2000kuramoto}
\begin{barticle}
\bauthor{\bsnm{Strogatz}, \binits{S.H.}}:
\batitle{From kuramoto to crawford: exploring the onset of synchronization in populations of coupled oscillators}.
\bjtitle{Physica D: Nonlinear Phenomena}
\bvolume{143}(\bissue{1-4}),
\bfpage{1}--\blpage{20}
(\byear{2000})
\end{barticle}
\endbibitem

%%% 53
\bibitem[\protect\citeauthoryear{Gu et~al.}{2025}]{gu2025emergence}
\begin{barticle}
\bauthor{\bsnm{Gu}, \binits{F.}},
\bauthor{\bsnm{Guiselin}, \binits{B.}},
\bauthor{\bsnm{Bain}, \binits{N.}},
\bauthor{\bsnm{Zuriguel}, \binits{I.}},
\bauthor{\bsnm{Bartolo}, \binits{D.}}:
\batitle{Emergence of collective oscillations in massive human crowds}.
\bjtitle{Nature}
\bvolume{638}(\bissue{8049}),
\bfpage{112}--\blpage{119}
(\byear{2025})
\end{barticle}
\endbibitem

%%% 54
\bibitem[\protect\citeauthoryear{Ying et~al.}{2018}]{ying2018graph}
\begin{bchapter}
\bauthor{\bsnm{Ying}, \binits{R.}},
\bauthor{\bsnm{He}, \binits{R.}},
\bauthor{\bsnm{Chen}, \binits{K.}},
\bauthor{\bsnm{Eksombatchai}, \binits{P.}},
\bauthor{\bsnm{Hamilton}, \binits{W.L.}},
\bauthor{\bsnm{Leskovec}, \binits{J.}}:
\bctitle{Graph convolutional neural networks for web-scale recommender systems}.
In: \bbtitle{Proceedings of the 24th ACM SIGKDD International Conference on Knowledge Discovery \& Data Mining},
pp. \bfpage{974}--\blpage{983}
(\byear{2018})
\end{bchapter}
\endbibitem

%%% 55
\bibitem[\protect\citeauthoryear{Zeng et~al.}{2019}]{zeng2019graphsaint}
\begin{botherref}
\oauthor{\bsnm{Zeng}, \binits{H.}},
\oauthor{\bsnm{Zhou}, \binits{H.}},
\oauthor{\bsnm{Srivastava}, \binits{A.}},
\oauthor{\bsnm{Kannan}, \binits{R.}},
\oauthor{\bsnm{Prasanna}, \binits{V.}}:
Graphsaint: Graph sampling based inductive learning method.
arXiv preprint arXiv:1907.04931
(2019)
\end{botherref}
\endbibitem

%%% 56
\bibitem[\protect\citeauthoryear{Manrique et~al.}{2018}]{manrique2018generalized}
\begin{barticle}
\bauthor{\bsnm{Manrique}, \binits{P.D.}},
\bauthor{\bsnm{Zheng}, \binits{M.}},
\bauthor{\bsnm{Cao}, \binits{Z.}},
\bauthor{\bsnm{Restrepo}, \binits{E.M.}},
\bauthor{\bsnm{Johnson}, \binits{N.F.}}:
\batitle{Generalized gelation theory describes onset of online extremist support}.
\bjtitle{Physical review letters}
\bvolume{121}(\bissue{4}),
\bfpage{048301}
(\byear{2018})
\end{barticle}
\endbibitem

%%% 57
\bibitem[\protect\citeauthoryear{Manrique et~al.}{2023}]{manrique2023shockwavelike}
\begin{barticle}
\bauthor{\bsnm{Manrique}, \binits{P.D.}},
\bauthor{\bsnm{Huo}, \binits{F.Y.}},
\bauthor{\bsnm{El~Oud}, \binits{S.}},
\bauthor{\bsnm{Zheng}, \binits{M.}},
\bauthor{\bsnm{Illari}, \binits{L.}},
\bauthor{\bsnm{Johnson}, \binits{N.F.}}:
\batitle{Shockwavelike behavior across social media}.
\bjtitle{Physical Review Letters}
\bvolume{130}(\bissue{23}),
\bfpage{237401}
(\byear{2023})
\end{barticle}
\endbibitem

%%% 58
\bibitem[\protect\citeauthoryear{Johnson et~al.}{2016}]{johnson2016new}
\begin{barticle}
\bauthor{\bsnm{Johnson}, \binits{N.F.}},
\bauthor{\bsnm{Zheng}, \binits{M.}},
\bauthor{\bsnm{Vorobyeva}, \binits{Y.}},
\bauthor{\bsnm{Gabriel}, \binits{A.}},
\bauthor{\bsnm{Qi}, \binits{H.}},
\bauthor{\bsnm{Vel{\'a}squez}, \binits{N.}},
\bauthor{\bsnm{Manrique}, \binits{P.}},
\bauthor{\bsnm{Johnson}, \binits{D.}},
\bauthor{\bsnm{Restrepo}, \binits{E.}},
\bauthor{\bsnm{Song}, \binits{C.}}, \betal:
\batitle{New online ecology of adversarial aggregates: Isis and beyond}.
\bjtitle{Science}
\bvolume{352}(\bissue{6292}),
\bfpage{1459}--\blpage{1463}
(\byear{2016})
\end{barticle}
\endbibitem

%%% 59
\bibitem[\protect\citeauthoryear{Johnson et~al.}{2020}]{johnson2020online}
\begin{barticle}
\bauthor{\bsnm{Johnson}, \binits{N.F.}},
\bauthor{\bsnm{Vel{\'a}squez}, \binits{N.}},
\bauthor{\bsnm{Restrepo}, \binits{N.J.}},
\bauthor{\bsnm{Leahy}, \binits{R.}},
\bauthor{\bsnm{Gabriel}, \binits{N.}},
\bauthor{\bsnm{El~Oud}, \binits{S.}},
\bauthor{\bsnm{Zheng}, \binits{M.}},
\bauthor{\bsnm{Manrique}, \binits{P.}},
\bauthor{\bsnm{Wuchty}, \binits{S.}},
\bauthor{\bsnm{Lupu}, \binits{Y.}}:
\batitle{The online competition between pro-and anti-vaccination views}.
\bjtitle{Nature}
\bvolume{582}(\bissue{7811}),
\bfpage{230}--\blpage{233}
(\byear{2020})
\end{barticle}
\endbibitem

%%% 60
\bibitem[\protect\citeauthoryear{Avalle et~al.}{2024}]{avalle2024persistent}
\begin{barticle}
\bauthor{\bsnm{Avalle}, \binits{M.}},
\bauthor{\bsnm{Di~Marco}, \binits{N.}},
\bauthor{\bsnm{Etta}, \binits{G.}},
\bauthor{\bsnm{Sangiorgio}, \binits{E.}},
\bauthor{\bsnm{Alipour}, \binits{S.}},
\bauthor{\bsnm{Bonetti}, \binits{A.}},
\bauthor{\bsnm{Alvisi}, \binits{L.}},
\bauthor{\bsnm{Scala}, \binits{A.}},
\bauthor{\bsnm{Baronchelli}, \binits{A.}},
\bauthor{\bsnm{Cinelli}, \binits{M.}}, \betal:
\batitle{Persistent interaction patterns across social media platforms and over time}.
\bjtitle{Nature}
\bvolume{628}(\bissue{8008}),
\bfpage{582}--\blpage{589}
(\byear{2024})
\end{barticle}
\endbibitem

%%% 61
\bibitem[\protect\citeauthoryear{Huo et~al.}{2024}]{huo2024multi}
\begin{barticle}
\bauthor{\bsnm{Huo}, \binits{F.Y.}},
\bauthor{\bsnm{Manrique}, \binits{P.D.}},
\bauthor{\bsnm{Johnson}, \binits{N.F.}}:
\batitle{Multispecies cohesion: Humans, machinery, ai, and beyond}.
\bjtitle{Physical Review Letters}
\bvolume{133}(\bissue{24}),
\bfpage{247401}
(\byear{2024})
\end{barticle}
\endbibitem

%%% 62
\bibitem[\protect\citeauthoryear{Gabriel et~al.}{2023}]{gabriel2023inductive}
\begin{barticle}
\bauthor{\bsnm{Gabriel}, \binits{N.A.}},
\bauthor{\bsnm{Broniatowski}, \binits{D.A.}},
\bauthor{\bsnm{Johnson}, \binits{N.F.}}:
\batitle{Inductive detection of influence operations via graph learning}.
\bjtitle{Scientific Reports}
\bvolume{13}(\bissue{1}),
\bfpage{22571}
(\byear{2023})
\end{barticle}
\endbibitem

%%% 63
\bibitem[\protect\citeauthoryear{Hyndman and Koehler}{2006}]{hyndman2006another}
\begin{barticle}
\bauthor{\bsnm{Hyndman}, \binits{R.J.}},
\bauthor{\bsnm{Koehler}, \binits{A.B.}}:
\batitle{Another look at measures of forecast accuracy}.
\bjtitle{International journal of forecasting}
\bvolume{22}(\bissue{4}),
\bfpage{679}--\blpage{688}
(\byear{2006})
\end{barticle}
\endbibitem

%%% 64
\bibitem[\protect\citeauthoryear{Makridakis et~al.}{2020}]{makridakis2020m4}
\begin{barticle}
\bauthor{\bsnm{Makridakis}, \binits{S.}},
\bauthor{\bsnm{Spiliotis}, \binits{E.}},
\bauthor{\bsnm{Assimakopoulos}, \binits{V.}}:
\batitle{The m4 competition: 100,000 time series and 61 forecasting methods}.
\bjtitle{International Journal of Forecasting}
\bvolume{36}(\bissue{1}),
\bfpage{54}--\blpage{74}
(\byear{2020})
\end{barticle}
\endbibitem

%%% 65
\bibitem[\protect\citeauthoryear{van~der Werf et~al.}{2023}]{van2023all}
\begin{barticle}
\bauthor{\bsnm{Werf}, \binits{J.M.E.}},
\bauthor{\bsnm{Polyvyanyy}, \binits{A.}},
\bauthor{\bsnm{Wensveen}, \binits{B.R.}},
\bauthor{\bsnm{Brinkhuis}, \binits{M.}},
\bauthor{\bsnm{Reijers}, \binits{H.A.}}:
\batitle{All that glitters is not gold: Four maturity stages of process discovery algorithms}.
\bjtitle{Information Systems}
\bvolume{114},
\bfpage{102155}
(\byear{2023})
\end{barticle}
\endbibitem

%%% 66
\bibitem[\protect\citeauthoryear{Jin et~al.}{2022}]{jin2022mionet}
\begin{barticle}
\bauthor{\bsnm{Jin}, \binits{P.}},
\bauthor{\bsnm{Meng}, \binits{S.}},
\bauthor{\bsnm{Lu}, \binits{L.}}:
\batitle{Mionet: Learning multiple-input operators via tensor product}.
\bjtitle{SIAM Journal on Scientific Computing}
\bvolume{44}(\bissue{6}),
\bfpage{3490}--\blpage{3514}
(\byear{2022})
\end{barticle}
\endbibitem

%%% 67
\bibitem[\protect\citeauthoryear{Kaplan et~al.}{2020}]{kaplan2020scaling}
\begin{botherref}
\oauthor{\bsnm{Kaplan}, \binits{J.}},
\oauthor{\bsnm{McCandlish}, \binits{S.}},
\oauthor{\bsnm{Henighan}, \binits{T.}},
\oauthor{\bsnm{Brown}, \binits{T.B.}},
\oauthor{\bsnm{Chess}, \binits{B.}},
\oauthor{\bsnm{Child}, \binits{R.}},
\oauthor{\bsnm{Gray}, \binits{S.}},
\oauthor{\bsnm{Radford}, \binits{A.}},
\oauthor{\bsnm{Wu}, \binits{J.}},
\oauthor{\bsnm{Amodei}, \binits{D.}}:
Scaling laws for neural language models.
arXiv preprint arXiv:2001.08361
(2020)
\end{botherref}
\endbibitem

%%% 68
\bibitem[\protect\citeauthoryear{Dosovitskiy et~al.}{2020}]{dosovitskiy2020image}
\begin{botherref}
\oauthor{\bsnm{Dosovitskiy}, \binits{A.}},
\oauthor{\bsnm{Beyer}, \binits{L.}},
\oauthor{\bsnm{Kolesnikov}, \binits{A.}},
\oauthor{\bsnm{Weissenborn}, \binits{D.}},
\oauthor{\bsnm{Zhai}, \binits{X.}},
\oauthor{\bsnm{Unterthiner}, \binits{T.}},
\oauthor{\bsnm{Dehghani}, \binits{M.}},
\oauthor{\bsnm{Minderer}, \binits{M.}},
\oauthor{\bsnm{Heigold}, \binits{G.}},
\oauthor{\bsnm{Gelly}, \binits{S.}}, et al.:
An image is worth 16x16 words: Transformers for image recognition at scale.
arXiv preprint arXiv:2010.11929
(2020)
\end{botherref}
\endbibitem

%%% 69
\bibitem[\protect\citeauthoryear{Bahri et~al.}{2024}]{bahri2024explaining}
\begin{barticle}
\bauthor{\bsnm{Bahri}, \binits{Y.}},
\bauthor{\bsnm{Dyer}, \binits{E.}},
\bauthor{\bsnm{Kaplan}, \binits{J.}},
\bauthor{\bsnm{Lee}, \binits{J.}},
\bauthor{\bsnm{Sharma}, \binits{U.}}:
\batitle{Explaining neural scaling laws}.
\bjtitle{Proceedings of the National Academy of Sciences}
\bvolume{121}(\bissue{27}),
\bfpage{2311878121}
(\byear{2024})
\end{barticle}
\endbibitem

%%% 70
\bibitem[\protect\citeauthoryear{Reuveni et~al.}{2010}]{reuveni2010anomalies}
\begin{barticle}
\bauthor{\bsnm{Reuveni}, \binits{S.}},
\bauthor{\bsnm{Granek}, \binits{R.}},
\bauthor{\bsnm{Klafter}, \binits{J.}}:
\batitle{Anomalies in the vibrational dynamics of proteins are a consequence of fractal-like structure}.
\bjtitle{Proceedings of the National Academy of Sciences}
\bvolume{107}(\bissue{31}),
\bfpage{13696}--\blpage{13700}
(\byear{2010})
\end{barticle}
\endbibitem

%%% 71
\bibitem[\protect\citeauthoryear{Lacasa and G{\'o}mez-Gardenes}{2013}]{lacasa2013correlation}
\begin{barticle}
\bauthor{\bsnm{Lacasa}, \binits{L.}},
\bauthor{\bsnm{G{\'o}mez-Gardenes}, \binits{J.}}:
\batitle{Correlation dimension of complex networks}.
\bjtitle{Physical review letters}
\bvolume{110}(\bissue{16}),
\bfpage{168703}
(\byear{2013})
\end{barticle}
\endbibitem

%%% 72
\bibitem[\protect\citeauthoryear{Arnaudon et~al.}{2020}]{arnaudon2020scale}
\begin{barticle}
\bauthor{\bsnm{Arnaudon}, \binits{A.}},
\bauthor{\bsnm{Peach}, \binits{R.L.}},
\bauthor{\bsnm{Barahona}, \binits{M.}}:
\batitle{Scale-dependent measure of network centrality from diffusion dynamics}.
\bjtitle{Physical Review Research}
\bvolume{2}(\bissue{3}),
\bfpage{033104}
(\byear{2020})
\end{barticle}
\endbibitem

%%% 73
\bibitem[\protect\citeauthoryear{Peach et~al.}{2019}]{peach2019semi}
\begin{botherref}
\oauthor{\bsnm{Peach}, \binits{R.L.}},
\oauthor{\bsnm{Arnaudon}, \binits{A.}},
\oauthor{\bsnm{Barahona}, \binits{M.}}:
Semi-supervised classification on graphs using explicit diffusion dynamics.
arXiv preprint arXiv:1909.11117
(2019)
\end{botherref}
\endbibitem

%%% 74
\bibitem[\protect\citeauthoryear{Peach et~al.}{2022}]{peach2022relative}
\begin{barticle}
\bauthor{\bsnm{Peach}, \binits{R.}},
\bauthor{\bsnm{Arnaudon}, \binits{A.}},
\bauthor{\bsnm{Barahona}, \binits{M.}}:
\batitle{Relative, local and global dimension in complex networks}.
\bjtitle{Nature Communications}
\bvolume{13}(\bissue{1}),
\bfpage{3088}
(\byear{2022})
\end{barticle}
\endbibitem

%%% 75
\bibitem[\protect\citeauthoryear{Lei~Ba et~al.}{2016}]{lei2016layer}
\begin{botherref}
\oauthor{\bsnm{Lei~Ba}, \binits{J.}},
\oauthor{\bsnm{Kiros}, \binits{J.R.}},
\oauthor{\bsnm{Hinton}, \binits{G.E.}}:
Layer normalization.
arXiv preprint arXiv:1607.06450
(2016)
\end{botherref}
\endbibitem

%%% 76
\bibitem[\protect\citeauthoryear{Vaswani et~al.}{2017}]{vaswani2017attention}
\begin{botherref}
\oauthor{\bsnm{Vaswani}, \binits{A.}},
\oauthor{\bsnm{Shazeer}, \binits{N.}},
\oauthor{\bsnm{Parmar}, \binits{N.}},
\oauthor{\bsnm{Uszkoreit}, \binits{J.}},
\oauthor{\bsnm{Jones}, \binits{L.}},
\oauthor{\bsnm{Gomez}, \binits{A.N.}},
\oauthor{\bsnm{Kaiser}, \binits{{\L}.}},
\oauthor{\bsnm{Polosukhin}, \binits{I.}}:
Attention is all you need.
Advances in neural information processing systems
\textbf{30}
(2017)
\end{botherref}
\endbibitem

%%% 77
\bibitem[\protect\citeauthoryear{Liu et~al.}{2024}]{liu2024kan}
\begin{botherref}
\oauthor{\bsnm{Liu}, \binits{Z.}},
\oauthor{\bsnm{Wang}, \binits{Y.}},
\oauthor{\bsnm{Vaidya}, \binits{S.}},
\oauthor{\bsnm{Ruehle}, \binits{F.}},
\oauthor{\bsnm{Halverson}, \binits{J.}},
\oauthor{\bsnm{Solja{\v{c}}i{\'c}}, \binits{M.}},
\oauthor{\bsnm{Hou}, \binits{T.Y.}},
\oauthor{\bsnm{Tegmark}, \binits{M.}}:
Kan: Kolmogorov-arnold networks.
arXiv preprint arXiv:2404.19756
(2024)
\end{botherref}
\endbibitem

%%% 78
\bibitem[\protect\citeauthoryear{Shukla et~al.}{2024}]{shukla2024comprehensive}
\begin{botherref}
\oauthor{\bsnm{Shukla}, \binits{K.}},
\oauthor{\bsnm{Toscano}, \binits{J.D.}},
\oauthor{\bsnm{Wang}, \binits{Z.}},
\oauthor{\bsnm{Zou}, \binits{Z.}},
\oauthor{\bsnm{Karniadakis}, \binits{G.E.}}:
A comprehensive and fair comparison between mlp and kan representations for differential equations and operator networks.
arXiv preprint arXiv:2406.02917
(2024)
\end{botherref}
\endbibitem

%%% 79
\bibitem[\protect\citeauthoryear{Ying et~al.}{2018}]{ying2018hierarchical}
\begin{botherref}
\oauthor{\bsnm{Ying}, \binits{Z.}},
\oauthor{\bsnm{You}, \binits{J.}},
\oauthor{\bsnm{Morris}, \binits{C.}},
\oauthor{\bsnm{Ren}, \binits{X.}},
\oauthor{\bsnm{Hamilton}, \binits{W.}},
\oauthor{\bsnm{Leskovec}, \binits{J.}}:
Hierarchical graph representation learning with differentiable pooling.
Advances in neural information processing systems
\textbf{31}
(2018)
\end{botherref}
\endbibitem

%%% 80
\bibitem[\protect\citeauthoryear{Lebanon}{2012}]{lebanon2012learning}
\begin{botherref}
\oauthor{\bsnm{Lebanon}, \binits{G.}}:
Learning riemannian metrics.
arXiv preprint arXiv:1212.2474
(2012)
\end{botherref}
\endbibitem

%%% 81
\bibitem[\protect\citeauthoryear{Hauberg et~al.}{2012}]{hauberg2012geometric}
\begin{botherref}
\oauthor{\bsnm{Hauberg}, \binits{S.}},
\oauthor{\bsnm{Freifeld}, \binits{O.}},
\oauthor{\bsnm{Black}, \binits{M.}}:
A geometric take on metric learning.
Advances in Neural Information Processing Systems
\textbf{25}
(2012)
\end{botherref}
\endbibitem

%%% 82
\bibitem[\protect\citeauthoryear{Arvanitidis et~al.}{2016}]{arvanitidis2016locally}
\begin{botherref}
\oauthor{\bsnm{Arvanitidis}, \binits{G.}},
\oauthor{\bsnm{Hansen}, \binits{L.K.}},
\oauthor{\bsnm{Hauberg}, \binits{S.}}:
A locally adaptive normal distribution.
Advances in Neural Information Processing Systems
\textbf{29}
(2016)
\end{botherref}
\endbibitem

%%% 83
\bibitem[\protect\citeauthoryear{Scarvelis and Solomon}{2022}]{scarvelis2022riemannian}
\begin{botherref}
\oauthor{\bsnm{Scarvelis}, \binits{C.}},
\oauthor{\bsnm{Solomon}, \binits{J.}}:
Riemannian metric learning via optimal transport.
arXiv preprint arXiv:2205.09244
(2022)
\end{botherref}
\endbibitem

%%% 84
\bibitem[\protect\citeauthoryear{Qiu and Dai}{2024}]{qiu2024estimating}
\begin{botherref}
\oauthor{\bsnm{Qiu}, \binits{J.}},
\oauthor{\bsnm{Dai}, \binits{X.}}:
Estimating riemannian metric with noise-contaminated intrinsic distance.
Advances in Neural Information Processing Systems
\textbf{36}
(2024)
\end{botherref}
\endbibitem

%%% 85
\bibitem[\protect\citeauthoryear{Diepeveen et~al.}{2024}]{diepeveen2024score}
\begin{botherref}
\oauthor{\bsnm{Diepeveen}, \binits{W.}},
\oauthor{\bsnm{Batzolis}, \binits{G.}},
\oauthor{\bsnm{Shumaylov}, \binits{Z.}},
\oauthor{\bsnm{Sch{\"o}nlieb}, \binits{C.-B.}}:
Score-based pullback riemannian geometry.
arXiv preprint arXiv:2410.01950
(2024)
\end{botherref}
\endbibitem

%%% 86
\bibitem[\protect\citeauthoryear{Chow and Knopf}{2004}]{chow2004ricci}
\begin{botherref}
\oauthor{\bsnm{Chow}, \binits{B.}},
\oauthor{\bsnm{Knopf}, \binits{D.}}:
The ricci flow: An introduction: An introduction
(2004)
\end{botherref}
\endbibitem

%%% 87
\bibitem[\protect\citeauthoryear{Gracyk}{2024}]{gracyk2024ricci}
\begin{botherref}
\oauthor{\bsnm{Gracyk}, \binits{A.}}:
Ricci flow-guided autoencoders in learning time-dependent dynamics.
arXiv preprint arXiv:2401.14591
(2024)
\end{botherref}
\endbibitem

%%% 88
\bibitem[\protect\citeauthoryear{Lee et~al.}{2019}]{lee2019self}
\begin{bchapter}
\bauthor{\bsnm{Lee}, \binits{J.}},
\bauthor{\bsnm{Lee}, \binits{I.}},
\bauthor{\bsnm{Kang}, \binits{J.}}:
\bctitle{Self-attention graph pooling}.
In: \bbtitle{International Conference on Machine Learning},
pp. \bfpage{3734}--\blpage{3743}
(\byear{2019}).
\bcomment{PMLR}
\end{bchapter}
\endbibitem

%%% 89
\bibitem[\protect\citeauthoryear{Belkin and Niyogi}{2003}]{belkin2003}
\begin{barticle}
\bauthor{\bsnm{Belkin}, \binits{M.}},
\bauthor{\bsnm{Niyogi}, \binits{P.}}:
\batitle{Laplacian eigenmaps for dimensionality reduction and data representation}.
\bjtitle{Neural computation}
\bvolume{15}(\bissue{6}),
\bfpage{1373}--\blpage{1396}
(\byear{2003})
\end{barticle}
\endbibitem

%%% 90
\bibitem[\protect\citeauthoryear{Dwivedi et~al.}{2020}]{LapEig}
\begin{botherref}
\oauthor{\bsnm{Dwivedi}, \binits{V.P.}},
\oauthor{\bsnm{Joshi}, \binits{C.K.}},
\oauthor{\bsnm{Laurent}, \binits{T.}},
\oauthor{\bsnm{Bengio}, \binits{Y.}},
\oauthor{\bsnm{Bresson}, \binits{X.}}:
Benchmarking graph neural networks.
CoRR
\textbf{abs/2003.00982}
(2020)
\end{botherref}
\endbibitem

%%% 91
\bibitem[\protect\citeauthoryear{Bahdanau et~al.}{2014}]{bahdanau2014neural}
\begin{botherref}
\oauthor{\bsnm{Bahdanau}, \binits{D.}},
\oauthor{\bsnm{Cho}, \binits{K.}},
\oauthor{\bsnm{Bengio}, \binits{Y.}}:
Neural machine translation by jointly learning to align and translate.
arXiv preprint arXiv:1409.0473
(2014)
\end{botherref}
\endbibitem

%%% 92
\bibitem[\protect\citeauthoryear{Kissas et~al.}{2022}]{LOCA2022}
\begin{barticle}
\bauthor{\bsnm{Kissas}, \binits{G.}},
\bauthor{\bsnm{Seidman}, \binits{J.H.}},
\bauthor{\bsnm{Guilhoto}, \binits{L.F.}},
\bauthor{\bsnm{Preciado}, \binits{V.M.}},
\bauthor{\bsnm{Pappas}, \binits{G.J.}},
\bauthor{\bsnm{Perdikaris}, \binits{P.}}:
\batitle{Learning operators with coupled attention}.
\bjtitle{Journal of Machine Learning Research}
\bvolume{23}(\bissue{215}),
\bfpage{1}--\blpage{63}
(\byear{2022})
\end{barticle}
\endbibitem

%%% 93
\bibitem[\protect\citeauthoryear{Wang et~al.}{2023}]{wang2023expert}
\begin{botherref}
\oauthor{\bsnm{Wang}, \binits{S.}},
\oauthor{\bsnm{Sankaran}, \binits{S.}},
\oauthor{\bsnm{Wang}, \binits{H.}},
\oauthor{\bsnm{Perdikaris}, \binits{P.}}:
An expert's guide to training physics-informed neural networks.
arXiv preprint arXiv:2308.08468
(2023)
\end{botherref}
\endbibitem

%%% 94
\bibitem[\protect\citeauthoryear{Raissi}{2018}]{raissi2018deep}
\begin{barticle}
\bauthor{\bsnm{Raissi}, \binits{M.}}:
\batitle{Deep hidden physics models: Deep learning of nonlinear partial differential equations}.
\bjtitle{The Journal of Machine Learning Research}
\bvolume{19}(\bissue{1}),
\bfpage{932}--\blpage{955}
(\byear{2018})
\end{barticle}
\endbibitem

%%% 95
\bibitem[\protect\citeauthoryear{Raissi et~al.}{2019}]{raissi2019physics}
\begin{barticle}
\bauthor{\bsnm{Raissi}, \binits{M.}},
\bauthor{\bsnm{Perdikaris}, \binits{P.}},
\bauthor{\bsnm{Karniadakis}, \binits{G.E.}}:
\batitle{Physics-informed neural networks: A deep learning framework for solving forward and inverse problems involving nonlinear partial differential equations}.
\bjtitle{Journal of Computational physics}
\bvolume{378},
\bfpage{686}--\blpage{707}
(\byear{2019})
\end{barticle}
\endbibitem

%%% 96
\bibitem[\protect\citeauthoryear{Lu et~al.}{2022}]{lu2022fair}
\begin{barticle}
\bauthor{\bsnm{Lu}, \binits{L.}},
\bauthor{\bsnm{Meng}, \binits{X.}},
\bauthor{\bsnm{Cai}, \binits{S.}},
\bauthor{\bsnm{Mao}, \binits{Z.}},
\bauthor{\bsnm{Goswami}, \binits{S.}},
\bauthor{\bsnm{Zhang}, \binits{Z.}},
\bauthor{\bsnm{Karniadakis}, \binits{G.E.}}:
\batitle{A comprehensive and fair comparison of two neural operators (with practical extensions) based on fair data}.
\bjtitle{Computer Methods in Applied Mechanics and Engineering}
\bvolume{393},
\bfpage{114778}
(\byear{2022})
\end{barticle}
\endbibitem

%%% 97
\bibitem[\protect\citeauthoryear{Yu et~al.}{2022}]{yu2022gradient}
\begin{barticle}
\bauthor{\bsnm{Yu}, \binits{J.}},
\bauthor{\bsnm{Lu}, \binits{L.}},
\bauthor{\bsnm{Meng}, \binits{X.}},
\bauthor{\bsnm{Karniadakis}, \binits{G.E.}}:
\batitle{Gradient-enhanced physics-informed neural networks for forward and inverse pde problems}.
\bjtitle{Computer Methods in Applied Mechanics and Engineering}
\bvolume{393},
\bfpage{114823}
(\byear{2022})
\end{barticle}
\endbibitem

%%% 98
\bibitem[\protect\citeauthoryear{Nickel and Kiela}{2017}]{nickel2017poincare}
\begin{botherref}
\oauthor{\bsnm{Nickel}, \binits{M.}},
\oauthor{\bsnm{Kiela}, \binits{D.}}:
Poincar{\'e} embeddings for learning hierarchical representations.
Advances in neural information processing systems
\textbf{30}
(2017)
\end{botherref}
\endbibitem

%%% 99
\bibitem[\protect\citeauthoryear{Newman}{2018}]{newman2018networks}
\begin{botherref}
\oauthor{\bsnm{Newman}, \binits{M.}}:
Networks.
Oxford university press
(2018)
\end{botherref}
\endbibitem

%%% 100
\bibitem[\protect\citeauthoryear{Grover and Leskovec}{2016}]{grover2016node2vec}
\begin{bchapter}
\bauthor{\bsnm{Grover}, \binits{A.}},
\bauthor{\bsnm{Leskovec}, \binits{J.}}:
\bctitle{node2vec: Scalable feature learning for networks}.
In: \bbtitle{Proceedings of the 22nd ACM SIGKDD International Conference on Knowledge Discovery and Data Mining},
pp. \bfpage{855}--\blpage{864}
(\byear{2016})
\end{bchapter}
\endbibitem

%%% 101
\bibitem[\protect\citeauthoryear{Perozzi et~al.}{2014}]{perozzi2014deepwalk}
\begin{bchapter}
\bauthor{\bsnm{Perozzi}, \binits{B.}},
\bauthor{\bsnm{Al-Rfou}, \binits{R.}},
\bauthor{\bsnm{Skiena}, \binits{S.}}:
\bctitle{Deepwalk: Online learning of social representations}.
In: \bbtitle{Proceedings of the 20th ACM SIGKDD International Conference on Knowledge Discovery and Data Mining},
pp. \bfpage{701}--\blpage{710}
(\byear{2014})
\end{bchapter}
\endbibitem

%%% 102
\bibitem[\protect\citeauthoryear{Qiu et~al.}{2018}]{qiu2018network}
\begin{bchapter}
\bauthor{\bsnm{Qiu}, \binits{J.}},
\bauthor{\bsnm{Dong}, \binits{Y.}},
\bauthor{\bsnm{Ma}, \binits{H.}},
\bauthor{\bsnm{Li}, \binits{J.}},
\bauthor{\bsnm{Wang}, \binits{K.}},
\bauthor{\bsnm{Tang}, \binits{J.}}:
\bctitle{Network embedding as matrix factorization: Unifying deepwalk, line, pte, and node2vec}.
In: \bbtitle{Proceedings of the Eleventh ACM International Conference on Web Search and Data Mining},
pp. \bfpage{459}--\blpage{467}
(\byear{2018})
\end{bchapter}
\endbibitem

%%% 103
\bibitem[\protect\citeauthoryear{Wang et~al.}{2019}]{wang2019dynamic}
\begin{barticle}
\bauthor{\bsnm{Wang}, \binits{Y.}},
\bauthor{\bsnm{Sun}, \binits{Y.}},
\bauthor{\bsnm{Liu}, \binits{Z.}},
\bauthor{\bsnm{Sarma}, \binits{S.E.}},
\bauthor{\bsnm{Bronstein}, \binits{M.M.}},
\bauthor{\bsnm{Solomon}, \binits{J.M.}}:
\batitle{Dynamic graph cnn for learning on point clouds}.
\bjtitle{ACM Transactions on Graphics (tog)}
\bvolume{38}(\bissue{5}),
\bfpage{1}--\blpage{12}
(\byear{2019})
\end{barticle}
\endbibitem

%%% 104
\bibitem[\protect\citeauthoryear{Corso et~al.}{2020}]{corso2020principal}
\begin{barticle}
\bauthor{\bsnm{Corso}, \binits{G.}},
\bauthor{\bsnm{Cavalleri}, \binits{L.}},
\bauthor{\bsnm{Beaini}, \binits{D.}},
\bauthor{\bsnm{Li{\`o}}, \binits{P.}},
\bauthor{\bsnm{Veli{\v{c}}kovi{\'c}}, \binits{P.}}:
\batitle{Principal neighbourhood aggregation for graph nets}.
\bjtitle{Advances in Neural Information Processing Systems}
\bvolume{33},
\bfpage{13260}--\blpage{13271}
(\byear{2020})
\end{barticle}
\endbibitem

%%% 105
\bibitem[\protect\citeauthoryear{Kingma}{2014}]{kingma2014adam}
\begin{botherref}
\oauthor{\bsnm{Kingma}, \binits{D.P.}}:
Adam: A method for stochastic optimization.
arXiv preprint arXiv:1412.6980
(2014)
\end{botherref}
\endbibitem

%%% 106
\bibitem[\protect\citeauthoryear{Loshchilov}{2017}]{loshchilov2017decoupled}
\begin{botherref}
\oauthor{\bsnm{Loshchilov}, \binits{I.}}:
Decoupled weight decay regularization.
arXiv preprint arXiv:1711.05101
(2017)
\end{botherref}
\endbibitem

\end{thebibliography}

%\end{document}
\newpage
\appendix

%\section{Neural Operators}
%Physical systems often have regularities that can be represented mathematically by function spaces. 

%\subsection{Function Spaces}

\section{Graph Learning}
A number of datasets come equipped, explicitly or implicitly, with a graph structure. For example, social networks, power grids, transportation systems, food webs, citation networks, gene regulatory networks, metabolic networks, and brain networks can each be represented as a discrete set of nodes/vertices and edges, often specified as a set $G = (V,E)$~\cite{newman2018networks,strogatz2000kuramoto}. In recent years, several techniques have been developed to apply machine learning to graph structured data sets, which we briefly describe here. \\

Recently, graph neural networks (GNNs) or message passing neural networks (MPNNs) have become the standard for machine learning on graphs~\cite{kipf2016semi,GAT2018,MPNN2017,you2022nonlocal}. The advantage of MPNNs over earlier graph embedding techniques (see next paragraph) is the ability to process node-wise features \textit{and} graph structure simultaneously, as well as modeling relationships between nodes while considering their context within the graph. \\

Early approaches in graph learning including unsupervised graph embedding techniques such as node2ve~\cite{grover2016node2vec}, DeepWalk~\cite{perozzi2014deepwalk}, and matrix-factorization based methods~\cite{qiu2018network} are compatible with standard machine learning models such as logistic regression, random forest, support vector machines, and MLPs. The graph embedding approach thereby offers a straightforward way to utilize graph structure in machine learning tasks. Interestingly, the message passing paradigm that has become the standard in graph learning is not only compatible with graph embedding approaches, but may enhance message passing approaches when applicable~\cite{chami2022machine}. For example, graph embeddings can be employed as positional embeddings for message passing neural networks when embeddings can be precomputed (i.e., for transductive problems). This is particularly beneficial for representing global position of nodes while simultaneously restricting the computational field during message passing computations to control variance, avoid  neighborhood explosion, and reduce oversmoothing~\cite{hamilton2017inductive,ying2018graph,zeng2019graphsaint} .  \\

\subsection{Message Passing Neural Networks}
Here we introduce several MPNNs (with GNNs as a subset) formulated as message passing layers. Most MPNNs can be specified by their $k$th layer as
\begin{flalign}
    &&\textbf{h}_i^{(k+1)} &= W^{(k)} \textbf{h}_i^{(k)} + \textbf{b}^{(k)}&&&&&&&&&&& \mbox{(affine transformation)}\\
    &&\textbf{m}_{ij}^{(k+1)} &= \phi\bigg(\textbf{h}_i^{(k)},\textbf{h}_j^{(k)},\textbf{e}_{ij}\bigg)&&&&&&&&&&& \mbox{(message passing)}\\
    &&\textbf{h}_i^{(k+1)} &= \underset{}{\textrm{AGG}_i}\bigg(\textbf{m}_{ij}^{(k+1)}\bigg)&&&&&&&&&&& \mbox{(feature aggregation)}\\
    &&\textbf{h}_i^{(k+1)} &= \sigma\bigg( \textbf{h}_i^{(k+1)} \bigg)&&&&&&&&&&& \mbox{(non-linearity)}
\end{flalign}
where $\phi$ is a message passing function such as an MLP, and $\textbf{e}_{ij}$ are edge-wise features. In this formulation, we can write the well known Graph Convolutional Network (GCN) layer by specifying the aggregation function as\cite{kipf2016semi}
\begin{flalign}
        &&\textbf{m}_{ij}^{(k+1)} &= \textbf{h}_j^{(k)} &&&&&&&&&&& \\
        &&\underset{}{\textrm{AGG}_i}\bigg(\textbf{m}_{ij}^{(k+1)}\bigg) &= \sum_{j \in \mathcal{N}(i)} \frac{1}{\sqrt{d_i d_j}} \textbf{m}_{ij}^{(k+1)} &&&&&&&&&&&&& 
\end{flalign}
where $d_i$ is the degree of the $i$th node. \\

Other choices of the message passing function define the GAT layer \cite{GAT2018}
\begin{flalign} 
    \textbf{m}_{ij}^{(k+1)} &= \underset{j \in \mathcal{N}(i)}{\textrm{softmax}}\bigg(\textrm{LeakyReLu}\Big(\mathbf{a}^T\big[ {W}_r \,\textbf{h}_i^{(k)}|| {W}_s \, h_j^{(k)}\big]\Big)\bigg) \\
    \underset{}{\textrm{AGG}_i}\bigg(\textbf{m}_{ij}^{(k+1)}\bigg) &= \underset{j \in \mathcal{N}(i)}{\sum} \textbf{m}_{ij}^{(k+1)} {W} \textbf{h}_j^{(k+1)}
\end{flalign}
where $||$ denotes concatenation, and the EdgeConv \cite{wang2019dynamic} layer
\begin{flalign} 
    \textbf{m}_{ij}^{(k+1)} &= \phi\Big( \textbf{h}_i^{(k)} || \textbf{h}_j^{(k)} - \textbf{h}_i^{(k)}\Big) \\
    \underset{}{\textrm{AGG}_i}\bigg(\textbf{m}_{ij}^{(k+1)}\bigg) &= \underset{j \in \mathcal{N}(i)}{\sum} \textbf{m}_{ij}^{(k+1)}
\end{flalign}
where $\phi$ is an MLP. The effect of different message passing rules can have a dramatic increase in MPNN performance, and in general the best choice is application dependent. \\

In both the GAT and EdgeConv layers, the resulting representation use a sum aggregation rule. However, this operation discards a significant amount of information about the statistics of each neighborhood features $j \in \mathcal{N}(i)$. Other aggregation functions such as 
\begin{flalign} 
    \underset{j \in \mathcal{N}(i)}{\textrm{min}(\mathbf{x}_j)}&  \\
    \underset{j \in \mathcal{N}(i)}{\textrm{max}(\mathbf{x}_j)}&  \\
    \underset{j \in \mathcal{N}(i)}{\textrm{std}(\mathbf{x}_j)}&  \\
    \underset{j \in \mathcal{N}(i)}{\textrm{softmax}}&(\mathbf{x}_j)
\end{flalign}
can provide additional information while still compressing neighborhood statistics. Combining several aggregation functions in some manner is a common approach to enhance performance in graph networks \cite{corso2020principal}, for example as 
\begin{equation}
    \underset{}{\textrm{AGG}_i}(\mathbf{x}) = \psi\Big( \, \underset{j \in \mathcal{N}(i)}{\textrm{sum}(\mathbf{x}_j)} ||  \underset{j \in \mathcal{N}(i)}{\textrm{max}(\mathbf{x}_j)} ||  \underset{j \in \mathcal{N}(i)}{\textrm{mean}(\mathbf{x}_j)} ||  \underset{j \in \mathcal{N}(i)}{\textrm{std}(\mathbf{x}_j)} \, \Big)
\end{equation}
where $\psi$ is an linear projection or MLP.
\subsection{Hyperbolic Representation}
Representing graphs possessing a hierarchical or tree-like structure with Euclidean representations is known to incur significant distortion \cite{ganea2018hyperbolic,liu2019hyperbolic,chami2019hyperbolic}, with negative impact on downstream performance. Computing graph representations in hyperbolic space is a a natural approach to represent hierarchical structures, which are common in scientific and social datasets. The two most common models of hyperbolic space for such purposes are the Poincar\'{e} disk and the Hyperboloid model. Topologically, these spaces are equivalent in $d$ dimensions since they are  unique, simply connected Riemannian manifolds with constant negative sectional curvature. The choice of manifold is often dictated by practical considerations such as numerical stability during optimization \cite{liu2019hyperbolic,chami2019hyperbolic}. In particular, for graphs with an extremely hierarchical structure, larger negative curvatures may be necessary for accurate representation, in which case the Hyperboloid model is known to be more stable. For applications with mild curvatures and smaller learning rates, optimization directly on the Poincar\'{e} is feasible. \\

Hyperbolic representations can be obtained following \cite{chami2019hyperbolic} by first assigning input features $\mathbf{x}_i \in \mathbb{R}^n$ to the tangent space of the hyperbolic manifold $\mathcal{M}^{n,K}$ with dimension $n$ and negative curvature $K$ at the origin $\mathbf{o}$, which we denote as $\mathcal{T}_{\mathbf{o}}\mathcal{M}^{n,K} \cong \mathbb{R}^n$. From here, we can project the input features to the manifold by the exponential map $\textrm{exp}^{K}_{\mathbf{o}}: \mathcal{T}_{\mathbf{o}}\mathcal{M}^{n,K} \rightarrow \mathcal{M}^{n,K}$ to obtain hyperbolic features

\begin{equation}
    \mathbf{x}_i^{{H}} =  \textrm{exp}^{K}_{\mathbf{o}}(\mathbf{x}_i) \in \mathcal{M}^{n,K},
\end{equation}
which has an inverse logarithmic map $\textrm{log}^{K}_{\mathbf{o}}: \mathcal{M}^{n,K} \rightarrow \mathcal{T}_{\mathbf{o}}\mathcal{M}^{n,K}$ that maps points on the manifold to the tangent space of the origin. After projecting to the manifold, hyperbolic representations can be computed at the $k$th layer as

\begin{subequations}
\label{eq:hgcn}
\begin{align}
    &&\textbf{h}_i^{(k+1)} &= \Big( W^{(k)} \otimes^{K} \textbf{h}_i^{(k)} \Big) \oplus^{K} \textbf{b}^{(l)}&&&&&&&&&&& \mbox{(affine transformation)}\\
    &&\textbf{h}_i^{(k+1)} &= \underset{}{\textrm{AGG}_i^{K}}\bigg(\textbf{h}_j^{(k+1)}\bigg)&&&&&&&&&&& \mbox{(feature aggregation)}\\
    &&\textbf{h}_i^{(k+1)} &= \sigma^{\otimes^{K}}\bigg( \textbf{h}_i^{(k+1)} \bigg)&&&&&&&&&&& \mbox{(non-linearity)}
\end{align}
\end{subequations}

where the curvature $K$ is constant in each hidden layer. While computing representations in hyperbolic space, one can define matrix multiplication, vector addition, and activation functions as 

\begin{subequations}
\label{eq:hyper}
\begin{align}
    && W \otimes^{K} \textbf{x} &= \textrm{exp}^{K}_{\mathbf{o}} \Big( \, W \, \textrm{log}^{K}_{\mathbf{o}} (\textbf{x} ) \Big) &&&&&&&&&&& \mbox{(matrix multiplication)}\\
    && \textbf{x} \oplus^{K} \textbf{b}^{(l)} &= \textrm{exp}^{K}_{\mathbf{x}} \Big(  P^{K}_{\mathbf{o}\rightarrow \mathbf{x}}(\textbf{b}^{(l)}) \Big) &&&&&&&&&&& \mbox{(vector addition)}\\
    && \sigma^{\otimes^{K}}\big( \textbf{x} \big) &= \textrm{exp}^{K}_{\mathbf{o}} \Big( \, \sigma \Big( \, \textrm{log}^{K}_{\mathbf{o}} (\textbf{x} ) \, \Big) \Big) &&&&&&&&&&& \mbox{(non-linearity)}
\end{align}
\end{subequations}
where each operation is performed in the appropriate tangent space, $\mathcal{T}_{\mathbf{o}}\mathcal{M}^{n,K}$ or $\mathcal{T}_{\mathbf{x}}\mathcal{M}^{n,K}$, such that each operation is mathematically consistent for each representation.  

\subsection{ROMA Message Passing Rule}
\label{ROMA_MP}
In the ROMA architecture, the renormalization procedure uses a first MPNN to embed the sugraph context
\begin{equation}
    \xb^{\sz} = \textrm{MPNN}^{\mathcal{M}}_{\textrm{pre}}\big(\hat{\ub}_{}^{\sz}, A^{\sz}\big) \in \mathcal{M}^{n_0,K}
\end{equation}
as well as two additional MPNNs for each coarse-graining step
\begin{align}
        S^{\sl} &= \textrm{MPNN}^{\mathcal{M};(l)}_{\textrm{CG}}\big(\xb^{\sl}, A^{\sl}\big) \in \mathcal{M}^{n_l,K} \\
        \zb^{\sl} &= \textrm{MPNN}^{\mathcal{M};\sl}_{\textrm{emb}}\big(\xb^{\sl}, A^{\sl}\big) \in \mathcal{M}^{n_l,K}
\end{align}
which compute the soft assignment and coarse-grained embeddings, respectively. For each MPNN, we use a common message passing rule

\begin{flalign} 
    \textbf{m}_{ij}^{(k+1)} &= \phi\bigg( \textrm{log}^{K}_{\mathbf{o}} ( \textbf{h}_i^{(k)} ) \, \Big|\Big| \, \textrm{log}^{K}_{\mathbf{o}} ( \textbf{h}_i^{(k)} ) - \textrm{log}^{K}_{\mathbf{o}} ( \textbf{h}_i^{(k)} ) \bigg) \\
    \underset{}{\textrm{AGG}_i}\bigg(\textbf{m}_{ij}^{(k+1)}\bigg) &= \psi\Big( \, \underset{j \in \mathcal{N}(i)}{\textrm{agg}_1}\big( \textbf{m}_{ij}^{(k+1)} \big) \, \Big|\Big|  \ldots \Big|\Big| \,  \underset{j \in \mathcal{N}(i)}{\textrm{agg}_m} \big( \textbf{m}_{ij}^{(k+1)}\big)\, \Big) \\
\end{flalign}
where $\phi$ and $\psi$ are MLPs, and the default aggregations are
\begin{equation}
    \textrm{agg}_m = [\textrm{sum}, \, \textrm{max}, \, \textrm{mean}, \, \textrm{var}]_m.
\end{equation}
This formulation integrates EdgeConv \cite{wang2019dynamic} and multiple aggregations \cite{corso2020principal} into the hyperbolic layers in (\ref{eq:hgcn}b). The functions $\phi$ and $\psi$ are two-layer, over-parameterized MLPs similar to the feed forward network of a transformer, which we find performs better than a linear projection or narrower and deeper MLP. 

\section{Experimental Details}

\subsection{Datasets}

Statistics for each dataset can be found in Table \ref{tab:dataset}.

\subsubsection*{Kuramoto Model}
Each Kuramoto model was simulated near the critical point $K_c=1.7$ for the graph configurations considered which have power law exponents of degree distribution $2.24 < \alpha < 2.29$ (Table \ref{tab:dataset}) . For numerical integration, we use the Dormand-Prince method with adaptive stepsize\footnote{\href{https://github.com/jax-ml/jax/blob/main/jax/experimental/ode.py}{https://github.com/jax-ml/jax/blob/main/jax/experimental/ode.py}} evaluated at equally spaced intervals $\Delta t = 5\times 10^{-3}$.

\subsubsection*{Burgers Dynamics}
The BD-3M example with update rule

\begin{equation}
    \frac{\partial u_i}{\partial t} \ \leftarrow \ \beta\frac{\partial u_i}{\partial t} + (1-\beta) \, \sqrt{ \frac{1}{k_i}\sum_{j\in \mathcal{N}(i)} \bigg(\frac{\partial u_j}{\partial t}\bigg)^2 } \\
\end{equation}
was simulated with $\beta=3/4$ and $m=6$ updates such that 80\% of the signal results from nonlinear interactions.

\subsection{Implementation Details}

\subsubsection*{Training}
For each neural operator, we follow the same recipe. We use the AdamW optimizer \cite{kingma2014adam,loshchilov2017decoupled} with a weight decay of $5\times 10^{-3}$. Similar to many transformer training procedures, we implement a learning rate scheduler with linear warmup of 10,000 steps followed by a linear decay schedule. To maintain stability during training, we use gradient clipping with a max norm of 1.0. \\

On the KM-\texttt{xx} datasets, each model is trained for $50,000 - 100,000$ iterations, while on the BD-3M dataset, each model is trained for $80,000 - 100,000$ iterations, depending on the experiment. In general, physics-informed models require a greater number of iterations to converge, but are able to continue learning with additional training. In such cases the number of iterations for all models in a particular experiment is increased.\\

Each GP model is trained for $10^3$ iterations with a learning rate of $5\times10^{-2}$ and an exact margninal log likelihood objective\footnote{\href{https://docs.gpytorch.ai/en/stable/_modules/gpytorch/mlls/exact_marginal_log_likelihood.html}{https://docs.gpytorch.ai/en/stable/\_modules/gpytorch/mlls/exact\_marginal\_log\_likelihood.html}}.

\subsubsection*{Model Details}

Each neural operator has latent dimension $d=3$, and the projection operation $\mathcal{P}: \mathbb{R}^d \rightarrow \mathbb{R}$ is
\begin{equation}
    \mathcal{P}(\bar{\sigma}) = ||\bar{\sigma}||_2 
\end{equation}
which is the flow speed of the latent field evaluated at the origin $\bar{\sigma}(t,\mathbf{y}=0)$. Additionally, each neural operator uses a Gaussian Random Field function space with a length scale $l=1.0$.

\subsubsection*{Computing Details}

All neural operators were trained on either an NVIDIA RTX A6000 GPU (48 GB) or NVIDIA GH200 Grace Hopper Superchip (96 GB). In particular, all models $>$324M parameters were trained on GH200s, and smaller models on  A6000s. Training times across all experiments ranged from 12-30h of wall time. 

\end{document}